\documentclass[conference]{IEEEtran}
\usepackage{times}  
\usepackage{helvet}  
\usepackage{courier}  
\usepackage[hyphens]{url}  
\usepackage{comment}
\usepackage{graphicx} 
\usepackage{caption} 
\usepackage{color}
\usepackage{subcaption}
\usepackage{algorithm, algorithmic}

\usepackage{amsmath,amsfonts,amssymb,amsthm}
\usepackage{multirow}
\usepackage{newfloat}
\usepackage{listings}

\newtheorem{theorem}{Theorem}
\newtheorem{proposition}{Proposition}

\newtheorem{assumption}{Assumption}
\newtheorem{lemma}{Lemma}
\newtheorem{observation}{Observation}

\def\N{\mathcal{N}}
\title{When MiniBatch SGD Meets SplitFed Learning: Convergence Analysis and Performance Evaluation}
\author{Chao Huang, Geng Tian, Ming Tang}

\begin{document}

\maketitle

\begin{abstract}
Federated learning (FL) enables collaborative model training across distributed clients (e.g., edge devices) without sharing raw data. Yet, FL can be computationally expensive as the clients need to train the entire model multiple times. SplitFed learning (SFL) is a recent distributed approach that alleviates computation workload at the client device by splitting the model at a cut layer into two parts, where clients only need to train part of the model. 
However, SFL still suffers from the \textit{client drift} problem when clients' data are highly non-IID. To address this issue, we propose MiniBatch-SFL. This algorithm incorporates MiniBatch SGD into SFL, where the clients train the client-side model in an FL fashion while the server trains the server-side model similar to MiniBatch SGD. 
We analyze the convergence of MiniBatch-SFL and show that the bound of the expected loss can be obtained by analyzing the expected server-side and client-side model updates, respectively. The server-side updates do not depend on the non-IID degree of the clients' datasets and can potentially mitigate client drift. However, the client-side model relies on the non-IID degree and can be optimized by properly choosing the cut layer. Perhaps counter-intuitive, our empirical result shows that a latter position of the cut layer leads to a smaller average gradient divergence and a better algorithm performance.  Moreover, numerical results show that MiniBatch-SFL achieves higher accuracy than conventional SFL and FL. The accuracy improvement can be up to 24.1\% and 17.1\% with highly non-IID data, respectively. 
\end{abstract}

\section{Introduction}
Federated learning (FL) \cite{mcmahan2017communication}  enables multiple clients to cooperatively train a global machine learning model while preserving data privacy. However, there are two major limitations of conventional FL algorithms.

\textbf{Limitation 1: computational-intensive local training}.  FL clients need to train the entire global model multiple times, which is  computational-intensive. Hence, clients (e.g., edge devices) may not be capable of accomplishing the training efficiently due to limited computational resources.

\textbf{Limitation 2: client drift under non-IID data}. FL algorithms may experience the \textit{client drift} problem under non-independent and identically distributed (non-IID) datasets among clients, leading to slow convergence and poor model performance. Despite that many existing studies proposed algorithms to address the non-IID issue (e.g., \cite{karimireddy2019scaffold,tang2023tackling,cui2022optimal, gao2022feddc,tan2023federated}, there are still potentials to further improve the model performance by making major modifications on the FL framework. 

\begin{figure}[t]
    \centering
    \includegraphics[height=5.5cm]{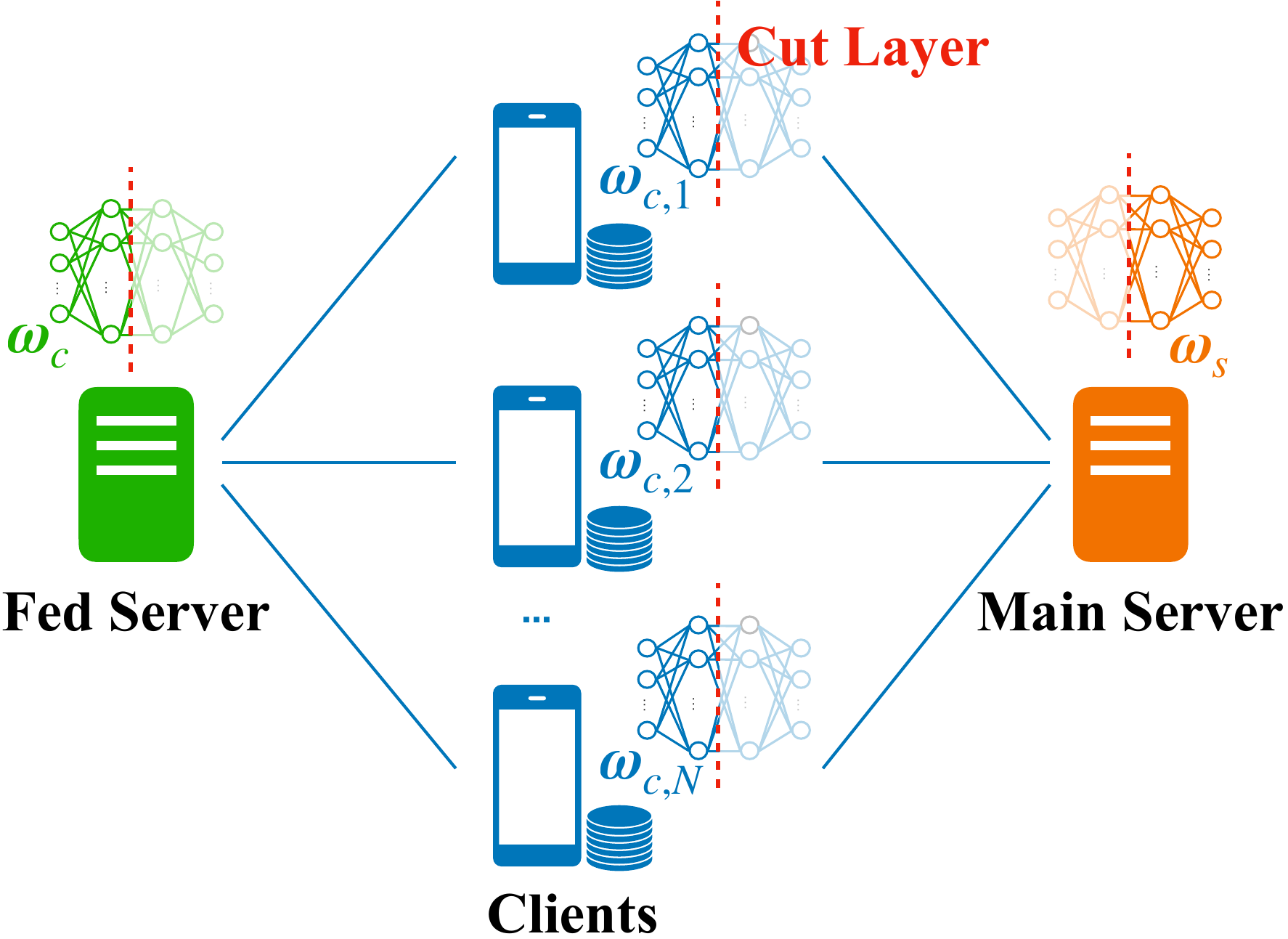}
    \caption{An illustration of SplitFed Learning framework. The global model $\boldsymbol{w}$ is split at the cut layers into two parts: client-side model $\boldsymbol{w}_c$  and server-side model $\boldsymbol{w}_s$. Each client $n$ trains its own version of client-side model $\boldsymbol{w}_{c,n}$, and the fed server periodically synchronizes all the client-side models. The main server trains the server-side model $\boldsymbol{w}_s$.}
    \label{fig:splitfed}
\end{figure}

One promising solution to addressing Limitation 1 is to incorporate split learning \cite{vepakomma2018split} into federated learning, known as SplitFed Learning (SFL) \cite{thapa2022splitfed}.  An illustration of SFL  is shown in Fig. \ref{fig:splitfed}. Specifically, 
the global model (to be trained) is first split at a cut layer into two parts: a client-side model and a server-side model. Then, the clients are responsible for training only the client-side model under the coordination of a \emph{fed server}. Another server, known as the \emph{main server}, is responsible for training the server-side model by collaborating with the clients. 
In SFL, since the clients only train a part of the model, 
 their training load can be reduced (compared to FL), which addresses Limitation 1.

However,  SFL may also suffer from slow convergence and a bad model performance under non-IID data due to the client drift problem. The experiments in \cite{Gao2022evaluation} showed that the performance of SFL decreases significantly when data becomes more non-IID (similar to FL). To address Limitation 2, one promising approach is MiniBatch SGD \cite{woodworth2020minibatch,gower2019sgd}. 
In MiniBatch SGD, the server takes all clients' stochastic gradients as one giant minibatch, calculates the averaged gradient, and then uses it to update the global model. This bears a similar spirit to centralized learning where the server updates the model using all clients' data (i.e., stochastic gradients), and hence MiniBatch SGD is immune to the non-IIDness across clients \cite{woodworth2020minibatch}. Despite that MiniBatch SGD has a good potential to address the client drift problem in SFL, it is still computation (and communication) intensive for clients.

In this work, we propose MiniBatch-SFL, an algorithm that incorporates MiniBatch SGD into SFL. In MiniBatch-SFL, the clients train the client-side model similar to SFL, which addresses Limitation 1. The key difference with SFL is that in MiniBatch-SFL, the main server trains the server-side model using the averaged gradients based on all clients' smashed data (i.e., output of the client-side models), which has a potential to address Limitation 2. 

Our work aims to answer three key questions as follows:
\begin{itemize}
    \item [Q1] What is the convergence rate of MiniBatch-SFL? 
    \item [Q2] How will the choice of the cut layer affect the performance of MiniBatch-SFL?
    \item [Q3] Will MiniBatch-SFL outperform FL and SFL under highly non-IID data? If so, by how much?
\end{itemize}

Answering question Q1 is important yet challenging. Different from prior FL analysis (e.g., \cite{li2019convergence, karimireddy2020scaffold}), the convergence of MiniBatch-SFL needs to take into account the dual-paced updates on both client-side and server-side models. That is, client-side models are updated in an FL fashion, while the server-side model is updated in a MiniBatch SGD fashion. Answering question Q2 is interesting yet non-trivial, because the split of the global model, controlled by the position of the cut layer, crucially affects the interaction between client-side and server-side models, and hence the performance of MiniBatch-SFL.  
To answer question Q3, we conduct simulations using various model structures on different datasets, and further provide insights into why MiniBatch-SFL outperforms the benchmarks. 

Our main contributions are as follows:
\begin{itemize}
\item We incorporate MiniBatch SGD into SFL, and propose a MiniBatch-SFL algorithm that enables low computational workload on edge clients while achieving good performance on highly non-IID data. We derive a theoretical upper 
bound of the expected loss of the trained global model, and show that the loss arising from clients-side model updates increases in the non-IIDness. However, the loss associated with the server-side model updates are independent of the non-IIDness. 
\item We empirically show that in MiniBatch-SFL, a latter position of the cut layer (i.e., more layers assigned to the client-side model) leads to a higher accuracy. This result is perhaps counter-intuitive. The reason is that when the client-side model has more layers (and parameters), it tends to update less aggressively and hence is less prone to the client drift problem. This is also consistent with our observation that more client layers lead to a smaller average gradient divergence of the client-side model under highly non-IID data. 
\item We conduct simulations using various model structures on different datasets. We show that MiniBatch-SFL converges faster and achieves a higher accuracy than conventional SFL and FL. The improvement can be up to 24.1\% and 17.1\%, respectively, under highly non-IID data. 
\end{itemize}

The rest of the paper is organized as follows. We first discuss related work. Next, we present the proposed algorithm MiniBatch-SFL, and then analyse its convergence. 
Finally, we present the numerical experiments.

\section{Related Work}
Federated learning \cite{mcmahan2017communication} and split learning (SL) \cite{vepakomma2018split} are two major distributed approaches that enable collaborative model training without accessing raw data. FL enables parallel model training across clients, but it can be computationally expensive for resource-constrained edge devices. SL alleviates the computational overhead by splitting the model into two parts, where the clients only need to train the first part of the model. However, clients in SL need to sequentially interact with the main server, which leads to a high latency.

SplitFed learning (SFL) was proposed in \cite{thapa2022splitfed} that combines the advantages of parallel training from FL and model splitting from SL. Some excellent recent studies improve SFL from various aspects, e.g., \cite{han2022splitgp, kafshgari2023smart,han2021accelerating,li2022mocosfl,khan2022security}. For example, \cite{ han2021accelerating,han2022splitgp} accelerates client-side model updates using locally generated loss. \cite{li2022mocosfl} combines momentum contrast with SFL to enable large-scale distributed  training.  \cite{khan2022security} analyzed the vulnerability of SFL w.r.t. model poisoning attacks. 

Although the algorithms proposed in the aforementioned works achieve good performance, they focused on addressing Limitation 1 (and/or improve privacy protection of SFL). However, SFL still suffers from the client drift problem when clients' data are highly non-IID \cite{Gao2022evaluation}. Different from prior studies, our work focuses on addressing Limitation 2 by proposing a new algorithm that incorporates MiniBatch SGD into SFL, which will be shown to accelerate convergence and boost model performance under highly non-IID data. 
 In addition, in a hope to understand whether and why our algorithm works, we proceed from both theoretical and empirical perspectives to answer the aforementioned questions Q1-Q3. 


\section{MiniBatch-SFL Algorithm}\label{sec: algorithm}
We first introduce some preliminary notations and then provide the algorithmic details of MiniBatch-SFL.
\subsection{Preliminaries}
We consider a set of clients, denoted by  $\N=\{1,2,\cdots, N\}$. Each client $n\in\N$ has a private dataset $\mathcal{D}_n$ of size $D_n=|\mathcal{D}_n|$. Each data sample contains a feature and a label. These clients in set $\N$ aim  to collectively train a machine learning model, denoted by  $\boldsymbol{w}$, that maps from feature to label using their private datasets. This model $\boldsymbol{w}$  consists of $L$ layers. It is split at the $L_c$-th layer (called cut layer) into two parts: a client-side model $\boldsymbol{w}_c$ (from the the first layer to layer $L_c$); a server-side model $\boldsymbol{w}_{s}$ (from layer $L_c+1$ to layer $L$). Note that $\boldsymbol{w}=[\boldsymbol{w}_c; \boldsymbol{w}_{s}]$. 
The clients  in set $\N$ cooperatively perform training over the client-side model $\boldsymbol{w}_c$ under the coordination of an \emph{fed server} (see Fig. \ref{fig:splitfed}), similar to conventional FL. Let $\boldsymbol{w}_{c,n}$ denote the local client-side model of client $n$. Meanwhile, a \emph{main server} is responsible for performing training over the server-side model $\boldsymbol{w}_{s}$ through interacting with the clients (see Fig. \ref{fig:splitfed}).


Consider model $\boldsymbol{w}$ obtained through the training process. Let $f_n (\boldsymbol{w}; \zeta_n)$ denote the loss of model $\boldsymbol{w}$ over client $n$'s mini-batch instance $\zeta_n$, which is randomly sampled from client $n$'s dataset $\mathcal{D}_n$. 
Let $ f_n (\boldsymbol{w})\triangleq \mathbb{E}_{\zeta_n\sim \mathcal{D}_n}[ f_n (\boldsymbol{w}; \zeta_n)]$ denote the expected loss of model $\boldsymbol{w}$ over client $n$'s dataset. The goal of  MiniBatch-SFL algorithm is to minimize the expected loss of the model $\boldsymbol{w}$ over the datasets of all clients:
\begin{equation}
\min_{\boldsymbol{w}} F(\boldsymbol{w}) \triangleq \sum_{n\in \mathcal{N}}p_n f_n (\boldsymbol{w}),
\end{equation}
where coefficient $p_n\geq 0$ denotes the weight assigned to client $n$'s dataset, and $\sum_{n\in \mathcal{N}}p_n=1$. We  typically have $p_n=D_n/(\sum_{n'\in \mathcal{N}}D_{n'})$. 

For ease of presentation on MiniBatch-SFL, we rewrite the loss function $f_n(\boldsymbol{w}; \zeta_n) $ as follows:
\begin{equation}
f_n(\boldsymbol{w}; \zeta_n) = h(\boldsymbol{w}_s; u(\boldsymbol{w}_{c,n}; \zeta_n)),
\end{equation}
where $u$ is a client-side function that maps the input data (of the sampled mini-batch instance $\zeta_n$) to the activation space (i.e., the output of client-side model $\boldsymbol{w}_{c,n}$), and $h$ is the server-side function that maps the activation space to a scalar loss value.

\begin{algorithm}[tb]
\caption{MiniBatch-SFL}
\label{alg:algorithm}
\textbf{Input}: $N, T, E, M, \mathcal{I},\mathcal{I}_G$, $\eta_c^i, \eta_s^i$\\
\textbf{Output}: $\boldsymbol{w}^{TEM}=[\boldsymbol{w}_c^{TEM}; \boldsymbol{w}_s^{TEM}]$\\
\textbf{Initialization}: $\boldsymbol{w}^{0}=[\boldsymbol{w}_c^{0}; \boldsymbol{w}_s^{0}]$
\begin{algorithmic}[1] 
\FOR{$i=1,2,\cdots, TEM$}
\vspace{1mm}
    \STATE \textit{\#Client forward pass in parallel} \\
    \FOR{each client $n\in \mathcal{N}$}
    \STATE Sample a mini-batch data $\zeta_n^{i-1}$ and compute $\boldsymbol{z}_n^{i-1}$
    \STATE  Send $\boldsymbol{z}_n^{i-1}$ (together with labels) to main server
    \ENDFOR
    \vspace{1mm}
\STATE \textit{\#Main server update}
\STATE \textbf{Main server}:
\STATE Compute server-side gradients using $\{\boldsymbol{z}_n^{i-1}\}_{n \in \mathcal{N}}$
\STATE Update $\boldsymbol{w}_s^i$ using averaged gradients in (\ref{eq:main-update})
\STATE Send gradient w.r.t. smashed data to clients $n\in \mathcal{N}$
\vspace{1mm}
\STATE \textit{\#Client backward pass in parallel}
    \FOR{each client $n\in \mathcal{N}$}
    \STATE Compute client-side gradients using chain rule
    \STATE Update client-side model $\boldsymbol{v}_{c,n}^i$ in (\ref{backward-prop}) 
    \ENDFOR
    \vspace{1mm}
\STATE \textit{\#Possible fed server synchronization}
\STATE \textbf{Fed server}:
\IF {$i\in \mathcal{I}_G$}
\STATE Aggregate client-side models in (\ref{eq:agg})
\ENDIF
\ENDFOR
\end{algorithmic}
\end{algorithm}

\subsection{Algorithm Description}
We first provide an overview of MiniBatch-SFL and then present the algorithmic details.  

\emph{Overview}: The MiniBatch-SFL algorithm continues for $T$ rounds. At the beginning of round $t$, the fed server sends the recent global client-side model to all clients. Then, clients and the main server perform training for $E$ epochs over the client-side and server-side models, respectively. In each epoch, the dataset of each client $n\in\N$ is randomly partitioned into $M$ mini-batches. Client $n$ and the main server cooperate to performs one stochastic gradient descent (SGD) step over each of these mini-batches to update their client-side and server-side models, respectively.  After the $E$ epochs,  the fed server aggregates the local client-side models of the clients to update the global client-side model.

\emph{Algorithm Details}: For ease of presentation, we introduce MiniBatch-SFL by explaining the actions of clients, the main server, and the fed server associated with each SGD step. Let $\mathcal{I}\triangleq \{1,2,\cdots, TEM\}$ denote the set of SGD steps across the entire training process. The SGD step over the $m^{\rm th}$ mini-batch in the $e^{\rm th}$ epoch of round $t$ is denoted as the $i^{\rm th}$ SGD step, where $i=(t-1)EM+(e-1)M+m$. Fed server synchronization  corresponds to the event when clients send their local client-side models to the fed server for global client-side model aggregation. We define set $\mathcal{I}_{G}\triangleq \{tEM| t=1, \cdots, T\}$. After the $i^{\rm th}$ SGD step, where $i\in \mathcal{I}_{G}$, the fed server performs synchronization. We use superscript to denote the SGD step in the rest of this paper.


We use $\boldsymbol{v}_{c,n}^i$ to represent the local client-side model of client $n$ immediately after the $i^{\rm th}$ SGD step. Let $\boldsymbol{w}_{c,n}^{i}$ denote the client-side model of client $n$ after the  synchronization associated with the $i^{\rm th}$ SGD step (if $i\in\mathcal{I}_G$). Let $\boldsymbol{w}_s^{i}$ denote the server-side model after the $i^{\rm th}$ SGD step.
In $i^{\rm th}$ SGD step, MiniBatch-SFL works as follows.
\begin{enumerate}
\item \textbf{Client Forward Pass}: Each client $n\in\N$ computes in parallel the output of the client-side model $\boldsymbol{z}_n^{i-1}=u(\boldsymbol{w}_
{c,n}^{i-1}; \zeta_n^{i-1})$, using the randomly sampled mini-batch $\zeta_n^{i-1}$ as the input. The output $\boldsymbol{z}_n^{i-1}$ is also known as smashed data. Then, each client $n$  
 sends the output $\boldsymbol{z}_n^{i-1}$ to the main server.
\item \textbf{Main Server Update}: The main server treats all the smashed data $\{\boldsymbol{z}_{n}^{i-1}\}_{n \in \mathcal{N}}$ as inputs, computes the averaged gradient, and performs one step of gradient descent (GD) on the server-side model:
\begin{equation}\label{eq:main-update}
\begin{aligned}
\hspace{-2mm}\boldsymbol{w}_{s}^{i} 
= \boldsymbol{w}_{s}^{i-1}  - \eta_s^{i-1} \frac{\sum_{n \in \mathcal{N}} p_n\nabla_{\boldsymbol{w}_s}h(\boldsymbol{w}_s^{i-1}; \boldsymbol{z}_n^{i-1})}{\sum_{n\in \mathcal{N}}p_n},
\end{aligned}
\end{equation}
where $\eta_s^{i-1}$ is the associated learning rate at the  server side. The main server also computes the gradients with respect to the smashed data, i.e., $\nabla_{z_n^{i-1}}h(\boldsymbol{w}_s^{i-1}; \boldsymbol{z}_n^{i-1}) $, and sends it back to the corresponding client $n$.
\item \textbf{Client Backward Pass}: Each client $n\in \mathcal{N}$ computes the gradient of the client-side model using chain rule and performs GD:
\begin{equation}\label{backward-prop}
\begin{aligned}
\boldsymbol{v}_{c,n}^{i} &= \boldsymbol{w}_{c,n}^{i-1}-\eta_c^{i-1} \nabla_{\boldsymbol{w}_c} f_n(\boldsymbol{w}^{i-1}; \zeta_n^{i-1})\\
& \hspace{-8mm}= \boldsymbol{w}_{c,n}^{i-1}-\eta_c^{i-1} \nabla_{\boldsymbol{z}_n^{i-1}} h(\boldsymbol{w}_s^{i-1}; \boldsymbol{z}_n^{i-1})\nabla_{\boldsymbol{w}_c}u(\boldsymbol{w}^{i-1}_{c,n}; \zeta_n^{i-1}),
\end{aligned}
\end{equation}
where $\eta_c^{i-1}$ denotes the associated learning rate at the client side. 
\item \textbf{Possible Fed Server Synchronization}: When $i \in \mathcal{I}_G$, the fed server synchronizes (i.e., aggregates) the client-side models. When $i \in \mathcal{I}\setminus \mathcal{I}_G$, no synchronization is performed. Hence, we have
\begin{equation}\label{eq:agg}
\begin{aligned}
\boldsymbol{w}_{c,n}^{i}=\begin{cases}
v_{c,n}^{i}, \quad &\text{if}\quad i\in \mathcal{I}\setminus  \mathcal{I}_G, \\
\sum_{n \in \mathcal{N}}p_n v_{c,n}^{i}, \quad &\text{if}\quad i\in \mathcal{I}_G.
\end{cases}
\end{aligned}
\end{equation}
\end{enumerate}

Different choices of cut layer leads to different client-side models $\boldsymbol{w}_{c}^{i}$ and server-side models $\boldsymbol{w}_{s}^{i}$, and hence  different full models $\boldsymbol{w}^i=[\boldsymbol{w}_c^i; \boldsymbol{w}_{s}^i]$. Importantly, the client-side model is trained by clients cooperatively in a distributed FL fashion, which relies on the non-IIDness of the clients' datasets. In contrast, the server-side model is trained by the main server similar to MiniBatch SGD in a ``centralized'' fashion, which will be shown to be independent of the non-IIDness. This is the main reason why MiniBatch-SFL may outperform SFL (and FL), as it mitigates the client drift problem by updating the server-side model in a centralized fashion that is robust to clients' non-IIDness.

In the following, we first analyze the convergence of MiniBatch-SFL. In particular, we focus on understanding how the non-IID degree of the clients' datasets affects the client-side model, the server-side model,  hence the full global model.  Then, we provide empirical analysis on the impact of cut layer on the algorithm performance.

\section{Convergence Analysis}\label{sec: convergence}
In this section, we present a convergence analysis of MiniBatch-SFL, focusing on the impact of non-IIDness. We start with the technical assumptions and then provide the theoretical performance bound. 
\subsection{Assumptions}
We make some commonly adopted assumptions in the distributed learning literature \cite{li2019convergence, woodworth2020minibatch}. Let $\nabla f_n\left(\mathrm{\boldsymbol{w}}\right)\triangleq \mathbb{E}_{\zeta_n} \nabla f_n(\boldsymbol{w}; \zeta_n) $.
\begin{assumption}{(S-smoothness)}\label{asm:lipschitz_grad}
For each client $n$, $ f_n(\boldsymbol{y}) \le f_n(\boldsymbol{x})+ \nabla f_n(\boldsymbol{x})^T(\boldsymbol{y}-\boldsymbol{x})+\frac{S}{2}||\boldsymbol{y}-\boldsymbol{x}||^2, \forall \boldsymbol{x}, \boldsymbol{y}.$
\end{assumption}

\begin{assumption}{($\mu$-strong convexity)}\label{asm:strong-convexity}
For each client $n$, $ f_n(\boldsymbol{y}) \ge f_n(\boldsymbol{x})+ \nabla f_n(\boldsymbol{x})^T(\boldsymbol{y}-\boldsymbol{x})+\frac{\mu}{2}||\boldsymbol{y}-\boldsymbol{x}||^2, \forall \boldsymbol{x}, \boldsymbol{y}$.
\end{assumption}

\begin{assumption}{(Bounded stochastic gradient variance)}\label{asm:bounded-grad-variance}
For each client $n$,  $\mathbb{E}_{\zeta_n}\left\Vert \nabla f_n(\boldsymbol{w}; \zeta_n) - \nabla f_n\left(\mathrm{\boldsymbol{w}}\right) \right\Vert^2  \le \sigma_n^2, \forall \boldsymbol{w}$.
\end{assumption}
\begin{assumption}{(Bounded stochastic gradient square norm)}\label{asm:bound-grad-norm}
 For each client $n$, $\mathbb{E}_{\zeta_n}\left\Vert \nabla f_n(\boldsymbol{w}; \zeta_n) \right\Vert^2  \le R^2, \forall \boldsymbol{w}$.
\end{assumption}
\begin{assumption}{(Bounded gradient divergence)}\label{asm:unbiased_global}
 For each client $n$, $\left\Vert \nabla f_n\left(\mathrm{\boldsymbol{w}}\right)-\nabla F\left(\mathrm{\boldsymbol{w}}\right)  \right\Vert^2 \le  \delta^2, \forall \boldsymbol{w}$.
\end{assumption}
A larger  $\delta^2$  implies a higher degree of non-IIDness. 


\begin{figure*}
       \begin{subfigure}{0.24\textwidth}
        \centering
        \includegraphics[height=3.4cm]{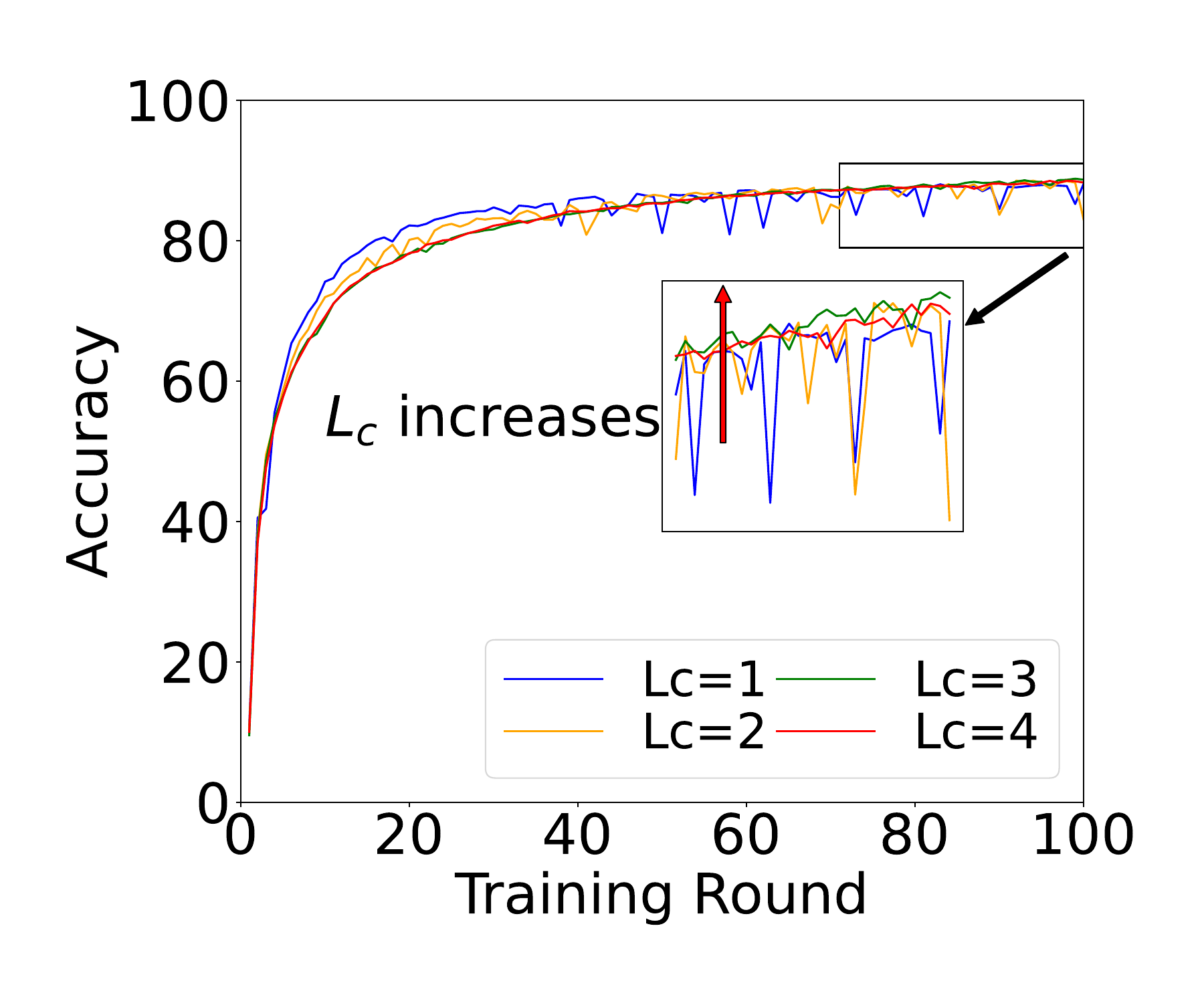}
        \caption{$r\%=0$ (CIFAR-10).}
        \label{subfig:accuracy0.5}
    \end{subfigure}
    \hfil
    \begin{subfigure}{0.24\textwidth}
        \centering
        \includegraphics[height=3.4cm]{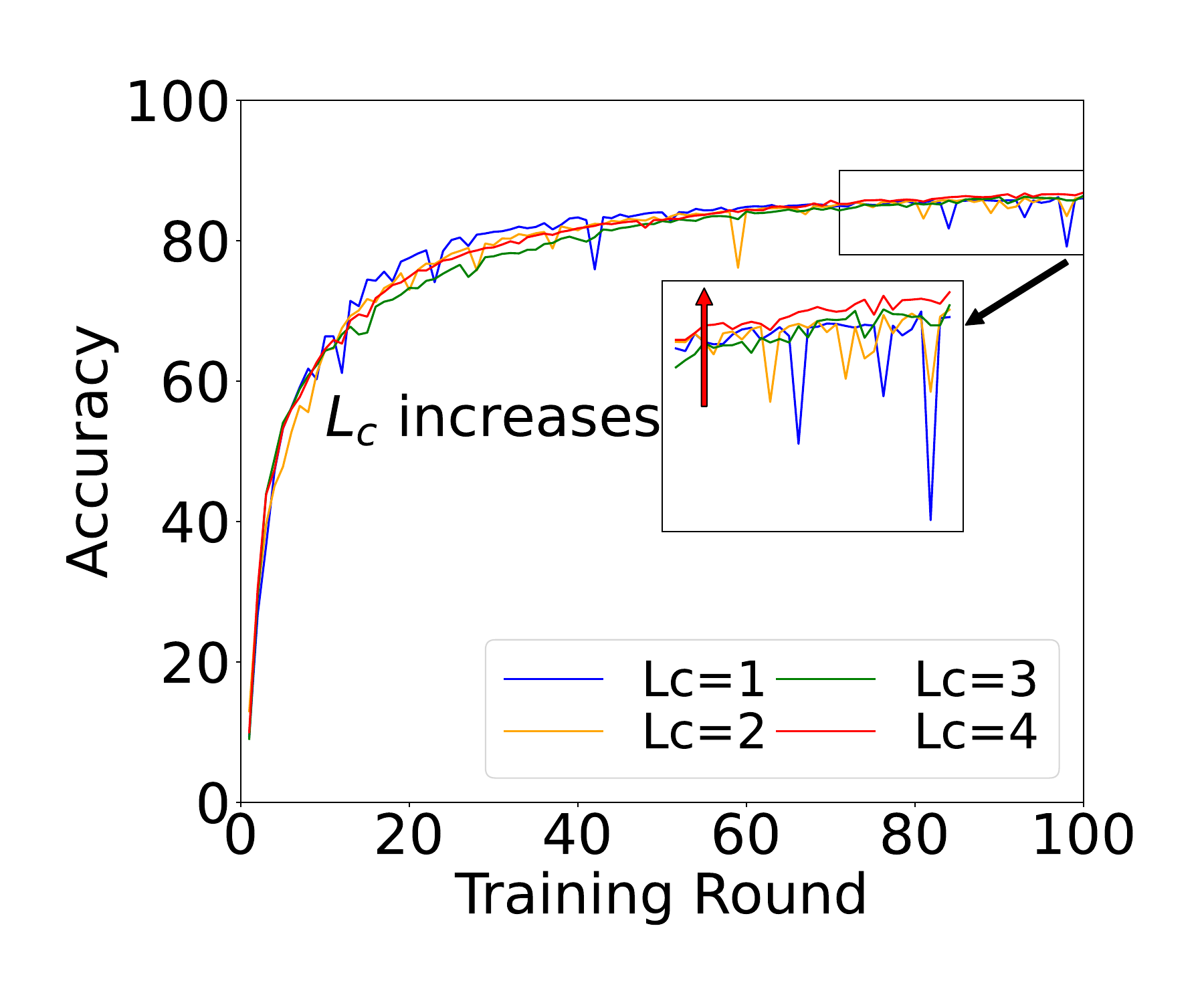}
        \caption{$r\%=0.5$ (CIFAR-10).}
        \label{subfig:accuracy0.5}
    \end{subfigure}
    \hfil
    \begin{subfigure}{0.24\textwidth}
        \centering
        \includegraphics[height=3.4cm]{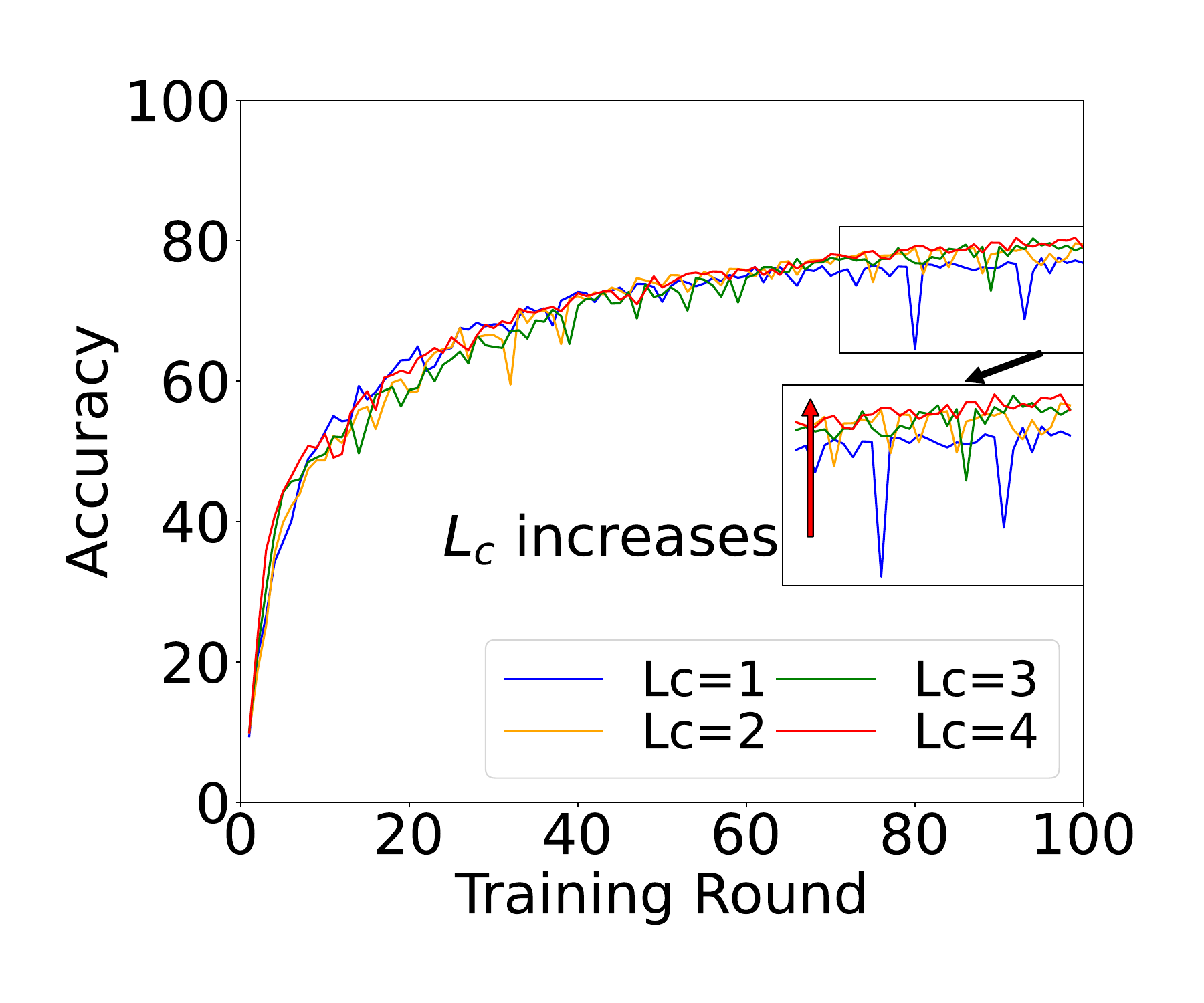}
        \caption{$r\%=0.8$ (CIFAR-10).}
        \label{subfig:accuracy0.8}
    \end{subfigure}
    \hfil 
    \begin{subfigure}{0.24\textwidth}
        \centering
        \includegraphics[height=3.4cm]{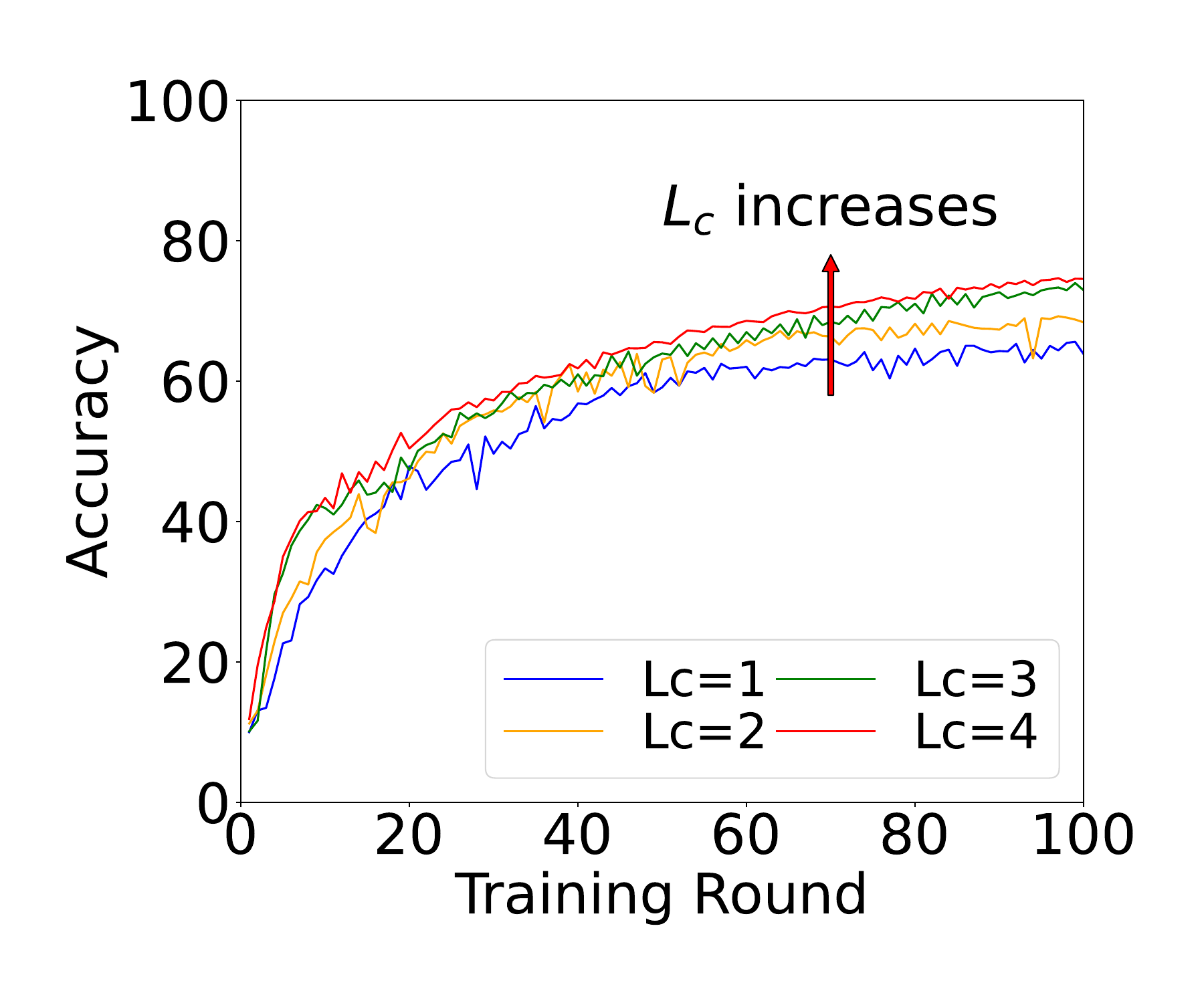}
        \caption{$r\%=0.9$ (CIFAR-10).}
        \label{subfig:accuracy0.9}
    \end{subfigure}
    \hfil \\
    \begin{subfigure}{0.24\textwidth}
        \centering
        \includegraphics[height=3.4cm]{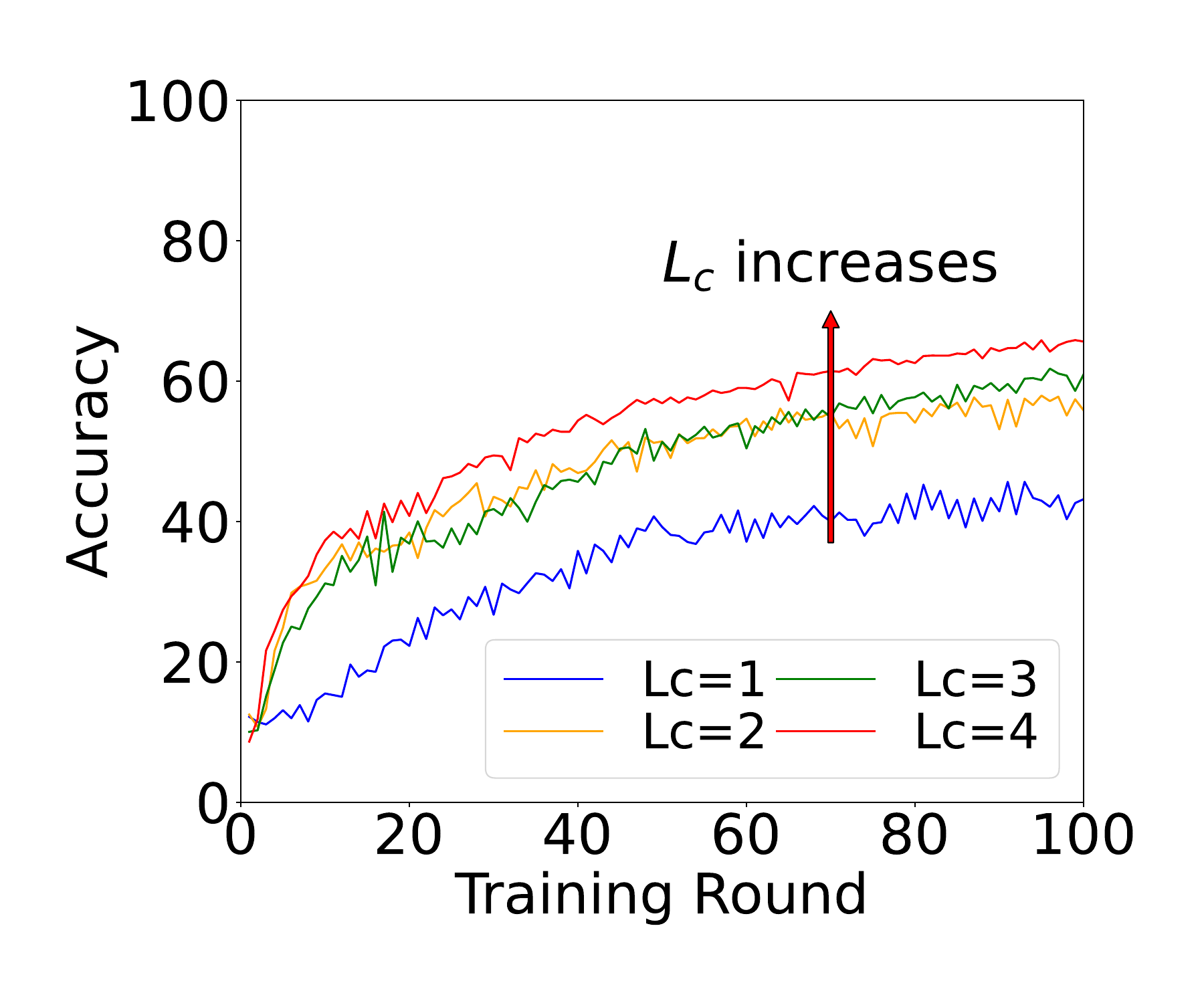}
        \caption{$r\%=0.95$ (CIFAR-10).}
        \label{subfig:accuracy0.95}
    \end{subfigure}
    \hfil
        \begin{subfigure}{0.24\textwidth}
        \centering
        \includegraphics[height=3.4cm]{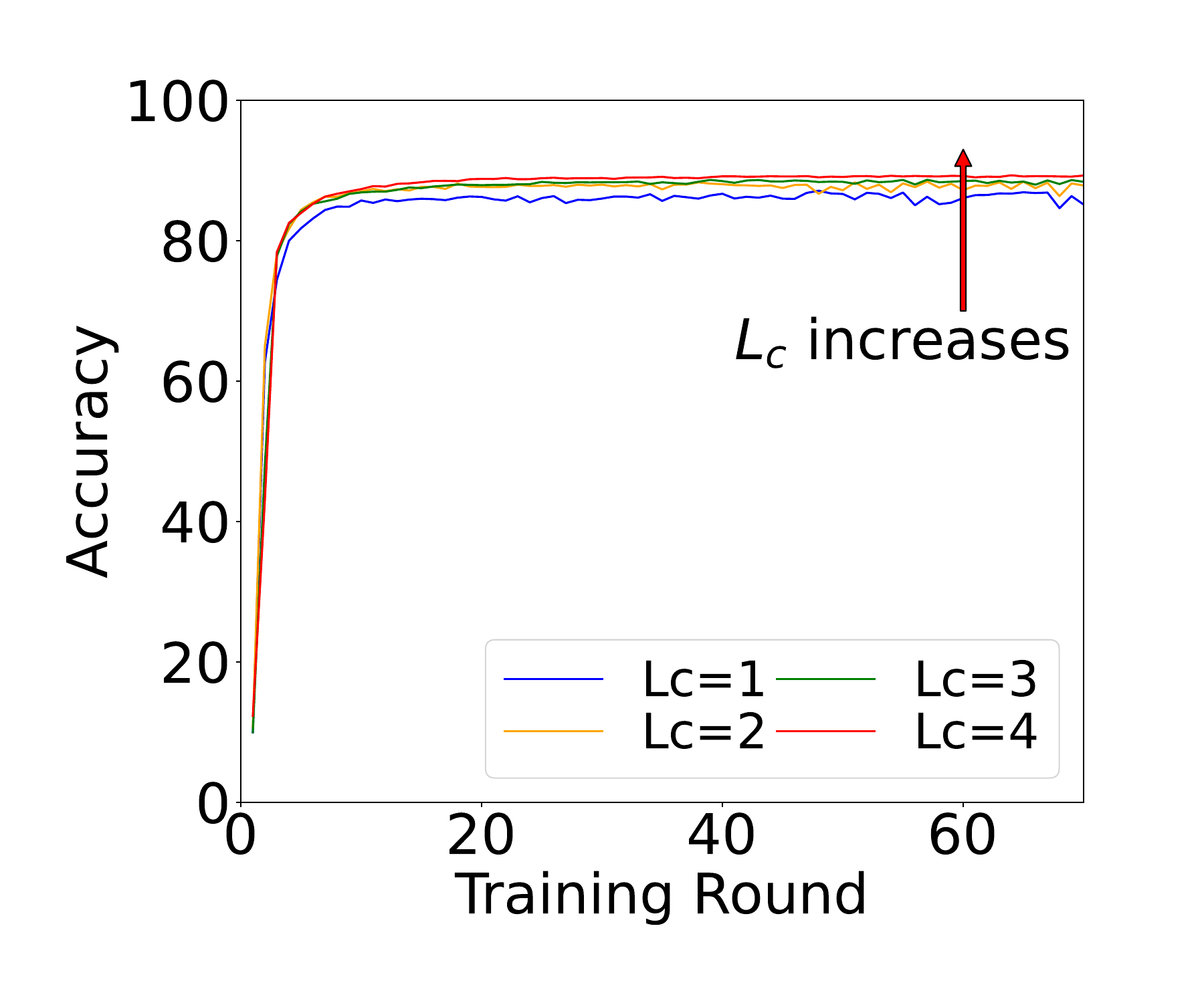}
        \caption{$r\%=0.8$ (FMNIST).}
    \end{subfigure}
    \hfil
    \begin{subfigure}{0.24\textwidth}
        \centering
        \includegraphics[height=3.4cm]{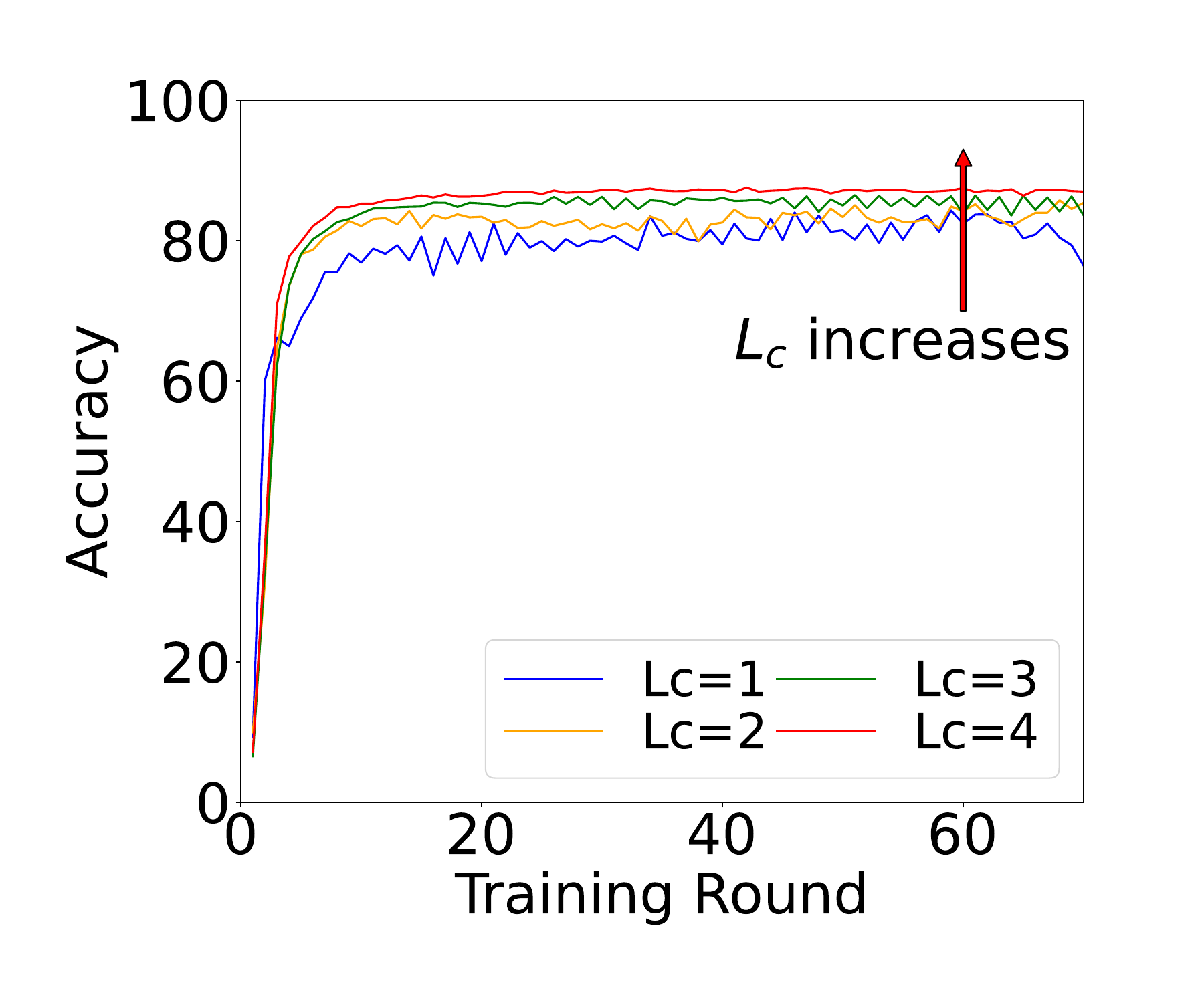}
        \caption{$r\%=0.9$ (FMNIST).}
    \end{subfigure}
    \hfil
    \begin{subfigure}{0.24\textwidth}
        \centering
        \includegraphics[height=3.4cm]{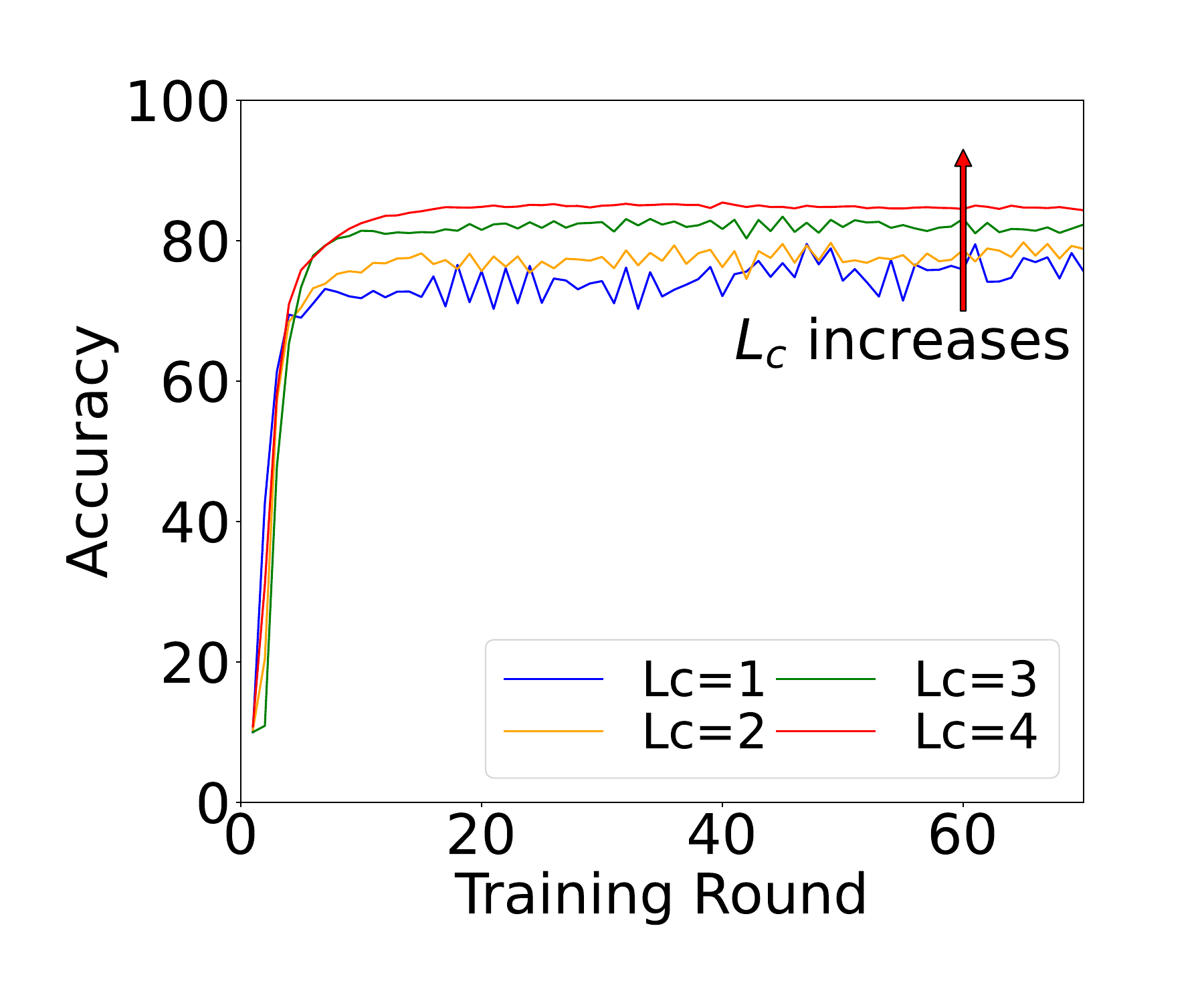}
        \caption{$r\%=0.95$ (FMNIST).}
    \end{subfigure}
    \caption{Impact of cut layer under different non-IID ratios: (a-e) CIFAR-10, (f-h) FMNIST.}
    \label{fig:cut-acc-cifar}
\end{figure*}

\subsection{Theoretical Performance Bound} 
Analyzing the performance bound of MiniBatch-SFL  is  more challenging than that of conventional FL counterparts due to the dual-paced model updates. That is, clients update the client-side model in a parallel distributed fashion (i.e., an FL fashion), while the main server updates the model in a centralized fashion. We address this challenge by decomposing the analysis into two parts: the server-side and client-side updates. We give the decomposition procedure below.

Let $\boldsymbol{w}^*\triangleq[\boldsymbol{w}_c^*; \boldsymbol{w}_s^*]$ denote the optimal global model that minimizes the expected loss function $F(\boldsymbol{w})$. Recall that $\boldsymbol{w}^{TEM} \triangleq [\boldsymbol{w}^{TEM}_c; \boldsymbol{w}^{TEM}_s]$ is the global model obtained using MiniBatch-SFL, where $TEM$ is the index of the final SGD step. We show that the expected loss of the global model obtained using our proposed algorithm and that of the optimal global model is upper-bounded by two terms, which are associated with the server-side and client-side models respectively.  We formalize this result in Lemma \ref{lemma: 1}. 
\begin{lemma}[Decomposition]\label{lemma: 1}
Under Assumption \ref{asm:lipschitz_grad}, we have 
\begin{equation}
\begin{aligned}
&\mathbb{E}\left[F(\boldsymbol{w}^
{TEM})\right]- F(\boldsymbol{w}^*)\\
&\le\frac{S}{2}\left(\mathbb{E}||\boldsymbol{w}_s^{TEM}-\boldsymbol{w}_s^*||^2+\mathbb{E}||\boldsymbol{w}_c^{TEM}-\boldsymbol{w}_c^*||^2\right).
\end{aligned}
\end{equation}
\end{lemma}
\begin{proof}
According to Assumption \ref{asm:lipschitz_grad}, we can prove that the loss function  $F(\cdot)$ is $S$-smooth. Thus, we have 
\begin{equation}\label{eq:composition1}
\begin{aligned}
\mathbb{E}\left[F(\boldsymbol{w}^
{TEM})\right]- F(\boldsymbol{w}^*)
\le \mathbb{E}\left[\langle\boldsymbol{w}^{TEM}-\boldsymbol{w}^*,\nabla F(\boldsymbol{w}^*)\rangle\right]\\
+\frac{S}{2}\mathbb{E}\left[||\boldsymbol{w}^{TEM}-\boldsymbol{w}^*||^2\right]=\frac{S}{2}\mathbb{E}\left[||\boldsymbol{w}^{TEM}-\boldsymbol{w}^*||^2\right].
\end{aligned}
\end{equation}
Since  $\boldsymbol{w}^{TEM}\triangleq [\boldsymbol{w}_c^{TEM};\boldsymbol{w}_s^{TEM}]$, and $\boldsymbol{w}^*\triangleq [\boldsymbol{w}_c^*; \boldsymbol{w}_s^*]$, the following equality holds:
   \begin{equation}\label{eq:composition2}
   \begin{aligned}
   &\mathbb{E}\left[||\boldsymbol{w}^{TEM}-\boldsymbol{w}^*||^2\right]\\
   =&\mathbb{E}\left[||[\boldsymbol{w}_c^{TEM};\boldsymbol{w}_s^{TEM}]-[\boldsymbol{w}_c^*; \boldsymbol{w}_s^*]||^2\right]\\
   =&\mathbb{E}\left[||[\boldsymbol{w}_c^{TEM}-\boldsymbol{w}_c^*; \boldsymbol{w}_s^{TEM}-\boldsymbol{w}_s^*]||^2\right]\\
   =&\mathbb{E}\left[||\boldsymbol{w}_c^{TEM}-\boldsymbol{w}_c^*||^2\right]+\mathbb{E}\left[||\boldsymbol{w}_s^{TEM}-\boldsymbol{w}_s^*||^2\right].
   \end{aligned}
   \end{equation}
   Substituting \eqref{eq:composition2} into \eqref{eq:composition1}, we complete the proof.
\end{proof}
Based on Lemma \ref{lemma: 1}, to bound the loss of the global model, it suffices to separately bound the expected gap between the model obtained using MiniBatch-SFL and the optimal model at the server-side and that of the client-side.

Next, we bound the expected model gap at the server-side, i.e., $\mathbb{E}||\boldsymbol{w}_s^{TEM}-\boldsymbol{w}_s^*||^2$, in Proposition \ref{bounding-server-side}. The proof is given in the supplementary material. For presentation simplicity, we define a coefficient $\gamma$ that will be useful in choosing the learning rates, where $\gamma\triangleq \max\left\{8S/\mu-1, EM\right\}$.
\begin{proposition}\label{bounding-server-side}
 Under Assumptions \ref{asm:lipschitz_grad}-\ref{asm:bound-grad-norm}, if we  choose $\eta_s^{i}=2/(\mu(\gamma+i))$, then  
\begin{equation}\label{server-side-result}
   \begin{aligned}
   \mathbb{E}||\boldsymbol{w}_s^{TEM}-\boldsymbol{w}_s^*||^2\le \frac{8R^2+ \mu^2N^2(\gamma+1)||\boldsymbol{w}_s^0-\boldsymbol{w}_s^*||^2}{\mu^2N^2(\gamma+TEM)}.
   \end{aligned}
   \end{equation}
   where $\boldsymbol{w}_s^0$ is the initialized server-side model.
\end{proposition}
Note that the bound of $\mathbb{E}||\boldsymbol{w}_s^{TEM}-\boldsymbol{w}_s^*||^2$ in \eqref{server-side-result} is irrelevant to $\delta^{2}$, which characterizes the non-IID degree of clients' datasets. This is because in MiniBatch-SFL, even if clients have non-IID data, the main server uses the smashed data of all clients' mini-batches (as one giant mini-batch) to update the server-side model, see \eqref{eq:main-update}. Thus, the server-side training is similar to centralized learning which ``cancels out'' the impact of non-IIDness on the server-side model. 

Then, we bound the expected model gap at the client-side, i.e., $\mathbb{E}\left[||\boldsymbol{w}_c^{TEM}-\boldsymbol{w}_c^*||^2\right]$. The proof is given in the supplementary material. Recall that ${\boldsymbol{w}}_c^{TEM}\triangleq \sum_{n \in \mathcal{N}} p_n \boldsymbol{w}_{c,n}^{TEM}$ is the global client-side model after the entire training process.  Let ${\boldsymbol{w}}_c^0$ denote the initial global client-side model, which is randomly generated by the fed server. 

\begin{proposition}\label{bounding-client-side}
Under Assumptions \ref{asm:lipschitz_grad}- \ref{asm:unbiased_global}, if we choose $\eta_c^i=2/(\mu(\gamma+i))$, then 
\begin{equation}\label{client-side-result}
\mathbb{E}||{\boldsymbol{w}}_c^{TEM}-\boldsymbol{w}_c^*||^2
  \le \frac{4H+\mu^2(\gamma+1)\mathbb{E}\left[||{\boldsymbol{w}}_c^0-\boldsymbol{w}_c^*||^2\right]}{\mu^2(\gamma+TEM)},
\end{equation}
where $H\triangleq 6EMR^2+12E^2M^2\delta^2 +6S\Gamma+  \sum_{n\in \mathcal{N}}p_n^2\sigma_n^2$, with $\Gamma\triangleq  F(\boldsymbol{w}^*)-\sum_{n \in \mathcal{N}}p_n f_n(\boldsymbol{w}_{c,n}^*, \boldsymbol{w}_s)$.
\end{proposition}
According to Proposition \ref{bounding-client-side}, as  $\delta^2$ increases (i.e., the non-IID degree of the clients' datasets increases), the bound of $\mathbb{E}\left[||\boldsymbol{w}_c^{TEM}-\boldsymbol{w}_c^*||^2\right]$ in \eqref{client-side-result} increases. This is due to the fact that clients perform model training in an FL fashion, which leads to the client drift issue. Thus, as in conventional FL, a higher non-IID degree of the clients' datasets leads to a larger gap between the global client-side model obtained from MiniBatch-SFL and the optimal client-side model. 

Now, we present the convergence result in Theorem \ref{final-convergence}.
\begin{theorem}{}\label{final-convergence}
Under Assumptions \ref{asm:lipschitz_grad}-\ref{asm:unbiased_global}, if we choose $\eta_c^{i}=\eta_s^{i}=2/(\mu(\gamma+i))$, then the gap between the expected loss obtained using MiniBatch-SFL and the minimum loss is upper-bounded by 
\begin{equation}
\begin{aligned}
\mathbb{E}\left[F(\boldsymbol{w}^
{TEM})\right]- F(\boldsymbol{w}^*)\le S\left(L_s+L_c\right)/2, 
\end{aligned}
\end{equation}
where $L_s$ and $L_c$ are defined as the right-hand side of (\ref{server-side-result}) and (\ref{client-side-result}), respectively.
\end{theorem}
Theorem \ref{final-convergence} follows from Lemma \ref{lemma: 1} and Propositions \ref{bounding-server-side}-\ref{bounding-client-side}.


\textbf{Theorem \ref{final-convergence} answers Q1}. Similar to SFL and FL, a larger non-IID degree leads to a slower algorithm convergence in MiniBatch-SFL, mainly because the client-side models still face the client-drift issue under non-IID data. Fortunately, Theorem \ref{final-convergence} indicates that the expected loss can be reduced (i.e., the convergence rate can be increased) by either reducing the optimality gap at the server-side or at the client-side. Since the server-side training can be regarded as  centralized learning, optimizing the client-side training has a larger potential for improving the algorithm performance. To this end, it can be a promising  idea to adjust the choice of cut layer properly in order to alleviate the impact of non-IID data.
In the next section, we will empirically show how the choice of cut layer affects the algorithm performance.





\begin{figure*}
       \begin{subfigure}{0.24\textwidth}
        \centering
        \includegraphics[height=3.4cm]{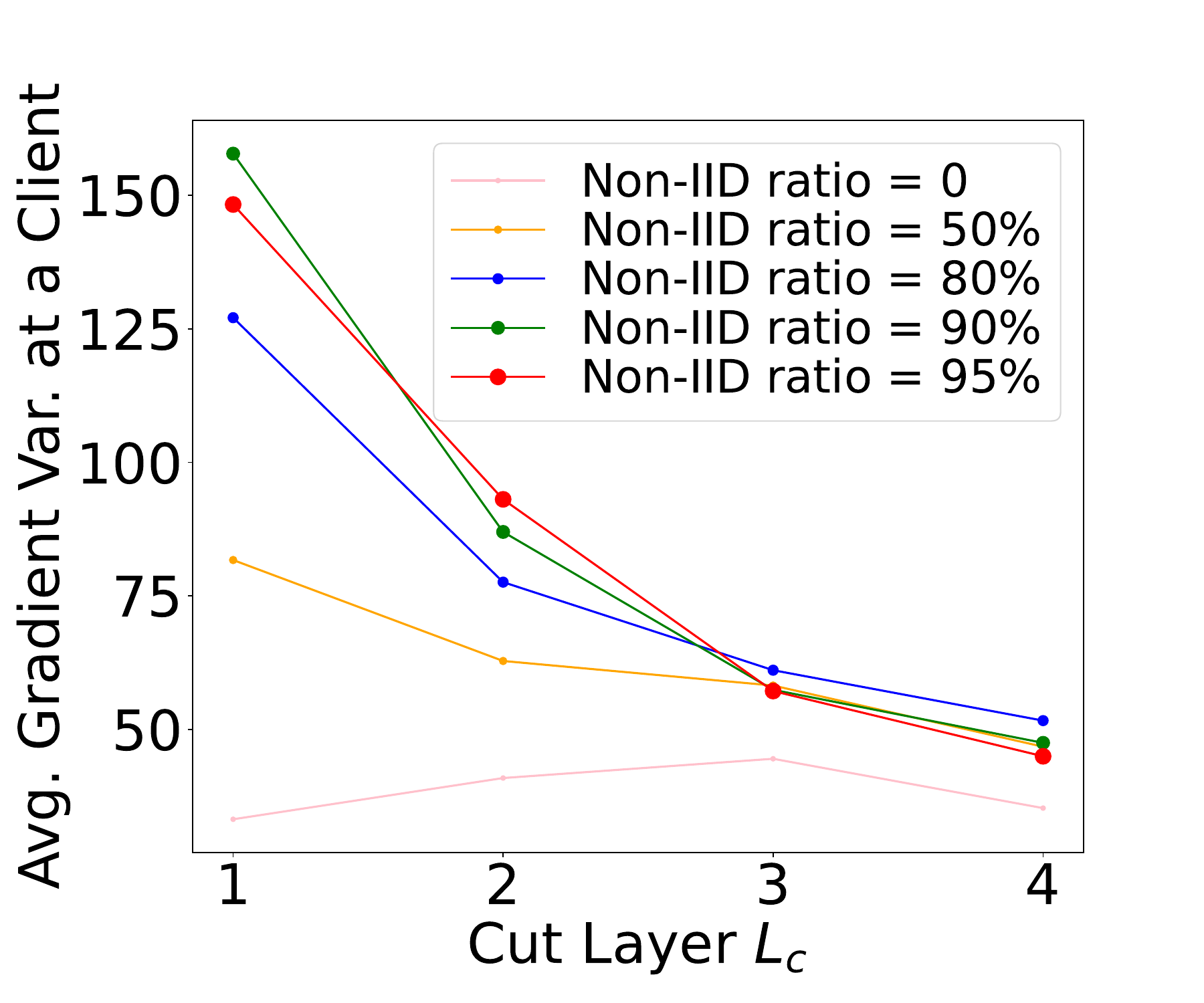}
        \caption{Gradient var. at a client.}
    \end{subfigure}
    \hfil
    \begin{subfigure}{0.24\textwidth}
        \centering
        \includegraphics[height=3.4cm]{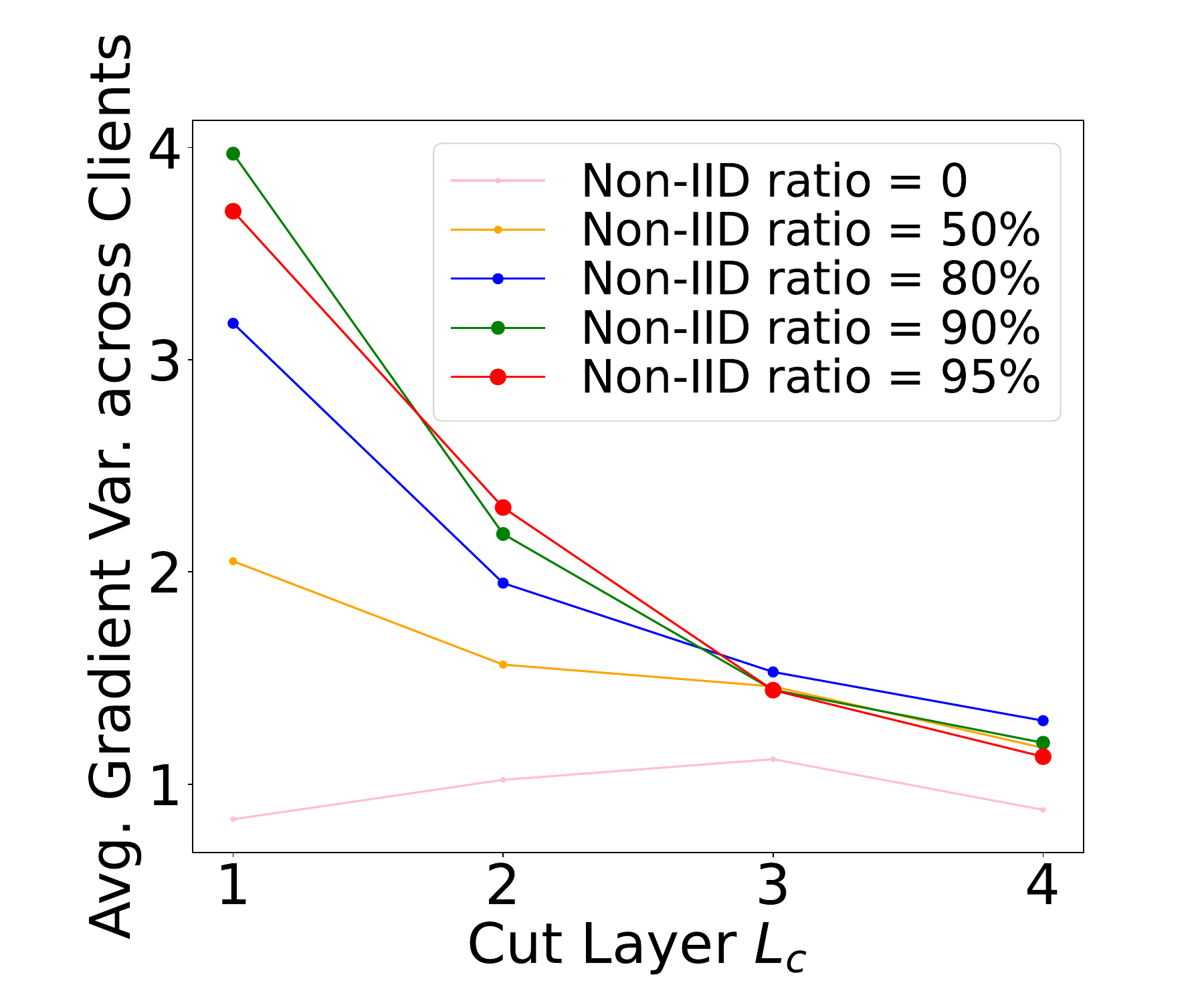}
        \caption{Gradient var.  across clients.}
    \end{subfigure}
    \hfil
    \begin{subfigure}{0.24\textwidth}
        \centering
        \includegraphics[height=3.4cm]{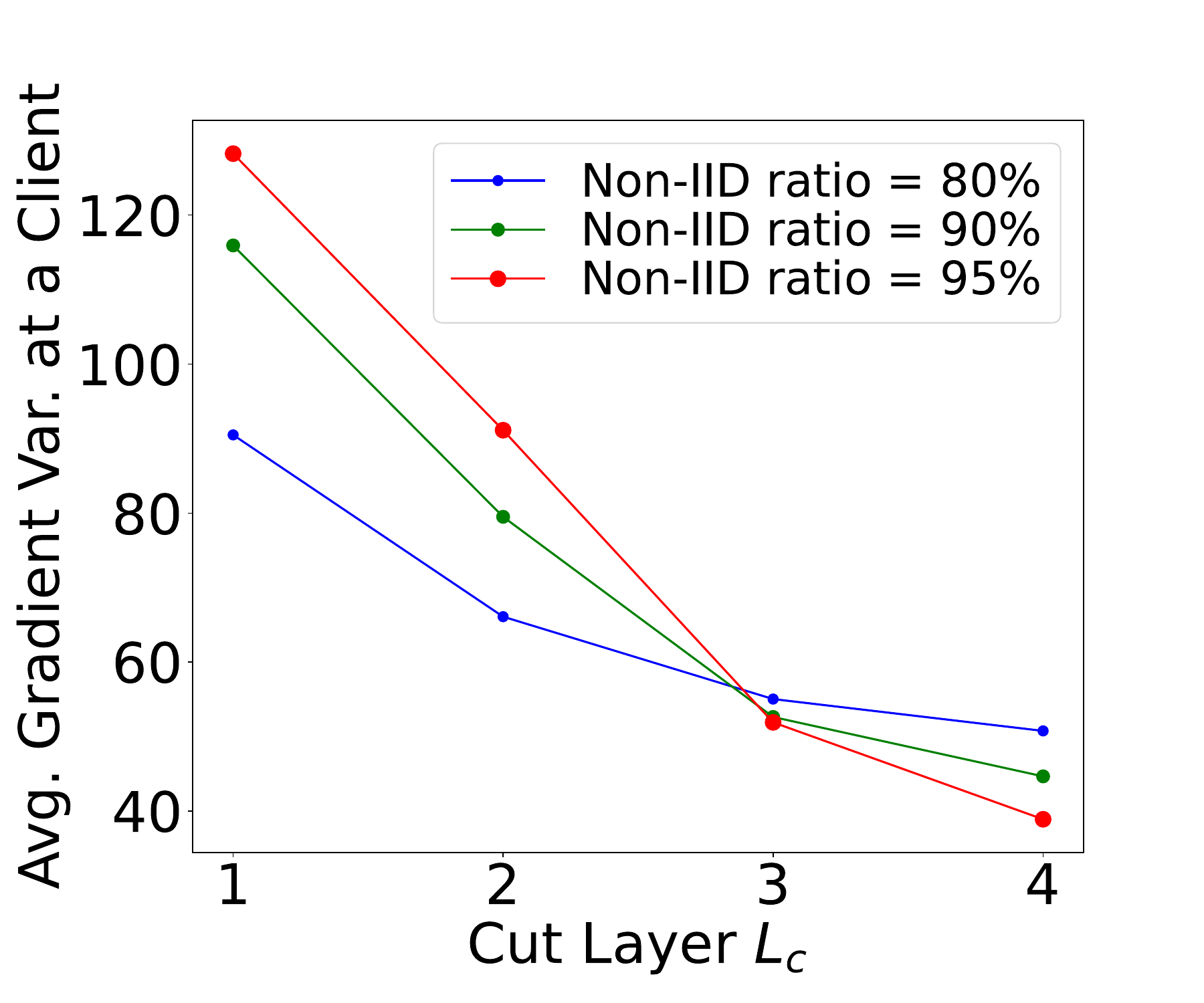}
        \caption{Gradient var. at a client.}
    \end{subfigure}
    \hfil
    \begin{subfigure}{0.24\textwidth}
        \centering
        \includegraphics[height=3.4cm]{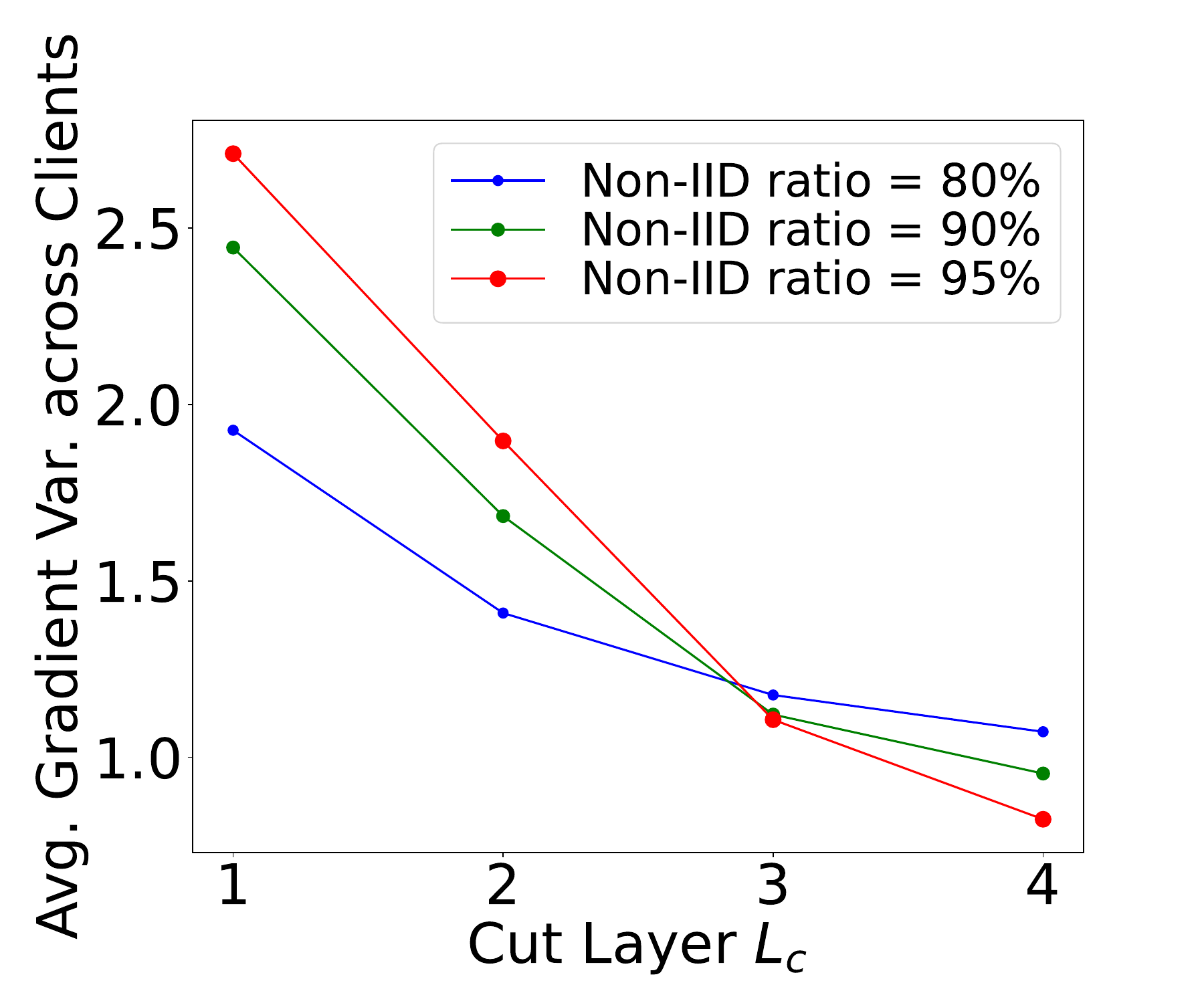}
        \caption{Gradient var. across clients.}
    \end{subfigure}
    \caption{Impact of cut layer on average gradient variance: (a-b) on CIFAR-10, (c-d) on FMNIST.}
    \label{fig:divergence}
\end{figure*}

\section{Performance Evaluation}
\subsection{Simulation Setup}
We conduct experiments on both CIFAR-10 \cite{cifar10} and Fashion-MNIST (FMNIST) \cite{xiao2017fashion}. 
To evaluate MiniBatch-SFL under different non-IID scenarios, we prepare the client data as follows. For each dataset, we first uniformly distribute $1-r\%$ of data samples to clients. Then, we sort the remaining  $r\%$ of data samples based on their labels, divide them into equal size blocks, and distribute the blocks to clients. We call $r\%$ as the \emph{non-IID ratio}, where  a larger $r\%$ implies a higher non-IID degree, and $r=0$ corresponds to the IID case. We consider $r\%\in \{0, 0.5, 0.8, 0.9, 0.95\}$ in the experiments. We use ResNet-18 as the model structure and consider four types of model splitting represented by $L_c\in \{1,2,3,4\}$, where a larger $L_c$ means more layers assigned to clients. The details of model splitting, hyper-parameters, and codes are given in the supplementary material.


\subsection{Choice of Cut Layer-Answering Q2}
We conduct experiments to answer question Q2, i.e., what is a proper choice of cut layer for MiniBatch-SFL, and why? We consider $N=10$ and report the results in Fig. \ref{fig:cut-acc-cifar}, from which we make the following observation.
\begin{observation}\label{obs: acc-cut}
The accuracy under MiniBatch-SFL generally increases in $L_c$. 
\end{observation}

Observation \ref{obs: acc-cut} is counter-intuitive. An intuitive guess to Q2 is that  a smaller $L_c$ should lead to a better performance, because it corresponds to a larger proportion of the server-side model which better mitigates the non-IID issue. However, Observation \ref{obs: acc-cut} shows that as $L_c$ increases (i.e., more layers assigned to clients), the accuracy of the trained global model increases. In addition, as the non-IID degree of the clients' datasets increases (e.g., non-IID ratios change from $0.8$ to $0.95$), the impact of cut layer becomes more significant, under which  a larger $L_c$ is more beneficial. 
\begin{figure*}
       \begin{subfigure}{0.24\textwidth}
        \centering
        \includegraphics[height=3.4cm]{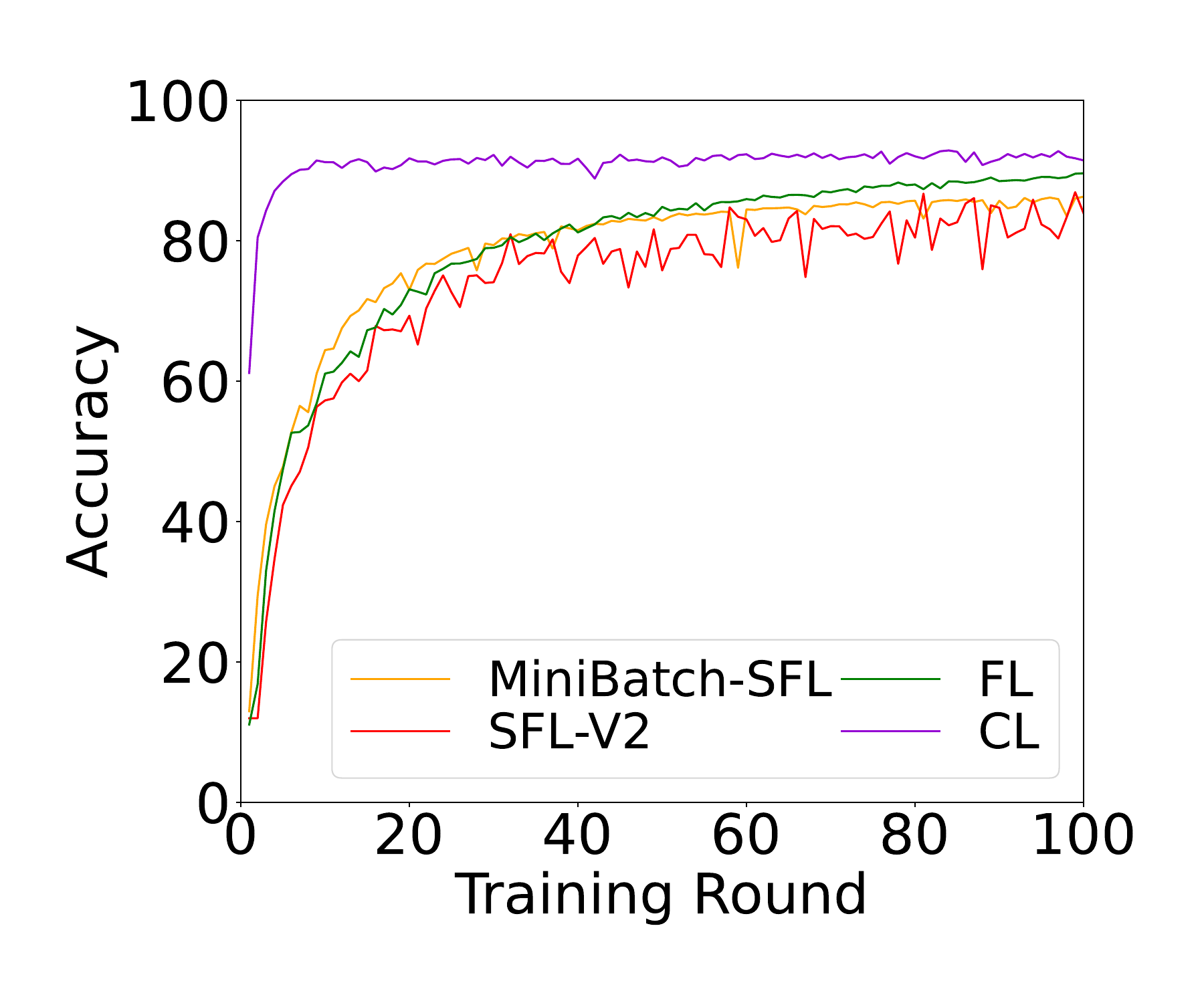}
        \caption{$r\%=0.5$.}
    \end{subfigure}
    \hfil
    \begin{subfigure}{0.24\textwidth}
        \centering
        \includegraphics[height=3.4cm]{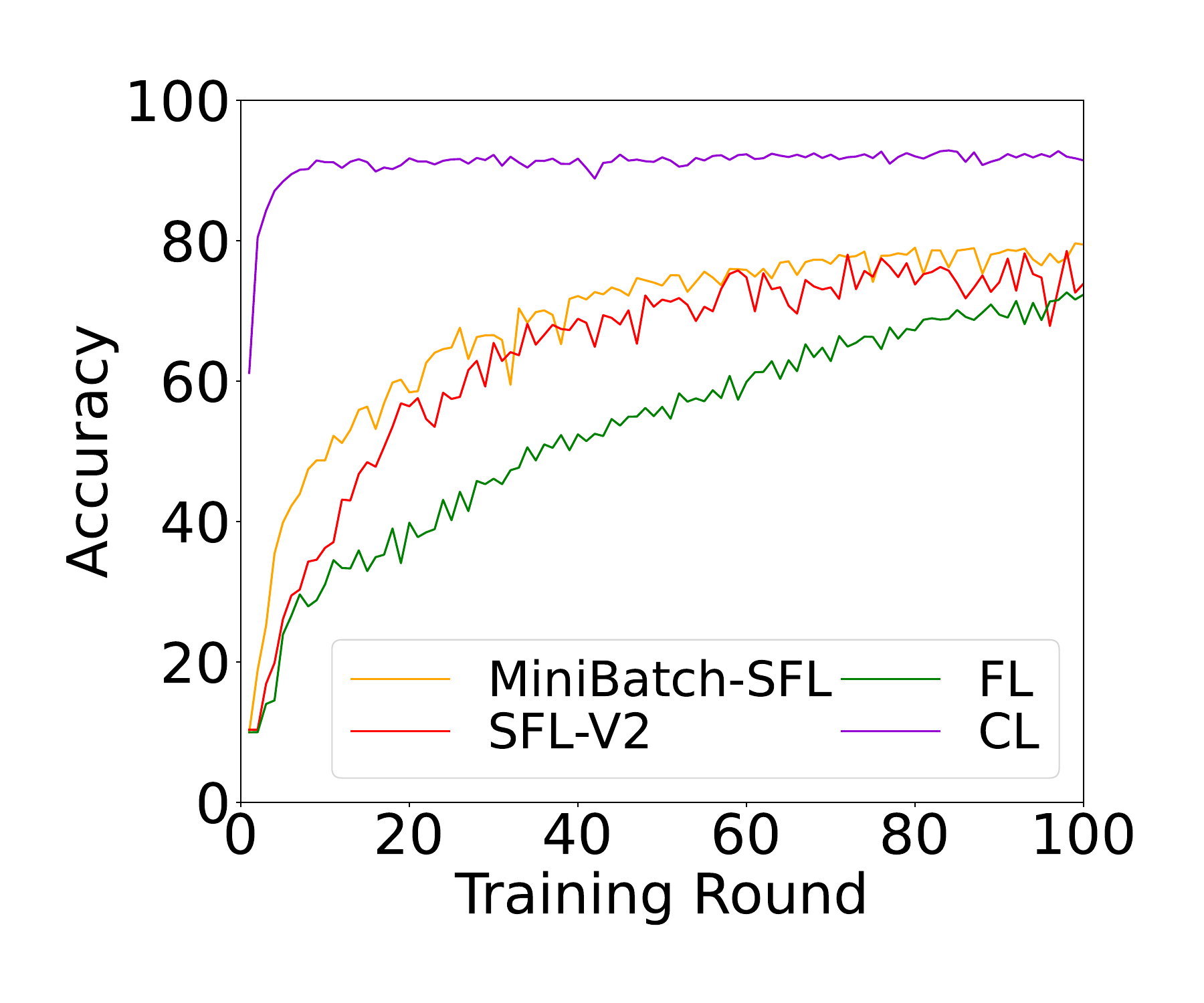}
        \caption{$r\%=0.8$.}
    \end{subfigure}
    \hfil
    \begin{subfigure}{0.24\textwidth}
        \centering
        \includegraphics[height=3.4cm]{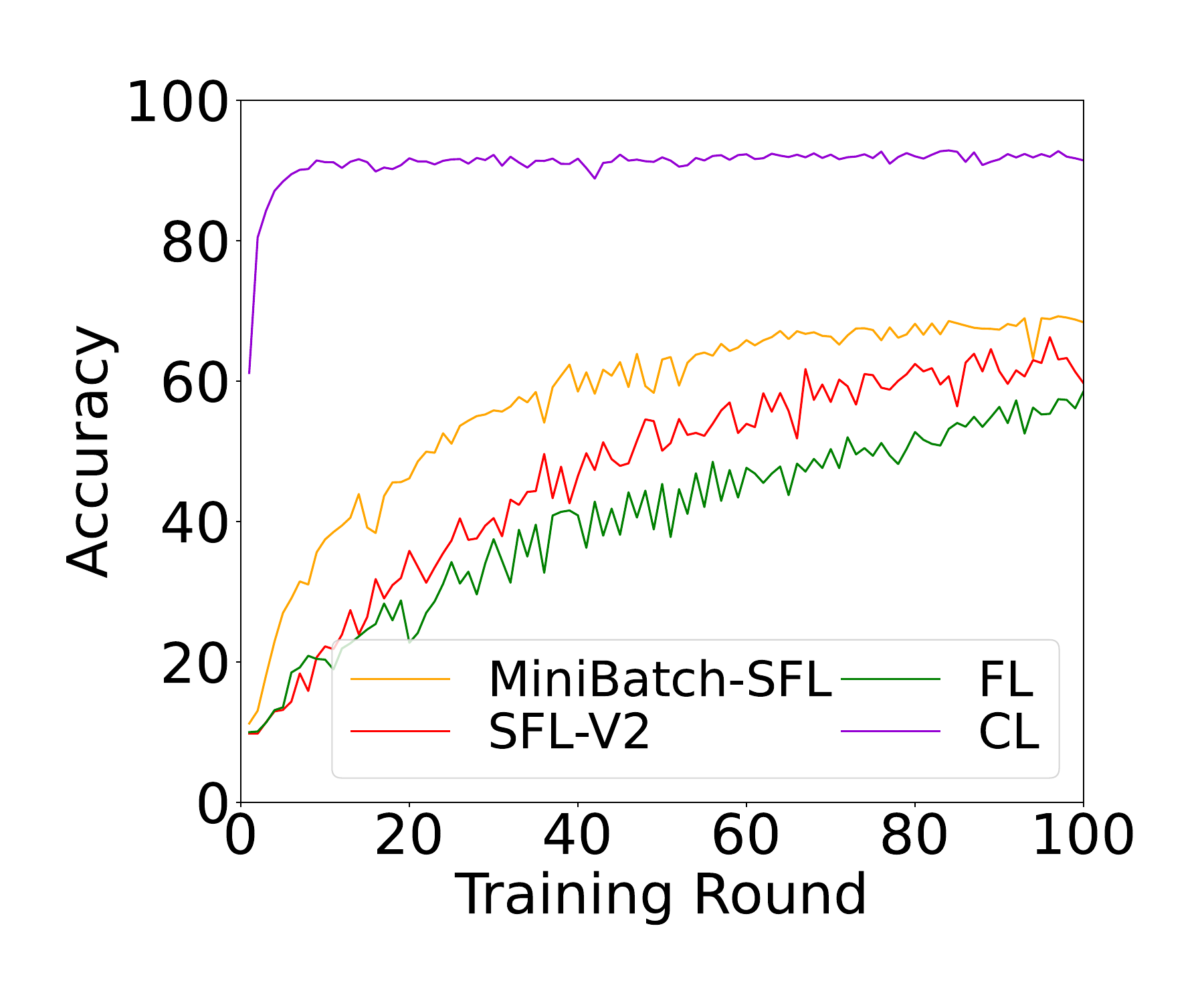}
        \caption{$r\%=0.9$.}
    \end{subfigure}
    \hfil 
    \begin{subfigure}{0.24\textwidth}
        \centering
        \includegraphics[height=3.4cm]{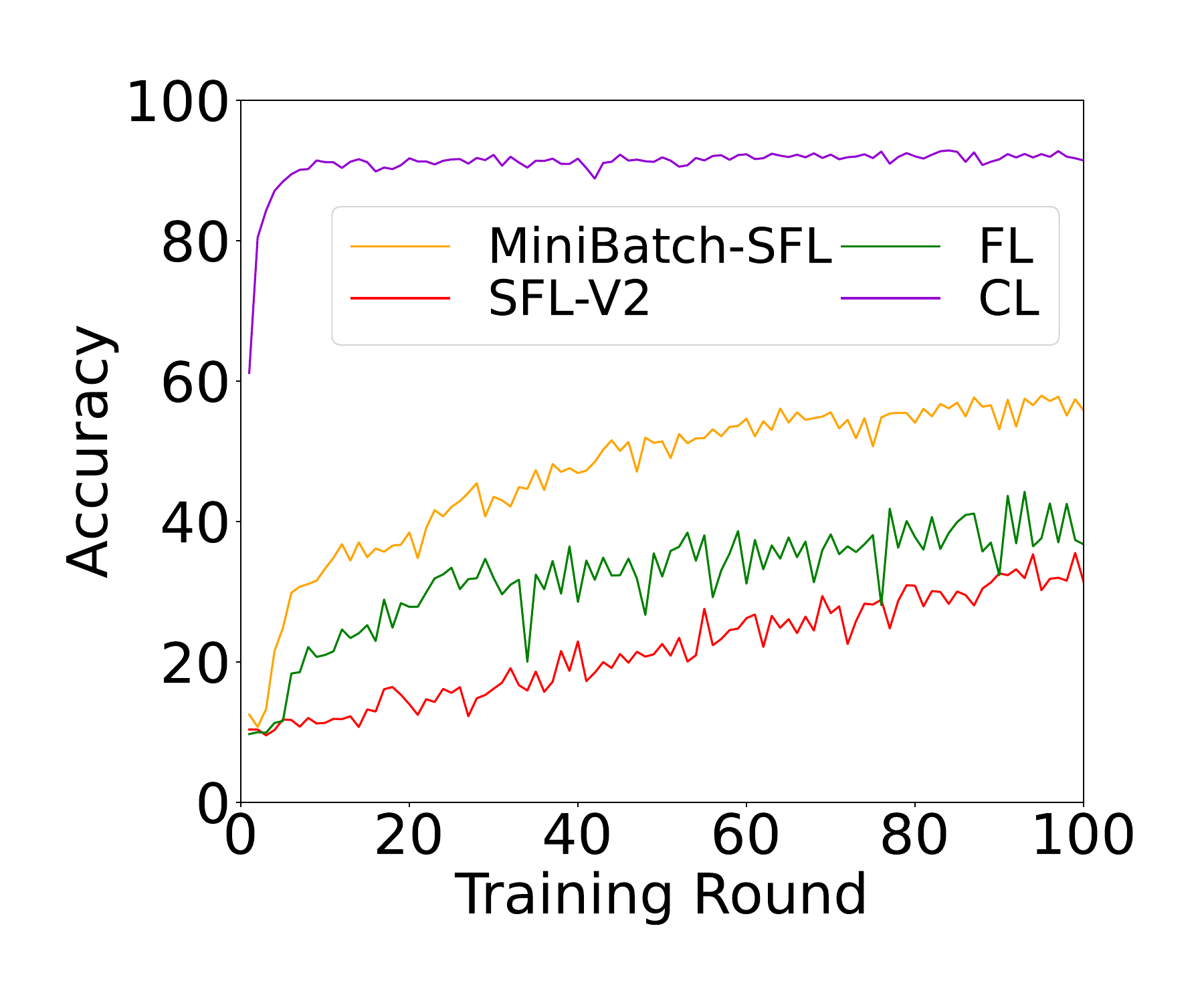}
        \caption{$r\%=0.95$.}
    \end{subfigure}
    \caption{Performance comparison with $N=10$ on CIFAR-10.}
    \label{fig:accuracy-comparison}
\end{figure*}

\begin{figure}[t]
    \centering
     \begin{subfigure}{0.23\textwidth}
        \centering
        \includegraphics[height=3.4cm]{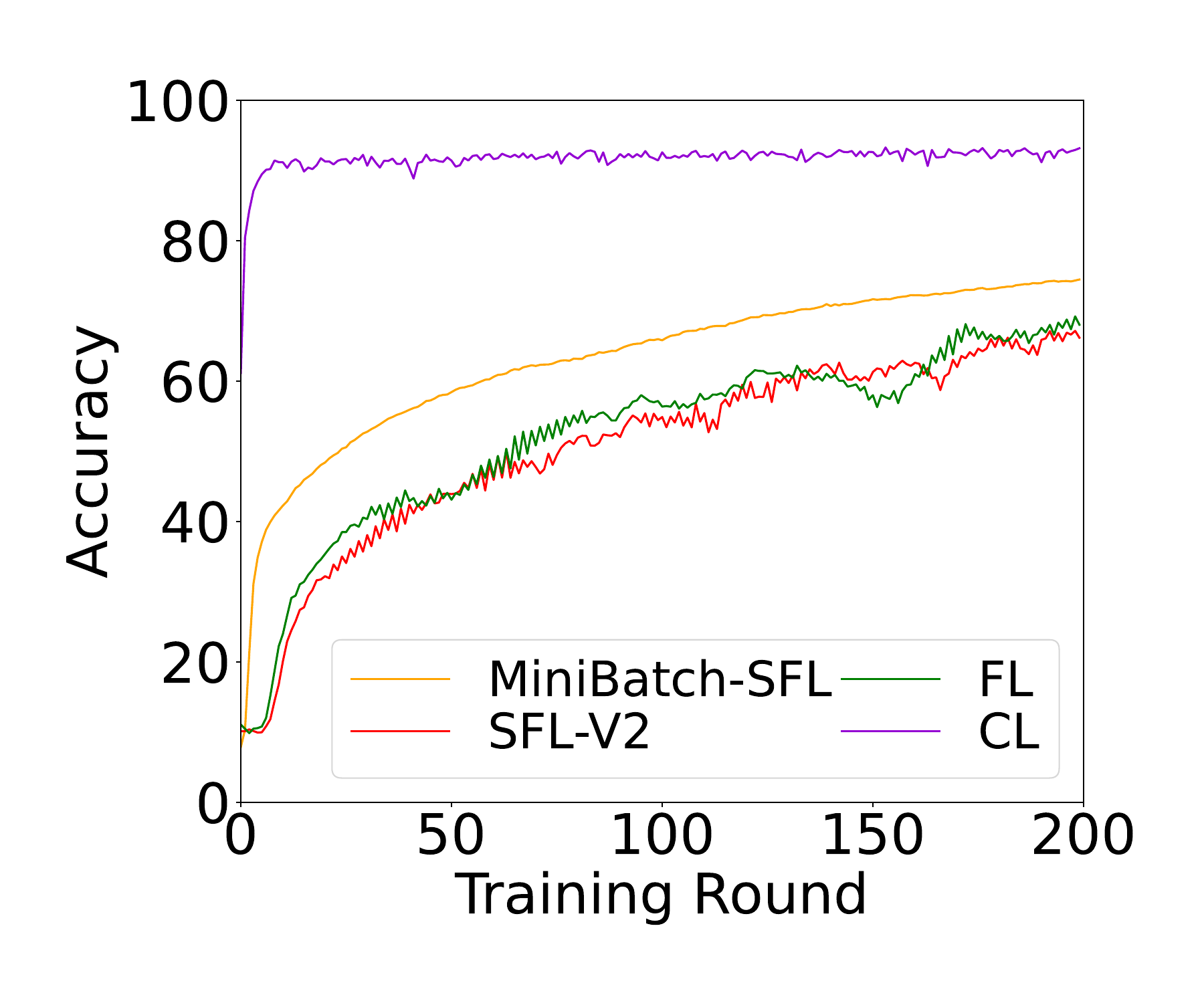}
        \caption{$r\%=0.5$.}
    \end{subfigure}
    \hfil
    \begin{subfigure}{0.23\textwidth}
        \centering
        \includegraphics[height=3.4cm]{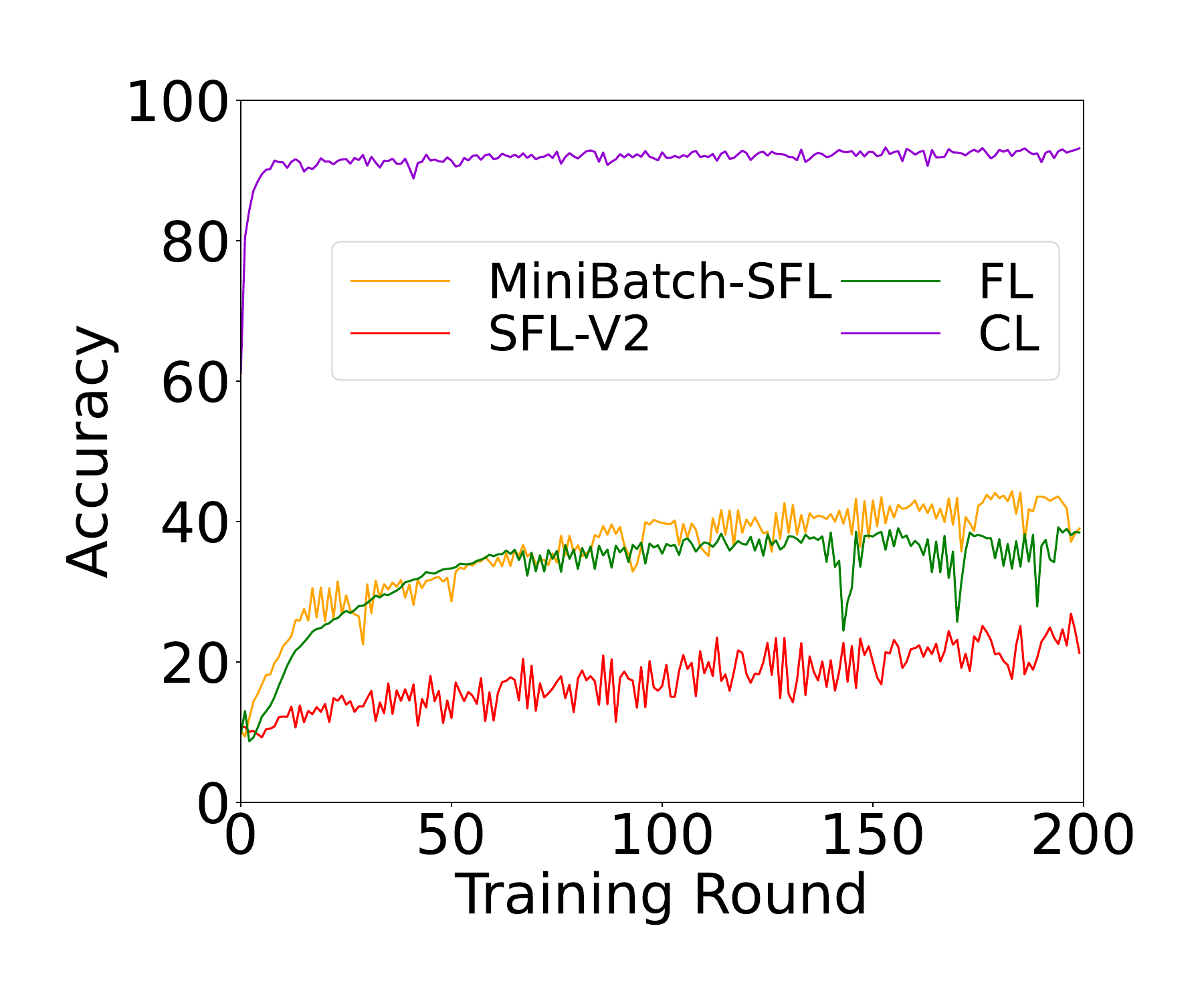}
        \caption{$r\%=0.95$.}
    \end{subfigure}
    \caption{Performance comparison with $N=100$ on CIFAR-10.}
    \label{fig:accuracy-comparison-100}
\end{figure}
 To understand the reason behind this, we consider two  metrics: (1) average gradient variance at a client, and (2) average gradient variance across clients. Specifically, the average gradient variance at a client is equal to the gradient variance of each client's client-side model divided by $L_c$.
It implies the degree of varying of a client's client-side model. The average gradient variance across clients is equal to the variance of the clients' gradients divided by $L_c$. It reveals the heterogeneity of the clients' gradients (or equivalently, their client-side models). Fig. \ref{fig:divergence} shows the impact of  $L_c$ on the above two metrics. We make the following observation.

\begin{observation}\label{obs: gradient-variance}
Both the average gradient variances at a client and across clients generally decrease in $L_c$.
\end{observation}
Observation \ref{obs: gradient-variance} implies that when $L_c$ is larger, the varying of the model of each client is smaller, the model heterogeneity among clients is also smaller. This is because a larger $L_c$ correspond to more parameters in the client-side model, which with the help from the main server,  updates less aggressively across randomly sampled mini-batches.
It is important to note that Observation \ref{obs: gradient-variance} supports the results in Fig. \ref{fig:cut-acc-cifar}. Intuitively, a larger average gradient variance at each client implies a higher chance of overfitting and forgetting at the client side, which leads to a worse algorithm performance. In addition, a larger average gradient variance across clients implies a higher heterogeneity level of the clients' local models. In this case, similar to conventional FL, clients update local models in different directions (i.e., client drift), which slows down the convergence rate. 

\subsection{Performance under Non-IID Data - Answering Q3}
Now we conduct experiments to answer question Q3, i.e., does MiniBatch-SFL have a good performance under highly non-IID data? We consider three benchmarks: 
\begin{itemize}
\item Centralized learning (CL). The main server trains a model using all clients' raw data. 
\item Federated learning (FL). The clients train the entire global model. The fed server aggregates the clients' models using FedAvg \cite{mcmahan2017communication}. 
\item SplitFed Learning (SFL) \cite{thapa2022splitfed}. In particular, we consider SFL-V2,  where the main-server sequentially updates the server-side model using clients' smashed data. We do not include SFL-V1 because it updates the global model similar to FL. The key difference is the reduced computation cost at client-side, which is not the focus of this paper. 
\end{itemize}
We first use a small cohort size $N=10$, choose $L_c=2$ for SFL-V2 and MiniBatch-SFL, and report the results in Fig. \ref{fig:accuracy-comparison}. We then use a larger cohort size $N=100$, choose $L_c=4$, and report the results in Fig. \ref{fig:accuracy-comparison-100}.

\subsubsection{Impact of Non-IID Ratio} We observe from Figs. \ref{fig:accuracy-comparison}-\ref{fig:accuracy-comparison-100}:
\begin{observation}\label{obs: impact-noniid}
The accuracies under FL, SFL-V2, and MiniBatch-SFL decrease in the non-IID ratio.
\end{observation}
Observation \ref{obs: impact-noniid} is due to the fact that in distributed learning, the client drift problem becomes severer when the non-IID ratio is higher. The performance of centralized learning (CL), however, is not affected by the non-IID ratio.

\subsubsection{Impact of Cohort Size}  We observe from Figs. \ref{fig:accuracy-comparison}-\ref{fig:accuracy-comparison-100}:
\begin{observation}\label{obs: impact-cohort}
The accuracies under FL, SFL-V2, and MiniBatch-SFL decrease in the number of clients.
\end{observation}
Under non-IID settings,  as there are more clients contributing updates, the potential for divergent updates intensifies. As a result, the global model may converge slower, leading to a decreased performance.

\subsubsection{Performance Comparison} We observe from Figs. \ref{fig:accuracy-comparison}-\ref{fig:accuracy-comparison-100}:
\begin{observation}\label{obs: comparison}
MiniBatch-SFL outperforms FL and SFL-V2, especially when the non-IID ratio is high. 
\end{observation}
MiniBatch-SFL outperforms FL mainly because the existence of the main server mitigates the model update heterogeneity among clients and hence alleviates the client drift issue. MiniBatch-SFL outperforms SFL-V2 because 
in SFL-V2, the main server updates the server-side model in a sequential manner, which may lead to the forgetting issue commonly observed in SL \cite{sheller2020federated,Gao2022evaluation}. In MiniBatch-SFL, however, the main server updates the model based on all clients' smashed data, which mitigates the forgetting issue and hence achieves a better performance. When compared with SFL-V2 and FL, the accuracy improvement can be up to 24.1\% and 17.1\% when the non-IID ratio is high (i.e., $r\%=0.95$), respectively. 


\subsection{Additional Experiments} We further include three sets of experiments in the supplementary material to validate our analysis and observations. 
\begin{enumerate}
    \item \textit{A different model structure}. The results in the main paper are based on the ResNet-18 model. We have further included the experiments using a simpler 4-layer CNN  and observed similar trends as in Figs. \ref{fig:cut-acc-cifar}-\ref{fig:divergence}. 
    \item \textit{A different metric}. The results in the main paper are based on the accuracy metric. We additionally reported the results using the loss metric and observed similar (but opposite) trends. That is, a higher accuracy is associated with a smaller loss.   
    \item \textit{Comparison on FMNIST}. The comparison among CL, FL, SFL-V2, and MiniBatch-SFL in the main paper are based on CIFAR-10. We have also included the comparison among the four algorithms on FMNIST (with both $N=10$ and $N=100$). We have made similar observations as in Observations \ref{obs: impact-noniid}-\ref{obs: comparison}.
\end{enumerate}

% ... 

\section{Conclusion}
In this paper, we propose MiniBatch-SFL, an algorithm that incorporates MiniBatch SGD into SFL to enable low computation overhead for edge devices while improving model performance on highly non-IID data. We provide a convergence analysis for MiniBatch-SFL and show that the model loss arising from the server-side model updates is independent of the data non-IIDness across clients. However, the client-side model relies on the non-IIDness, which motivates a proper choice of the cut layer to optimize the algorithm performance. Perhaps counter-intuitively, we show that a latter position of the cut layer results in a better algorithm performance. The main reason is that it leads to a smaller average gradient divergence across clients and hence mitigates the client drift problem. Our simulations on various datasets using different model structures show that the proposed MiniBatch-SFL outperforms conventional SFL and FL, and the improvement can be up to 24.1\% and 17.1\%, respectively,  when the clients' data are highly non-IID. 

For the future work, there are many important and interesting challenges to be addressed. First, it would be interesting to investigate whether a tighter convergence bound exists, and if so, how it compares to that of other distributed learning algorithms. Second, it is important to study the tradeoff between accuracy and overhead (in terms of communication/computation) for MiniBatch-SFL. Third, it is also interesting to develop customized privacy-preserving methods for MiniBatch-SFL, e.g., adding noise to the transmitted smashed data. 
\bibliography{aaai24.bib}
\bibliographystyle{IEEEtran}

\clearpage 
\onecolumn

\section{Proof of Proposition 1}
\begin{proof}
We aim to bound $\mathbb{E}||\boldsymbol{w}_s^{TEM}-\boldsymbol{w}_s^*||^2$. For ease of presentation, we use $\eta_s$ instead of $\eta_s^i$ for now, and we have 
   \begin{equation}\label{bounding-SGD}
   \begin{aligned}
   &\mathbb{E}\left[||\boldsymbol{w}_s^i-\boldsymbol{w}_s^*||^2\right]\\ &= \mathbb{E}\left[\bigg|\bigg| \boldsymbol{w}_s^{i-1}-\eta_s \frac{1}{N}\sum_{n \in \mathcal{N}}\nabla_{\boldsymbol{w}_s}f_n(\boldsymbol{w}^{i-1}; \boldsymbol{\zeta}_n^{i-1})-\boldsymbol{w}_s^*\bigg|\bigg|^2\right]\\
   &=\mathbb{E}\left[||\boldsymbol{w}_s^{i-1}-\boldsymbol{w}_s^*||^2 \right]- 2\eta_s \mathbb{E}\langle \nabla_{\boldsymbol{w}_s} F(\boldsymbol{w}^{i-1}), \boldsymbol{w}_s^{i-1}-\boldsymbol{w}_s^*\rangle + \eta_s^2 \mathbb{E}\left[\bigg|\bigg|\frac{1}{N}\sum_{n \in \mathcal{N}}\nabla_{\boldsymbol{w}_s}f_n(\boldsymbol{w}^{i-1}; \zeta_n^{i-1})\bigg|\bigg|^2\right]\\
   &\le (1-\mu\eta_s)\mathbb{E}\left[||\boldsymbol{w}_s^{i-1}-\boldsymbol{w}_s^*||^2 \right]-2\eta_s \mathbb{E}\left[\underbrace{F(\boldsymbol{w}^{i-1})}_{=F(\boldsymbol{w}_c^{i-1}, \boldsymbol{w}_s^{i-1})}-F(\boldsymbol{w}_c^{i-1}, \boldsymbol{w}_s^*))\right]
+\eta_s^2 \underbrace{\mathbb{E}\left[\bigg|\bigg|\frac{1}{N}\sum_{n \in \mathcal{N}}\nabla_{\boldsymbol{w}_s}f_n(\boldsymbol{w}^{i-1}; \zeta_n^{i-1})\bigg|\bigg|^2\right]}_{C_1},
   \end{aligned}
   \end{equation}
   where the inequality is due to the $\mu$-strong convexity of $F$.

   Next, we bound $C_1$ as follows:
   \begin{equation}\label{bounding-C1}
   \begin{aligned}
  C_1&= \mathbb{E}\bigg|\bigg|\frac{1}{N}\sum_{n \in \mathcal{N}}\nabla_{\boldsymbol{w}_s}f_n(\boldsymbol{w}^{i-1}; \zeta_n^{i-1})\bigg|\bigg|^2\\
   &=\mathbb{E}\bigg|\bigg|\frac{1}{N}\sum_{n \in \mathcal{N}}\left[\nabla_{\boldsymbol{w}_s}f_n(\boldsymbol{w}^{i-1}; \zeta_n^{i-1})-\nabla_{\boldsymbol{w}_s}f_n(\boldsymbol{w}^*; \zeta_n^{i-1})+\nabla_{\boldsymbol{w}_s}f_n(\boldsymbol{w}^{*}; \zeta_n^{i-1})\right]\bigg|\bigg|^2\\
   &\le2 \mathbb{E}\bigg|\bigg|\frac{1}{N}\sum_{n \in \mathcal{N}}\left[\nabla_{\boldsymbol{w}_s}f_n(\boldsymbol{w}^{i-1}; \zeta_n^{i-1})-\nabla_{\boldsymbol{w}_s}f_n(\boldsymbol{w}^*; \zeta_n^{i-1})\right]\bigg|\bigg|^2 + 2\mathbb{E}\bigg|\bigg|\frac{1}{N}\sum_{n \in \mathcal{N}}\left[\nabla_{\boldsymbol{w}_s}f_n(\boldsymbol{w}^{*}; \zeta_n^{i-1})\right]\bigg|\bigg|^2\\
   &\le \frac{2}{N} \sum_{n\in \mathcal{N}}\mathbb{E}\bigg|\bigg| \nabla_{\boldsymbol{w}_s}f_n(\boldsymbol{w}^{i-1}; \zeta_n^{i-1})-\nabla_{\boldsymbol{w}_s}f_n(\boldsymbol{w}^*; \zeta_n^{i-1})\bigg|\bigg|^2+  2\mathbb{E}\bigg|\bigg|\frac{1}{N}\sum_{n \in \mathcal{N}}\left[\nabla_{\boldsymbol{w}_s}f_n(\boldsymbol{w}^{*}; \zeta_n^{i-1})\right]\bigg|\bigg|^2\\
   &\le \frac{2}{N} \sum_{n\in \mathcal{N}}\mathbb{E}\bigg|\bigg| \nabla_{\boldsymbol{w}_s}f_n(\boldsymbol{w}^{i-1}; \zeta_n^{i-1})-\nabla_{\boldsymbol{w}_s}f_n(\boldsymbol{w}^*; \zeta_n^{i-1})\bigg|\bigg|^2 + \frac{2R^2}{N^2}\\
   &\le \frac{4S}{N}\sum_{n\in \mathcal{N}}\mathbb{E}\left[f_n(\boldsymbol{w}^{i-1};\zeta_{n}^{i-1})-f_n(\boldsymbol{w}^{*};\zeta_{n}^{i-1})-\langle \nabla_{\boldsymbol{w}_s}f_n(\boldsymbol{w}^*; \zeta_{n}^{i-1}), \boldsymbol{w}^{i-1}-\boldsymbol{w}^*\rangle \right]+ \frac{2R^2}{N^2}\\
   &=4S \mathbb{E}\left[F(\boldsymbol{w}_c^{i-1}, \boldsymbol{w}_s^{i-1})- F(\boldsymbol{w}_c^{i-1}, \boldsymbol{w}_s^{*})\right]+\frac{2R^2}{N^2}
   \end{aligned}
   \end{equation}
   where 
   \begin{itemize}
   \item First inequality: $(x-y+y)^2\le 2(x-y)^2+2y^2$.
   \item Second inequality: $\big|\big|\sum_{i=1}^{N}a_i\big|\big|^2\le N \sum_{i=1}^{N} ||a_i||^2$
   \item Third inequality: Assumption \ref{asm:bound-grad-norm} and that $\mathbb{E}\bigg|\bigg|\sum_{n \in \mathcal{N}}\left[\nabla_{\boldsymbol{w}_s}f_n(\boldsymbol{w}^{*}; \zeta_n^{i-1})\right]\bigg|\bigg|^2 \le \mathbb{E}\bigg|\bigg|\sum_{n \in \mathcal{N}}\left[\nabla_{\boldsymbol{w}}f_n(\boldsymbol{w}^{*}; \zeta_n^{i-1})\right]\bigg|\bigg|^2$
   \item Fourth inequality: Lemma 1 from \cite{woodworth2020minibatch} (Co-Coercivity of the gradient for $S$-smooth and convex functions)
   \end{itemize}
   Plugging (\ref{bounding-C1}) back to (\ref{bounding-SGD}), and if we choose $\eta_s\le \frac{1}{4S}$, we have 
   \begin{equation}\label{15}
   \begin{aligned}
   \mathbb{E}\left[||\boldsymbol{w}_s^i-\boldsymbol{w}_s^*||^2\right] &\le (1-\mu\eta_s)\mathbb{E}\left[||\boldsymbol{w}_s^{i-1}-\boldsymbol{w}_s^*||^2 \right]-2\eta_s\mathbb{E}\left[F(\boldsymbol{w}_c^{i-1}, \boldsymbol{w}_s^{i-1})- F(\boldsymbol{w}_c^{i-1}, \boldsymbol{w}_s^{*})\right]\\
   &\hspace{10mm}+ \eta_s^2\left(4L \mathbb{E}\left[F(\boldsymbol{w}_c^{i-1}, \boldsymbol{w}_s^{i-1})- F(\boldsymbol{w}_c^{i-1}, \boldsymbol{w}_s^{*})\right]+\frac{2R^2}{N^2}\right)\\
   &=(1-\mu\eta_s)\mathbb{E}\left[||\boldsymbol{w}_s^{i-1}-\boldsymbol{w}_s^*||^2 \right] -2\eta_s(1-2L\eta_s)\mathbb{E}\left[F(\boldsymbol{w}_c^{i-1}, \boldsymbol{w}_s^{i-1})- F(\boldsymbol{w}_c^{i-1}, \boldsymbol{w}_s^{*})\right]+\frac{2\eta_s^2R^2}{N^2}\\
   &\le (1-\mu\eta_s)\mathbb{E}\left[||\boldsymbol{w}_s^{i-1}-\boldsymbol{w}_s^*||^2 \right] -\eta_s\mathbb{E}\left[F(\boldsymbol{w}_c^{i-1}, \boldsymbol{w}_s^{i-1})- F(\boldsymbol{w}_c^{i-1}, \boldsymbol{w}_s^{*})\right]+\frac{2\eta_s^2R^2}{N^2}.
   \end{aligned}
   \end{equation}
   Let $\Delta^{i+1}\triangleq \mathbb{E}\left[||\boldsymbol{w}_s^{i+1}-\boldsymbol{w}_s^*||^2\right]$. We can rewrite (\ref{15}) as:
\begin{equation}
   \begin{aligned}
   \Delta^{i+1}
   &\le (1-\mu\eta_s)\Delta^{i} -\eta_s\mathbb{E}\left[F(\boldsymbol{w}_c^{i}, \boldsymbol{w}_s^{i})- F(\boldsymbol{w}_c^{i}, \boldsymbol{w}_s^{*})\right]+\frac{2\eta_s^2R^2}{N^2},\\
   & \le (1-\mu\eta_s)\Delta^{i} +\frac{2\eta_s^2R^2}{N^2}.
   \end{aligned}
   \end{equation}
   Consider a diminishing stepsize $\eta_s=\frac{\beta}{\gamma+i}$ where $\beta=\frac{2}{\mu}$,  $\gamma=\max\left\{\frac{8S}{\mu}-1, EM\right\}$. It is easy to show that $\eta_s\le \frac{1}{4S}$ for all $i$. Next, we will prove that $\Delta^{i+1}\le \frac{v}{\gamma +i+1}$, where $v=\max\left\{\frac{4B}{\mu^2}, (\gamma+1)\Delta^0\right\}$, $B=\frac{2R^2}{N^2}$. We prove this by induction. 

   First,  the definition of $v$ ensures that it holds for $i=-1$. Assume the conclusion holds for some $i$, it follows that 
   \begin{equation}\label{induction-1}
   \begin{aligned}
   \Delta^{i+1}&\le (1-\mu\eta_s)\Delta^i +\eta_s^2B\\
   &\le\left(1-\frac{\mu \beta}{\gamma+i}\right)\frac{v}{\gamma+i}+\eta_s^2B\\
   &=\frac{\gamma+i-1}{(\gamma+i)^2}v + \left[\frac{\beta^2 B}{(\gamma+i)^2}-\frac{\beta \mu-1}{(\gamma+i)^2}v\right]\\
   &=\frac{\gamma+i-1}{(\gamma+i)^2}v + \left[\frac{\beta^2 B}{(\gamma+i)^2}-\frac{\beta \mu-1}{(\gamma+i)^2} \max\left\{\frac{4B}{\mu^2}, (\gamma+1)\Delta^0\right\} \right]\\
   &= \frac{\gamma+i-1}{(\gamma+i)^2}v + \left[\frac{\beta^2 B}{(\gamma+i)^2}-\frac{\beta \mu-1}{(\gamma+i)^2} \max\left\{\frac{\beta^2 B}{\beta\mu -1}, (\gamma+1)\Delta^0\right\} \right]\\
   &\le \frac{\gamma+i-1}{(\gamma+i)^2}v\\
   &\le \frac{v}{\gamma+i+1}.
   \end{aligned}
   \end{equation}
   Hence, we have proven that $\Delta^i\le \frac{v}{\gamma +i}, \forall i$. Therefore, we have 
   \begin{equation}\label{induction-2}
   \mathbb{E}\left[||\boldsymbol{w}_s^i-\boldsymbol{w}_s^*||^2\right]= \Delta^{i}\le \frac{v}{\gamma+i}=\frac{\max\left\{\frac{8R^2}{\mu^2N^2},  (\gamma+1)\mathbb{E}\left[||\boldsymbol{w}_s^0-\boldsymbol{w}_s^*||^2\right]\right\}}{\gamma+i}. 
   \end{equation}
   Now let $i=TEM$, we have 
   \begin{equation}\label{induction-3}
   \begin{aligned}
   \mathbb{E}\left[||\boldsymbol{w}_s^{TEM}-\boldsymbol{w}_s^*||^2\right]&\le \frac{\max\left\{\frac{8R^2}{\mu^2N^2},  (\gamma+1)\mathbb{E}\left[||\boldsymbol{w}_s^0-\boldsymbol{w}_s^*||^2\right]\right\}}{\gamma+TEM}\\
  & \le \frac{8R^2}{\mu^2N^2(\gamma+TEM)}+\frac{  (\gamma+1)\mathbb{E}\left[||\boldsymbol{w}_s^0-\boldsymbol{w}_s^*||^2\right]}{\gamma+TEM}\\
  &=\frac{8R^2+ \mu^2N^2(\gamma+1)||\boldsymbol{w}_s^0-\boldsymbol{w}_s^*||^2}{\mu^2N^2(\gamma+TEM)}.
   \end{aligned}
   \end{equation}
   Thu we finish the proof.
\end{proof}

\newpage 
\section{Proof of Proposition 2}
The proof of Proposition 2 mainly follows the proof of Theorem 1 in \cite{li2019convergence}. We provide the details below to make the paper self-contained. 

\subsection{Preliminaries}
We recall/define the following notations for presentation convenience:
\begin{itemize}
\item $\boldsymbol{w}_{c,n}^{i}$, $\boldsymbol{w}_s^{i}$: client $n$'s model and server side model actually maintained after the $i^{\rm th}$ SGD step
\item $\boldsymbol{v}_{c,n}^i$: the immediate result of client $n$'s model updated from $\boldsymbol{w}_{c,n}^{i-1}$ after the $i^{\rm th}$ SGD step
\item $\bar{\boldsymbol{w}}_c^i\triangleq \sum_{n \in \mathcal{N}} p_n \boldsymbol{w}_{c,n}^i$; $\bar{\boldsymbol{v}}_c^i\triangleq \sum_{n \in \mathcal{N}} p_n \boldsymbol{v}_{c,n}^i$: averaged values of $\boldsymbol{w}_{c,n}^{i}$ and $\boldsymbol{v}_{c,n}^i$ for all $n \in \mathcal{N}$ 
\item $\boldsymbol{g}_c^i(\boldsymbol{\zeta}_c^i)=\sum_{n\in \mathcal{N}}p_n \nabla f_n(\boldsymbol{v}^i_{c,n}, \boldsymbol{w}_s^{i}; \zeta_{c,n}^i)$, where $\boldsymbol{\zeta}_c^i\triangleq \{\zeta_{c,n}^i\}_{n \in \mathcal{N}} $;  $\bar{\boldsymbol{g}}_c^i=\sum_{n\in \mathcal{N}}p_n \nabla f_n(\boldsymbol{v}^i_{c,n}, \boldsymbol{w}_s^{i})$. It is easy to show that $\mathbb{E}_{\boldsymbol{\zeta}_c^i}\left[\boldsymbol{g}_c^i(\boldsymbol{\zeta}_c^i)\right]=\bar{\boldsymbol{g}}_c^i$, and $\bar{\boldsymbol{v}}_c^{i+1}=\bar{\boldsymbol{w}}_c^i - \eta_c \boldsymbol{g}_c^i(\boldsymbol{\zeta}_c^i)$
\end{itemize}
We will also use $\eta_c$ instead of $\eta_c^i$ for now. 
We first present some useful lemmas, and then use them to prove Proposition 2. 
\begin{lemma}{(Bound of SGD step)}\label{lem: bound-sgd}
Under Assumptions \ref{asm:lipschitz_grad} and \ref{asm:strong-convexity}, if $\eta_c\le \frac{1}{4S}$, then for any $i$, we have 
\begin{equation}\label{39-1}
\mathbb{E}\left[||\bar{\boldsymbol{v}}_c^{i+1}-\boldsymbol{w}_c^*||^2\right]\le (1-\mu\eta_c)\mathbb{E}\left[||\bar{\boldsymbol{w}}_c^i-\boldsymbol{w}_c^*||^2\right] +2\mathbb{E}\left[\sum_{n \in \mathcal{N}}p_n||\bar{\boldsymbol{w}}_c^i- \boldsymbol{w}_{c,n}^i||^2\right]+6S\eta_c^2\Gamma+\eta_c^2\mathbb{E}\left[||\bar{\boldsymbol{g}}_c^i-\boldsymbol{g}_c^i(\boldsymbol{\zeta}_c^i)||^2\right],
\end{equation}
where $\Gamma\triangleq  F(\boldsymbol{w}^*)-\sum_{n \in \mathcal{N}}p_n f_n(\boldsymbol{w}_{c,n}^*, \boldsymbol{w}_s)$.
\end{lemma}
\begin{proof}
The proof is given after the proof of Proposition 2. 
\end{proof}

\begin{lemma}{(Bounding the variance)}\label{lem: bound-variance}
Under Assumption \ref{asm:bounded-grad-variance}, we have 
\begin{equation}
\mathbb{E}\left[||\bar{\boldsymbol{g}}_c^i-\boldsymbol{g}_c^i(\boldsymbol{\zeta}_c^i)||^2\right]\le \sum_{n\in \mathcal{N}}p_n^2\sigma_n^2.
\end{equation}
\end{lemma}
\begin{proof}
The proof follows immediately from Assumption \ref{asm:bounded-grad-variance}.
\end{proof}

\begin{lemma}{(Bounding the divergence of client-side parameter)}\label{lem: model-variance}
Under Assumptions \ref{asm:lipschitz_grad}, \ref{asm:strong-convexity}, \ref{asm:bound-grad-norm}, and \ref{asm:unbiased_global}, then for a decreasing step size $\eta_c=\frac{2}{\mu(\gamma+i)}$ for any $\gamma>0$, we have 
\begin{equation}
\sum_{n=1}^{N}p_n\mathbb{E}\left[||\boldsymbol{w}^{i}_{c,n}-\bar{\boldsymbol{w}}_c^i||^2\right]\le 3EMR^2\eta_c^2+6E^2M^2\eta_c^2\delta^2.
\end{equation}
\end{lemma}
\begin{proof}
The proof follows immediately from the proof of Lemma 8 in \cite{woodworth2020minibatch}.
\end{proof}

\subsection{Proof of Proposition 2}
\begin{proof}
Let $\Delta_c^{i+1}=\mathbb{E}\left[||\boldsymbol{w}_c^{i+1}-\boldsymbol{w}_c^*||^2\right]$. Based on Lemmas 1-3, we can rewrite (\ref{39-1}) as:
\begin{equation}
   \begin{aligned}
   \Delta_c^{i+1}\le (1-\mu\eta_c)\Delta_c^{i} +6EMR^2\eta_c^2+12E^2M^2\eta_c^2\delta^2 +6S\eta_c^2\Gamma+  \eta_c^2\sum_{n\in \mathcal{N}}p_n^2\sigma_n^2
   \end{aligned}
   \end{equation}
   Consider a diminishing stepsize $\eta_c=\frac{\beta}{\gamma+i}$ where $\beta=\frac{2}{\mu}$,  $\gamma=\max\left\{\frac{8S}{\mu}-1, EM\right\}$. It is easy to show that $\eta_c\le \frac{1}{4S}$ for all $i$.  
   Next, we will prove that $\Delta^{i+1}\le \frac{v}{\gamma +i+1}$, where $v=\max\left\{\frac{4H}{\mu^2}, (\gamma+1)\Delta^0\right\}$, $H\triangleq 6EMR^2+12E^2M^2\delta^2   +6S\Gamma+  \sum_{n\in \mathcal{N}}p_n^2\sigma_n^2$. This is proved by induction below.

    First,  the definition of $v$ ensures that it holds for $i=-1$. Assume the conclusion holds for some $i$, it follows that 
   \begin{equation}\label{induction-1}
   \begin{aligned}
   \Delta^{i+1}&\le (1-\mu\eta_s)\Delta^i +\eta_c^2H\\
   &\le\left(1-\frac{\mu \beta}{\gamma+i}\right)\frac{v}{\gamma+i}+\eta_c^2H\\
   &=\frac{\gamma+i-1}{(\gamma+i)^2}v + \left[\frac{\beta^2 H}{(\gamma+i)^2}-\frac{\beta \mu-1}{(\gamma+i)^2}v\right]\\
   &=\frac{\gamma+i-1}{(\gamma+i)^2}v + \left[\frac{\beta^2 H}{(\gamma+i)^2}-\frac{\beta \mu-1}{(\gamma+i)^2} \max\left\{\frac{4H}{\mu^2}, (\gamma+1)\Delta^0\right\} \right]\\
   &= \frac{\gamma+i-1}{(\gamma+i)^2}v + \left[\frac{\beta^2 H}{(\gamma+i)^2}-\frac{\beta \mu-1}{(\gamma+i)^2} \max\left\{\frac{\beta^2 H}{\beta\mu -1}, (\gamma+1)\Delta^0\right\} \right]\\
   &\le \frac{\gamma+i-1}{(\gamma+i)^2}v\\
   &\le \frac{v}{\gamma+i+1}.
   \end{aligned}
   \end{equation}
Therefore, we have 
   \begin{equation}
   \mathbb{E}\left[||\bar{\boldsymbol{w}}_c^i-\boldsymbol{w}_c^*||^2\right]= \Delta_c^{i}\le \frac{v}{\gamma+i}=\frac{\max\left\{\frac{4H}{\mu^2},  (\gamma+1)\mathbb{E}\left[||\bar{\boldsymbol{w}}_c^0-\boldsymbol{w}_c^*||^2\right]\right\}}{\gamma+i}. 
   \end{equation}
   Now let $i=TEM$, we have 
   \begin{equation}
   \begin{aligned}
   \mathbb{E}\left[||\bar{\boldsymbol{w}}_c^{TEM}-\boldsymbol{w}_c^*||^2\right]&\le \frac{\max\left\{\frac{4H}{\mu^2},  (\gamma+1)\mathbb{E}\left[||\bar{\boldsymbol{w}}_c^0-\boldsymbol{w}_c^*||^2\right]\right\}}{\gamma+TEM}\\
  & \le \frac{4H}{\mu^2(\gamma+TEM)}+\frac{ (\gamma+1)\mathbb{E}\left[||\bar{\boldsymbol{w}}_c^0-\boldsymbol{w}_c^*||^2\right]}{\gamma+TEM}\\
&=\frac{4H+\mu^2(\gamma+1)\mathbb{E}\left[||\bar{\boldsymbol{w}}_c^0-\boldsymbol{w}_c^*||^2\right]}{\mu^2(\gamma+TEM)}\\
  &=\frac{4H+\mu^2(\gamma+1)\mathbb{E}\left[||{\boldsymbol{w}}_c^0-\boldsymbol{w}_c^*||^2\right]}{\mu^2(\gamma+TEM)}.
   \end{aligned}
   \end{equation}
\end{proof}

\subsection{Deferred Proof of Lemma \ref{lem: bound-sgd}}
\begin{proof}
Notice that $\bar{\boldsymbol{v}}_c^{i+1}=\bar{\boldsymbol{w}}_c^{i}-\eta_c \boldsymbol{g}_c^i(\boldsymbol{\zeta}_c^i) $. Then, 
\begin{equation}\label{15}
\begin{aligned}
||\bar{\boldsymbol{v}}_c^{i+1}-\boldsymbol{w}_c^*||^2 &= || \bar{\boldsymbol{w}}_c^{i}-\eta_c \boldsymbol{g}_c^i(\boldsymbol{\zeta}_c^i) -\boldsymbol{w}_c^* - \eta_c \bar{\boldsymbol{g}}_c^i +\eta_c \bar{\boldsymbol{g}}_c^i||^2\\
&=\underbrace{|| \bar{\boldsymbol{w}}_c^{i} -\boldsymbol{w}_c^*-\eta_c \bar{\boldsymbol{g}}_c^i||^2}_{A_1}  + \underbrace{2\eta_c \langle \bar{\boldsymbol{w}}_c^{i} -\boldsymbol{w}_c^*-\eta_c \bar{\boldsymbol{g}}_c^i, \bar{\boldsymbol{g}}_c^i - \boldsymbol{g}_c^i(\boldsymbol{\zeta}_c^i)\rangle}_{A_2} + \eta_c^2||\bar{\boldsymbol{g}}_c^i-\boldsymbol{g}_c^i(\boldsymbol{\zeta}_c^i)||^2.
\end{aligned}
\end{equation}
It is obvious that $\mathbb{E} \left[A_2\right]=0$ since $\mathbb{E}_{\boldsymbol{\zeta}_c^i}\left[\boldsymbol{g}_c^i(\boldsymbol{\zeta}_c^i)\right]=\bar{\boldsymbol{g}}_c^i$. We next focus on bounding $A_1$, and 
\begin{equation}\label{37}
A_1 = || \bar{\boldsymbol{w}}_c^{i} -\boldsymbol{w}_c^*-\eta_c \bar{\boldsymbol{g}}_c^i||^2= ||\bar{\boldsymbol{w}}_c^{i} -\boldsymbol{w}_c^*||^2 \underbrace{- 2\eta_c \langle \bar{\boldsymbol{w}}_c^{i} -\boldsymbol{w}_c^*, \bar{\boldsymbol{g}}_c^i\rangle}_{B_1} + \underbrace{\eta_c^2 ||\bar{\boldsymbol{g}}_c^i||^2}_{B_2}.
\end{equation}
We can bound $B_2$ as follows:
\begin{equation}
\begin{aligned}
B_2 = \eta_c^2 ||\bar{\boldsymbol{g}}_c^i||^2 &= \eta_c^2 ||\sum_{n\in \mathcal{N}}p_n \nabla f_n(\boldsymbol{v}^i_{c,n}, \boldsymbol{w}_s^{i})||^2\\
& \le \eta_c^2 \sum_{n\in \mathcal{N}}p_n ||\nabla f_n(\boldsymbol{v}^i_{c,n}, \boldsymbol{w}_s^{i})||^2\\
& \le \eta_c^2 \sum_{n\in \mathcal{N}}p_n \cdot 2S (f_n(\boldsymbol{v}^i_{c,n}, \boldsymbol{w}_s^{i})- f^*_n)=2S\eta_c^2 \sum_{n\in \mathcal{N}}p_n   (f_n(\boldsymbol{v}^i_{c,n}, \boldsymbol{w}_s^{i})- f^*_n),
\end{aligned}
\end{equation}
where
\begin{itemize}
\item First inequality: $||\cdot||^2$ is convex and Jensen's inequality 
\item Second inequality: $L$-smoothness and descent lemma (see also Theorem 8.3 in \url{https://stanford.edu/~rezab/dao/notes/L08/cme323_lec8.pdf})
\end{itemize}
We can write $B_1$ as follows:
\begin{equation}\label{39}
\begin{aligned}
B_1 &= - 2\eta_c \langle \bar{\boldsymbol{w}}_c^{i} -\boldsymbol{w}_c^*, \bar{\boldsymbol{g}}_c^i\rangle = -2\eta_c \sum_{n\in \mathcal{N}}p_n \langle \bar{\boldsymbol{w}}_c^{i} -\boldsymbol{w}_c^*, \nabla f_n(\boldsymbol{v}^i_{c,n}, \boldsymbol{w}_s^{i})\rangle\\
&= -2\eta_c \sum_{n\in \mathcal{N}}p_n \langle \bar{\boldsymbol{w}}_c^{i} -\boldsymbol{w}_{c,n}^{i}, \nabla f_n(\boldsymbol{v}^i_{c,n}, \boldsymbol{w}_s^{i})\rangle -2\eta_c \sum_{n\in \mathcal{N}}p_n \langle \boldsymbol{w}_{c,n}^{i} -\boldsymbol{w}_c^*, \nabla f_n(\boldsymbol{v}^i_{c,n}, \boldsymbol{w}_s^{i})\rangle.
\end{aligned}
\end{equation}
By Cauchy-Schwarz inequality ($|\langle b - a, \nabla f(a) \rangle| \leq ||b - a|| \cdot ||\nabla f(a)||$) and AM-GM inequality ($||b - a|| \cdot ||\nabla f(a)|| \leq \frac{1}{2c} ||b - a||^2 + \frac{c}{2} ||\nabla f(a)||^2
$), we have 
\begin{equation}\label{40}
-2\langle \bar{\boldsymbol{w}}_c^{i} -\boldsymbol{w}_{c,n}^{i}, \nabla f_n(\boldsymbol{v}^i_{c,n}, \boldsymbol{w}_s^{i})\rangle \le \frac{1}{\eta_c}||\bar{\boldsymbol{w}}_c^{i} -\boldsymbol{w}_{c,n}^{i}||^2+ \eta_c||\nabla f_n(\boldsymbol{v}^i_{c,n}, \boldsymbol{w}_s^{i})||^2.
\end{equation}
By $\mu$-strong convexity of $f_n(\cdot)$, we have 
\begin{equation}\label{41}
- \langle \boldsymbol{w}_{c,n}^{i} -\boldsymbol{w}_c^*, \nabla f_n(\boldsymbol{v}^i_{c,n}, \boldsymbol{w}_s^{i})\rangle \le -(f_n(\boldsymbol{v}^i_{c,n}, \boldsymbol{w}_s^{i})-f_n(\boldsymbol{w}_c^*, \boldsymbol{w}_s^i))-\frac{\mu}{2}||\boldsymbol{w}_{c,n}^{i} -\boldsymbol{w}_c^*||^2.
\end{equation}
Combing (\ref{37}, \ref{39}, \ref{40}, \ref{41}), we have 
\begin{equation}
\begin{aligned}
A_1&= || \bar{\boldsymbol{w}}_c^{i} -\boldsymbol{w}_c^*-\eta_c \bar{\boldsymbol{g}}_c^i||^2= ||\bar{\boldsymbol{w}}_c^{i} -\boldsymbol{w}_c^*||^2 + \underbrace{\eta_c^2 ||\bar{\boldsymbol{g}}_c^i||^2}_{B_2} \underbrace{- 2\eta_c \langle \bar{\boldsymbol{w}}_c^{i} -\boldsymbol{w}_c^*, \bar{\boldsymbol{g}}_c^i\rangle}_{B_1} \\
& \le ||\bar{\boldsymbol{w}}_c^{i} -\boldsymbol{w}_c^*||^2 + 2S\eta_c^2 \sum_{n\in \mathcal{N}}p_n   (f_n(\boldsymbol{v}^i_{c,n}, \boldsymbol{w}_s^{i})- f^*_n) \\
  &\hspace{5mm} + \eta_c \sum_{n\in \mathcal{N}}p_n \left(\frac{1}{\eta_c}||\bar{\boldsymbol{w}}_c^{i} -\boldsymbol{w}_{c,n}^{i}||^2+ \eta_c||\nabla f_n(\boldsymbol{v}^i_{c,n}, \boldsymbol{w}_s^{i})||^2\right) \\
  &\hspace{5mm}-2\eta_c \sum_{n\in \mathcal{N}}p_n\left(f_n(\boldsymbol{v}^i_{c,n}, \boldsymbol{w}_s^{i})-f_n(\boldsymbol{w}_c^*, \boldsymbol{w}_s^i)+\frac{\mu}{2}||\boldsymbol{w}_{c,n}^{i} -\boldsymbol{w}_c^*||^2\right)\\
  &\le (1-\mu\eta_c)||\bar{\boldsymbol{w}}_c^i-\boldsymbol{w}_c^*||^2 +\sum_{n \in \mathcal{N}}p_n||\bar{\boldsymbol{w}}_c^i- \boldsymbol{w}_{c,n}^i||^2 \\
  & \hspace{5mm}+\underbrace{4S\eta_c^2 \sum_{n\in \mathcal{N}}p_n(f_n(\boldsymbol{v}^i_{c,n}, \boldsymbol{w}_s^{i})-f^*_n)-2\eta_c \sum_{n\in \mathcal{N}}p_n\left(f_n(\boldsymbol{v}^i_{c,n}, \boldsymbol{w}_s^{i})-f_n(\boldsymbol{w}^*_{c}, \boldsymbol{w}_s^{i})\right)}_{C}.
\end{aligned}
\end{equation}
We next aim to bound $C$.
We define $\gamma_c=2\eta_c(1-2S\eta_c)$. Since $\eta_c\le \frac{1}{4S}$ by assumption, we have $\eta_c\le \gamma_c\le 2\eta_c$. Then we split $C$ into two terms:
\begin{equation}
\begin{aligned}
C= &-2\eta_c(1-2S\eta_c)\sum_{n\in \mathcal{N}}p_n(f_n(\boldsymbol{v}^i_{c,n}, \boldsymbol{w}_s^{i})-f^*_n) + 2\eta_c\sum_{n\in \mathcal{N}}p_n(f_n(\boldsymbol{w}^*_{c}, \boldsymbol{w}_s^{i})-f_n^*)\\
& =-\gamma_c\sum_{n\in \mathcal{N}}p_n(f_n(\boldsymbol{v}^i_{c,n}, \boldsymbol{w}_s^{i})-f^*_n) + 2\eta_c\sum_{n\in \mathcal{N}}p_n(f_n(\boldsymbol{w}^*_{c}, \boldsymbol{w}_s^{i})-f_n^*)\\
&\le -\gamma_c\sum_{n\in \mathcal{N}}p_n(f_n(\boldsymbol{v}^i_{c,n}, \boldsymbol{w}_s^{i})-F^*)+(2\eta_c-\gamma_c)\Gamma\\
& = -\gamma_c\sum_{n\in \mathcal{N}}p_n(f_n(\boldsymbol{v}^i_{c,n}, \boldsymbol{w}_s^{i})-F^*)+4S\eta_c^2\Gamma\\
& \le 6S\eta_c^2\Gamma + \sum_{n\in \mathcal{N}}p_n||\boldsymbol{w}_{c,n}^i-\boldsymbol{w}_c^*||^2,
\end{aligned}
\end{equation}
where $\Gamma\triangleq F(\boldsymbol{w}^*)-\sum_{n \in \mathcal{N}}p_n f_n(\boldsymbol{w}_{c,n}^*, \boldsymbol{w}_s).$
Hence, we have 
\begin{equation}\label{A1-bound}
A_1\le (1-\mu\eta_c)||\bar{\boldsymbol{w}}_c^i-\boldsymbol{w}_c^*||^2 +2\sum_{n \in \mathcal{N}}p_n||\bar{\boldsymbol{w}}_c^i- \boldsymbol{w}_{c,n}^i||^2+6S\eta_c^2\Gamma.
\end{equation}
Combining (\ref{A1-bound}) and (\ref{15}), and taking expectations over (\ref{15}), we finish the proof of Lemma 1. 
\end{proof}

\newpage
\section{Hyper-parameters and Model Splitting}
For the experiments, we use the local epoch number as $E=5$, batch size as $32$, and the learning rates for MiniBatch-SFL, CL, FL, and SFL-V2 are 0.01, 0.01, 0.01, 0.01, respectively. 

For the model ResNet-18, the model splitting details are as follows. 
\begin{itemize}
\item $L_c=1$: ResNet-18 is split after the first residual block.
\item $L_c=2$: ResNet-18 is split after the second residual block.
\item $L_c=3$: ResNet-18 is split after the third residual block.
\item $L_c=4$: ResNet-18 is split after the fourth residual block.
\end{itemize}

\section{Section 4: Additional Experiments}
The additional experiments are organized as follows:
\begin{itemize}
\item Section 4.1: a different model structure using CNN
\item Section 4.2: a different performance metric using loss
\item Section 4.3: comparison on a different dataset using FMNIST
\end{itemize}
\subsection{Section 4.1: Experiments on CNN}
\begin{figure}[h]
       \begin{subfigure}{0.24\textwidth}
        \centering
        \includegraphics[height=3.4cm]{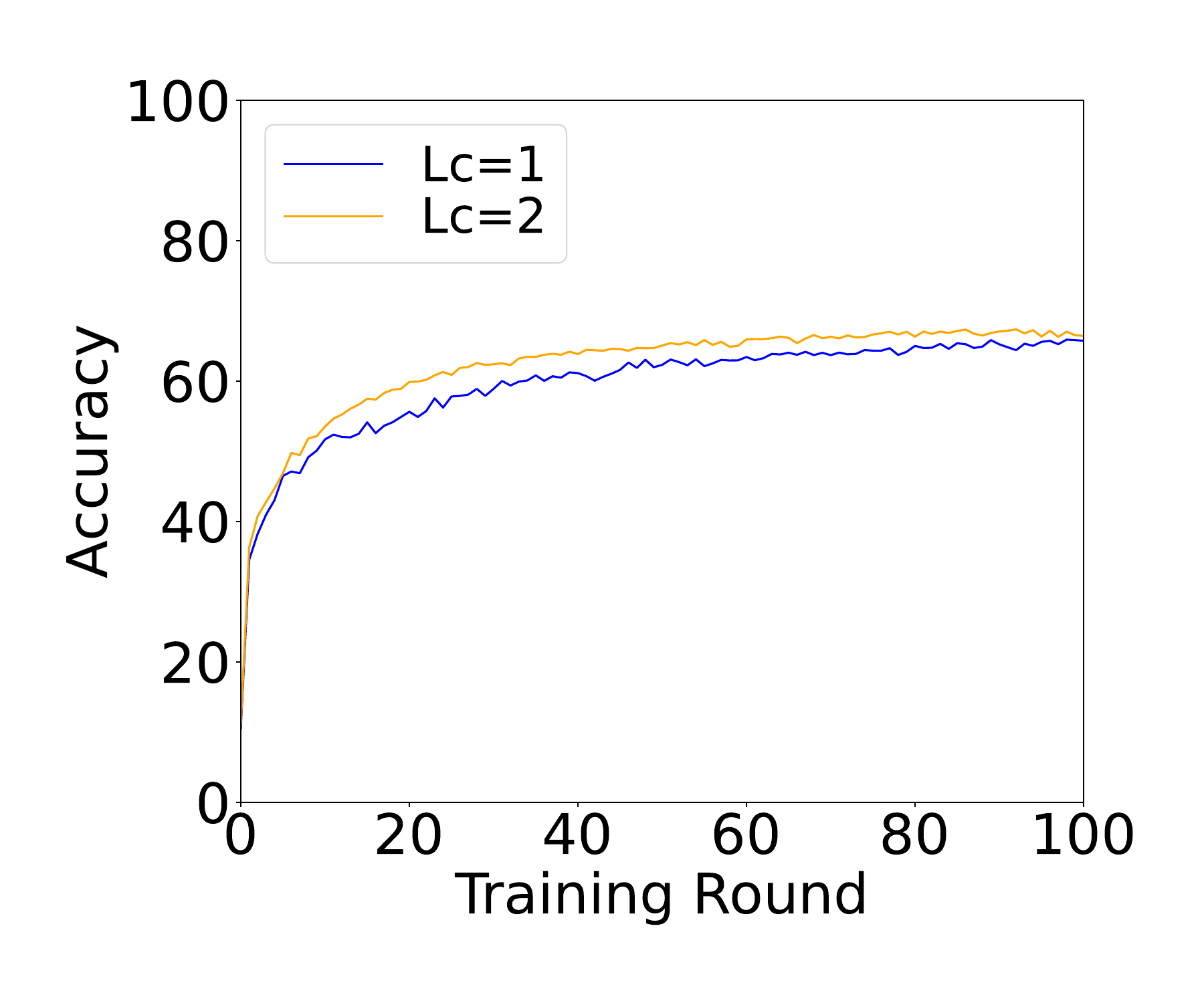}
        \caption{$r\%=0.5$.}
    \end{subfigure}
    \hfil
    \begin{subfigure}{0.24\textwidth}
        \centering
        \includegraphics[height=3.4cm]{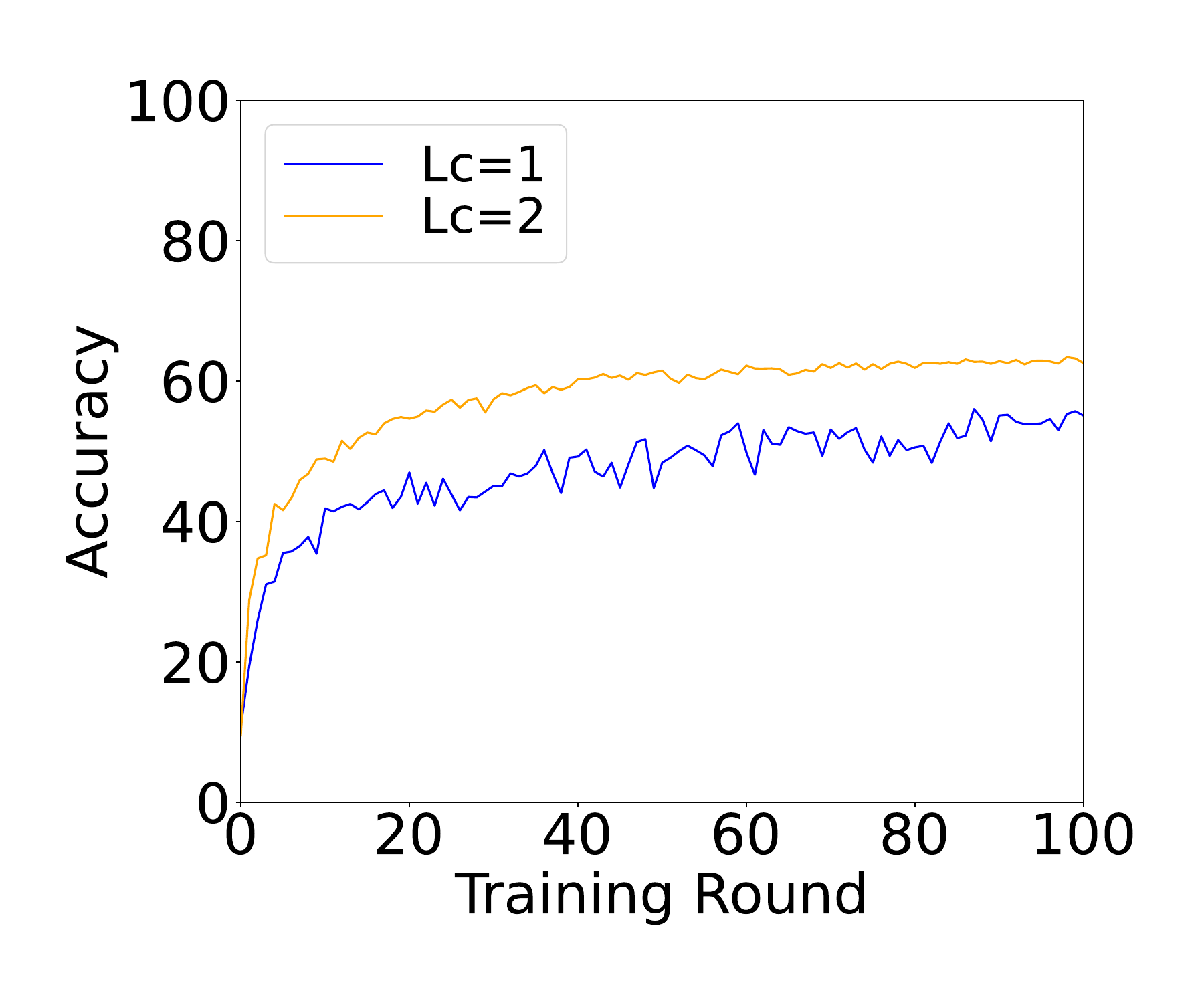}
        \caption{$r\%=0.8$.}
    \end{subfigure}
    \hfil
    \begin{subfigure}{0.24\textwidth}
        \centering
        \includegraphics[height=3.4cm]{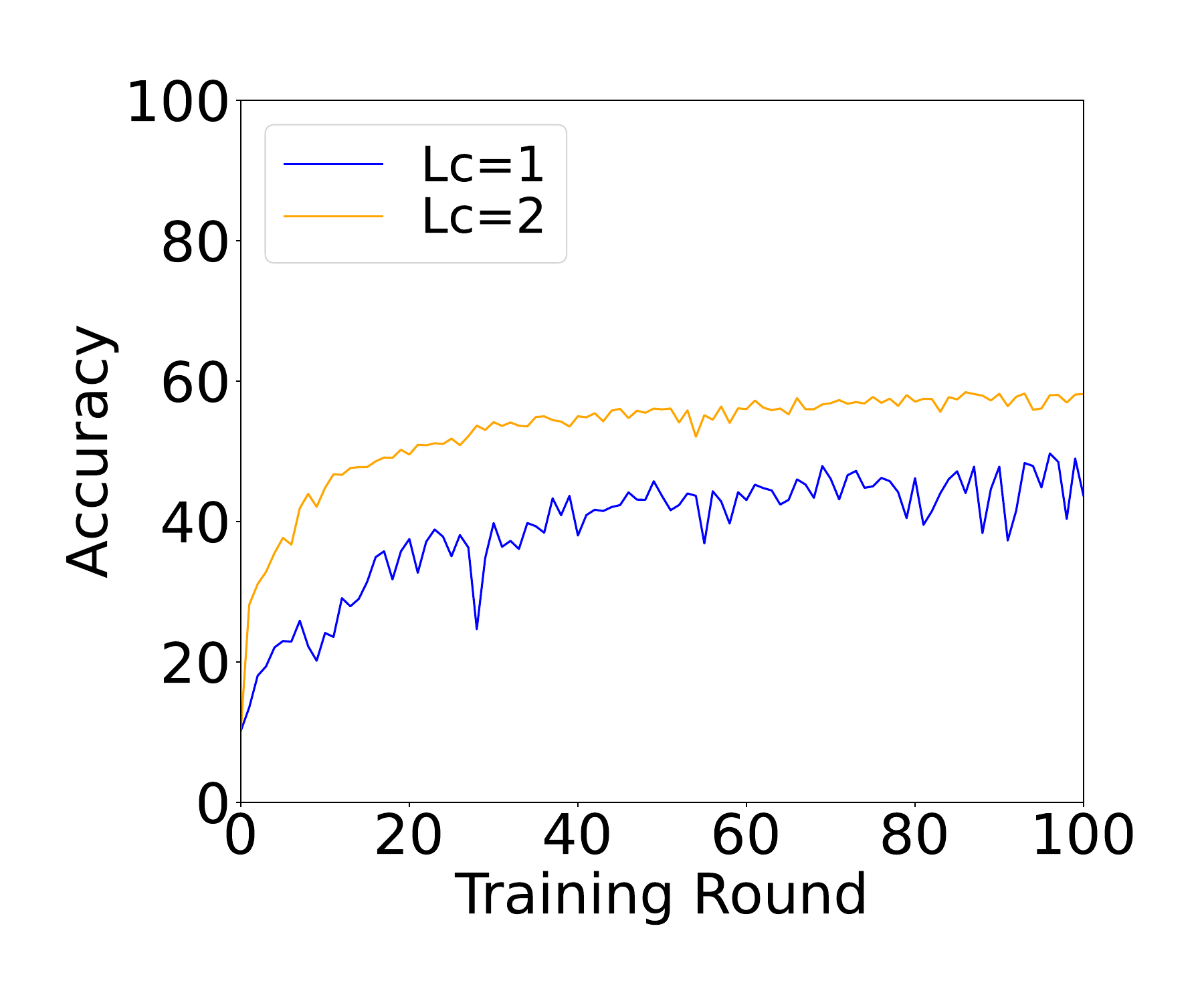}
        \caption{$r\%=0.9$.}
    \end{subfigure}
    \hfil 
    \begin{subfigure}{0.24\textwidth}
        \centering
        \includegraphics[height=3.4cm]{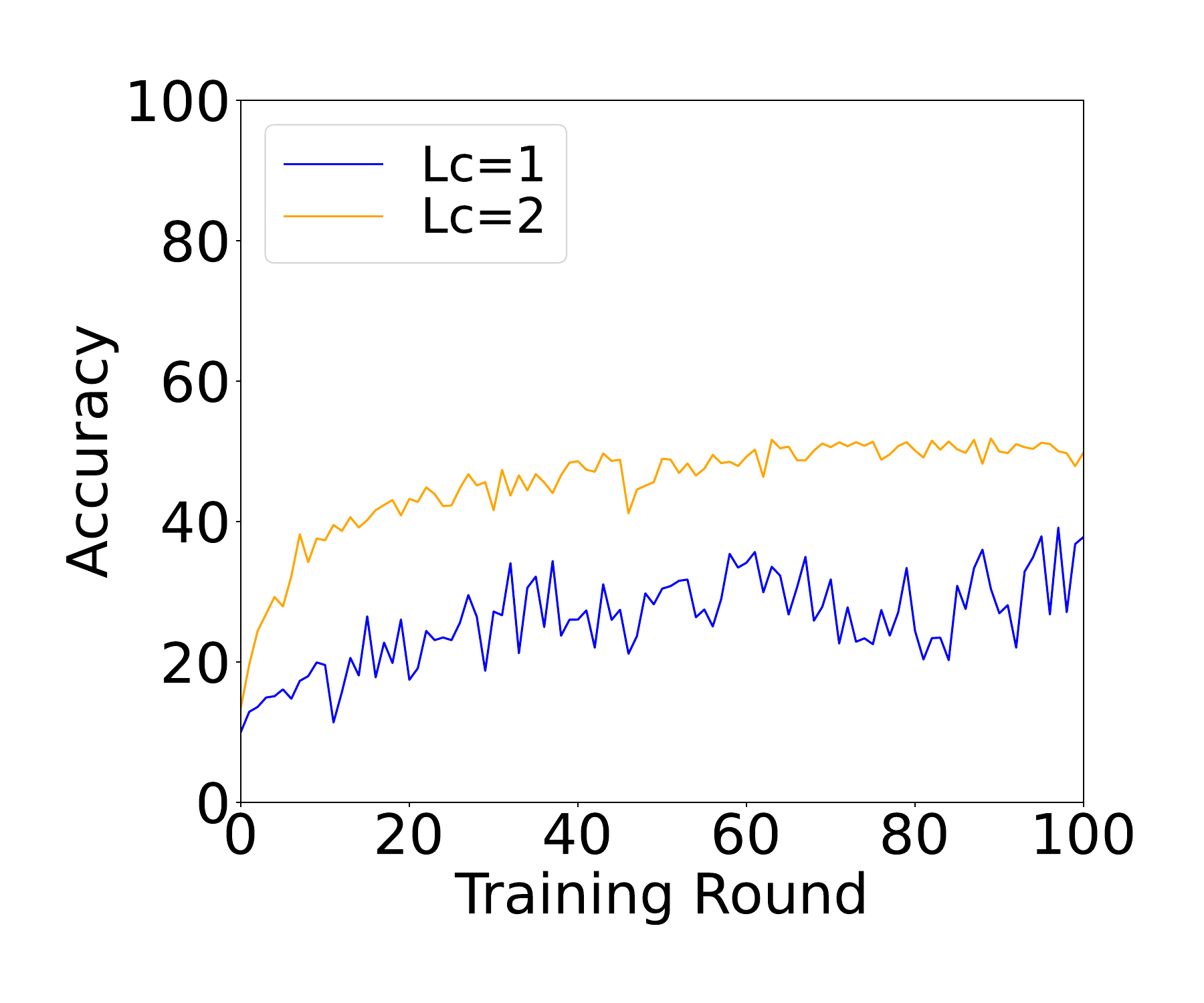}
        \caption{$r\%=0.95$.}
    \end{subfigure}
    \caption{Accuracy of MiniBatch-SFL using 4-layer CNN with $N=10$ on CIFAR-10.}
    \label{fig:accuracy-cnn}
\end{figure}

In addition to the ResNet-18 model used in the main paper, we also consider a simpler model structure, i.e., a 4-layer CNN. We consider two types of model splitting:
\begin{itemize}
\item $L_c=1$ corresponds to the split after the first group of convolutional layer and max-pooling layer
\item $L_c=2$ corresponds to the split after the second group of convolutional layer and max-pooling layer
\end{itemize}
The accuracy results are reported in Fig. \ref{fig:accuracy-cnn}. 

\begin{figure}[h]
       \begin{subfigure}{0.24\textwidth}
        \centering
        \includegraphics[height=3.4cm]{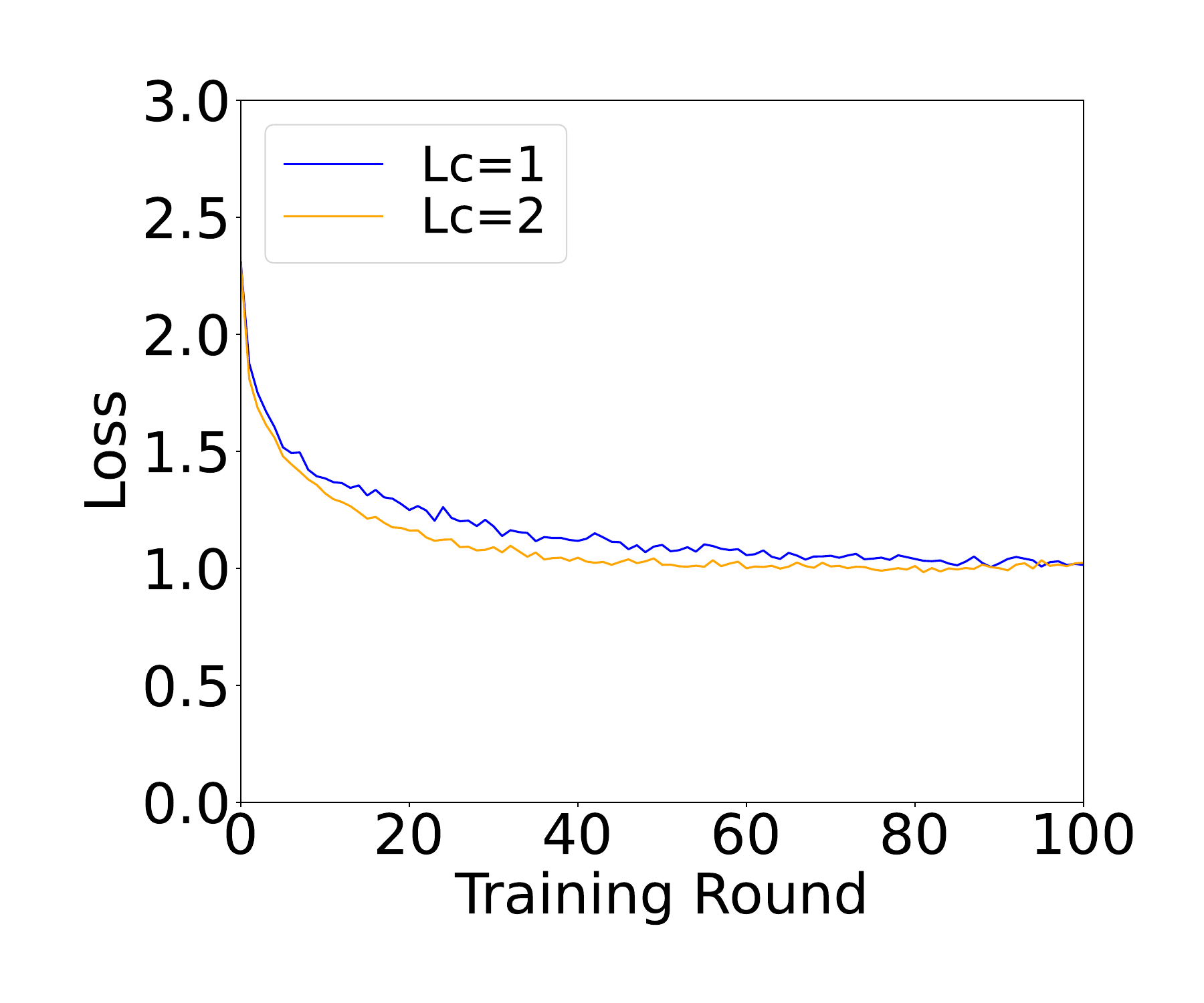}
        \caption{$r\%=0.5$.}
    \end{subfigure}
    \hfil
    \begin{subfigure}{0.24\textwidth}
        \centering
        \includegraphics[height=3.4cm]{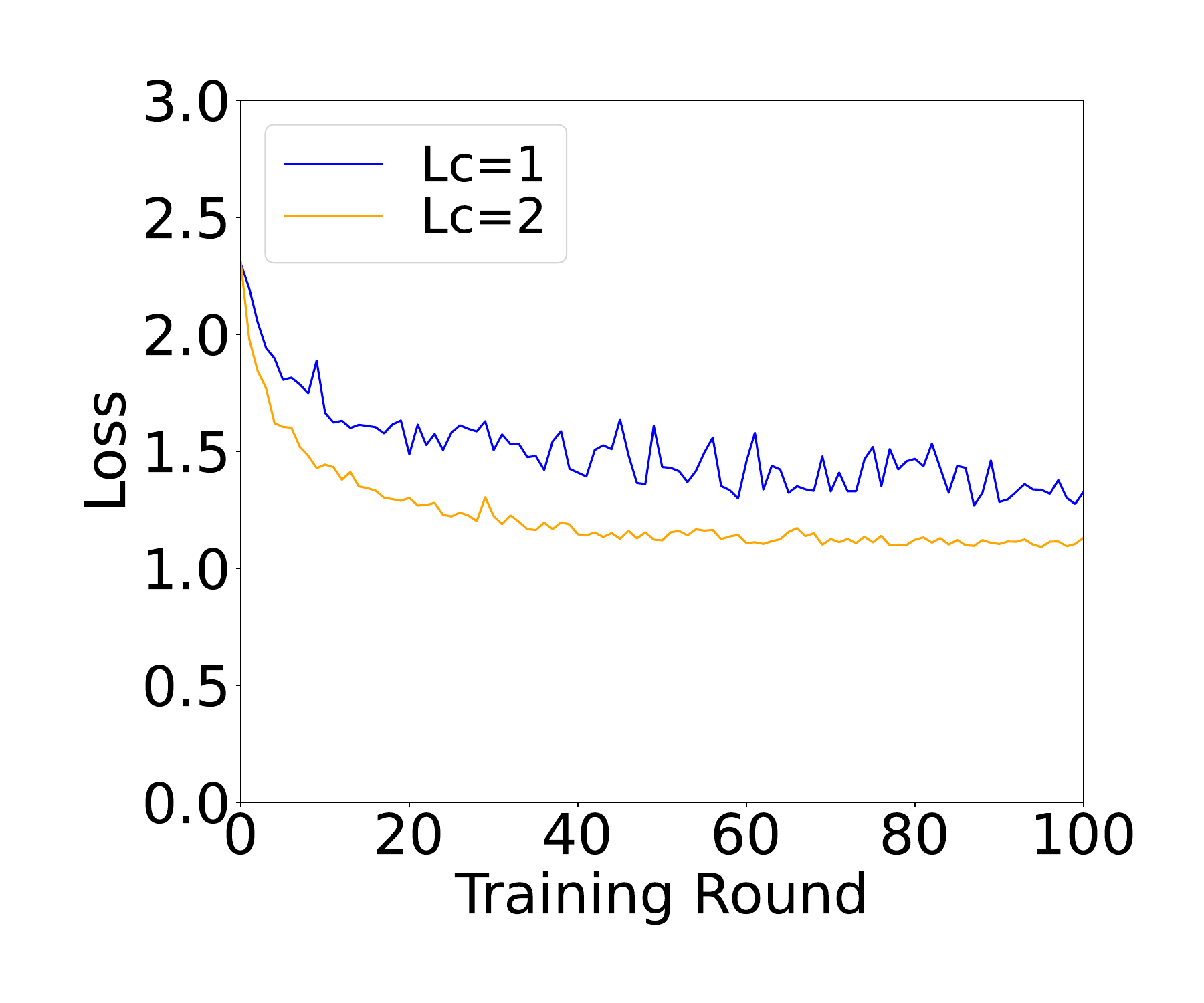}
        \caption{$r\%=0.8$.}
    \end{subfigure}
    \hfil
    \begin{subfigure}{0.24\textwidth}
        \centering
        \includegraphics[height=3.4cm]{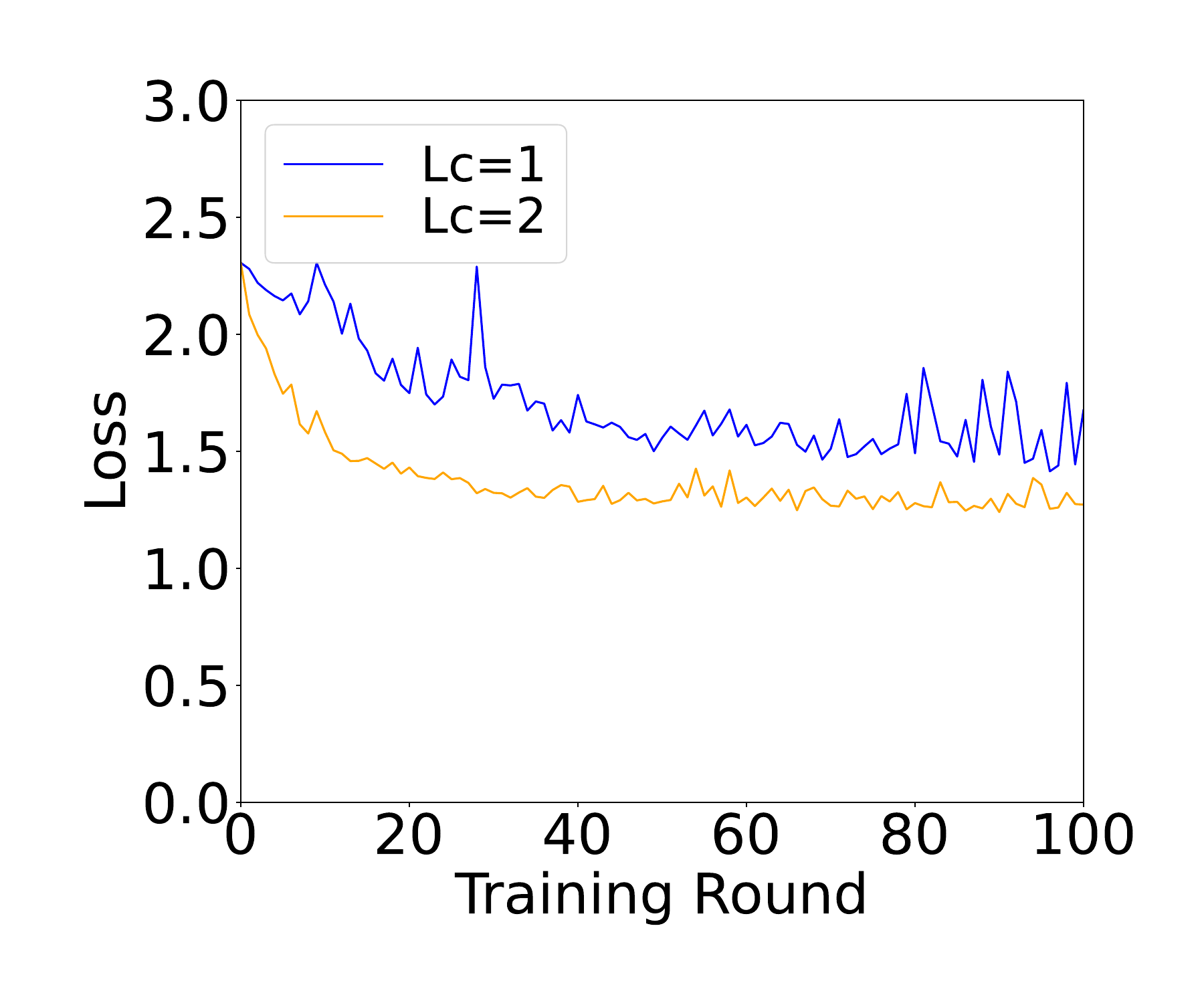}
        \caption{$r\%=0.9$.}
    \end{subfigure}
    \hfil 
    \begin{subfigure}{0.24\textwidth}
        \centering
        \includegraphics[height=3.4cm]{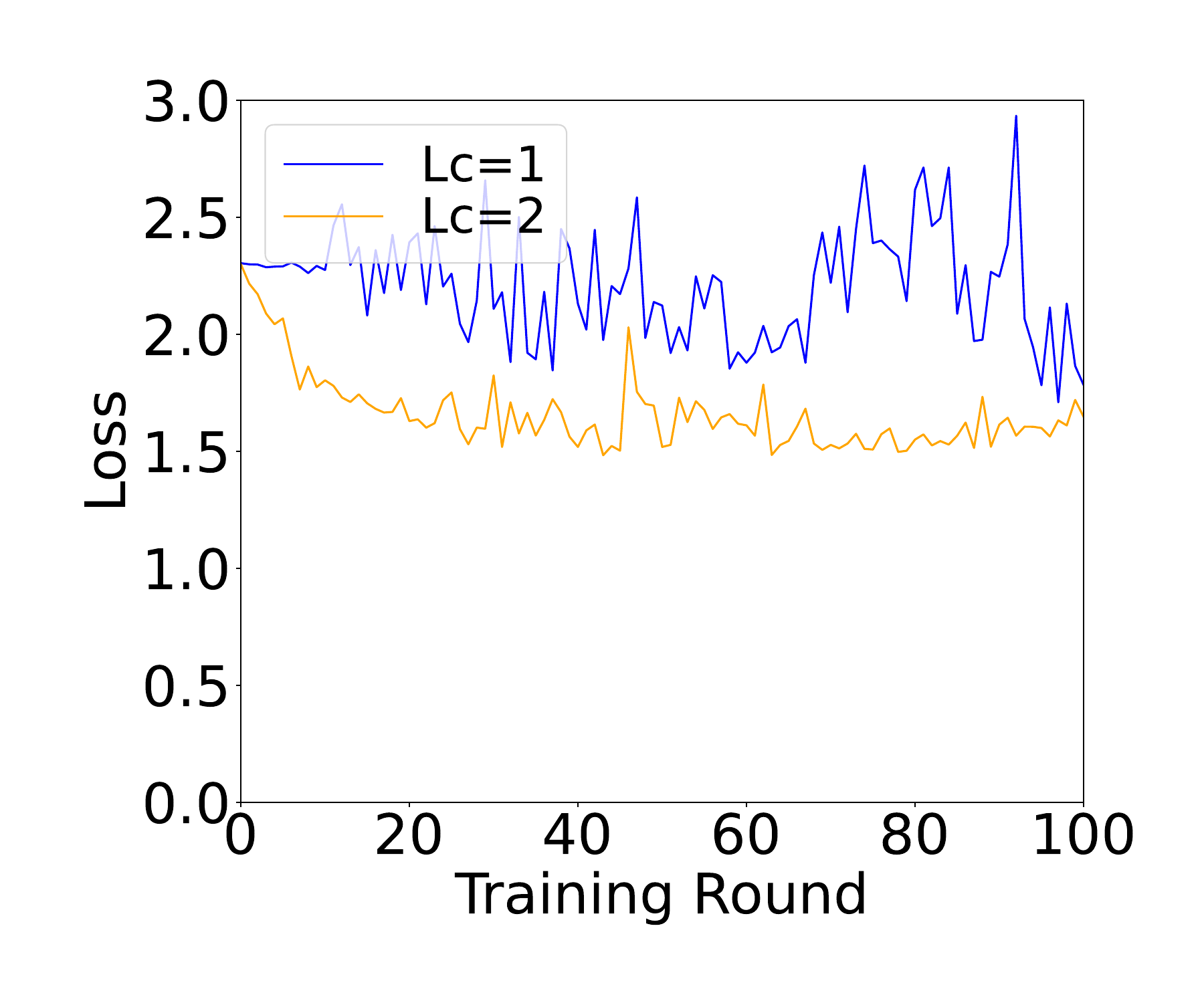}
        \caption{$r\%=0.95$.}
    \end{subfigure}
    \caption{Loss of MiniBatch-SFL using 4-layer CNN with $N=10$ on CIFAR-10.}
    \label{fig:loss-cnn}
\end{figure}

From Figs. \ref{fig:loss-cnn}-\ref{fig:accuracy-cnn}, we observe that the loss decreases while the model accuracy increases in $L_c$, and the impact is more significant when the non-IID ratio is higher. This is consistent with Observation 1 in the main paper.



\subsection{Section 4.2: Results represented by Loss}
We now present the loss results. 
\subsubsection{Impact of cut layer in MiniBatch-SFL}
\begin{figure}[h]
       \begin{subfigure}{0.24\textwidth}
        \centering
        \includegraphics[height=3.4cm]{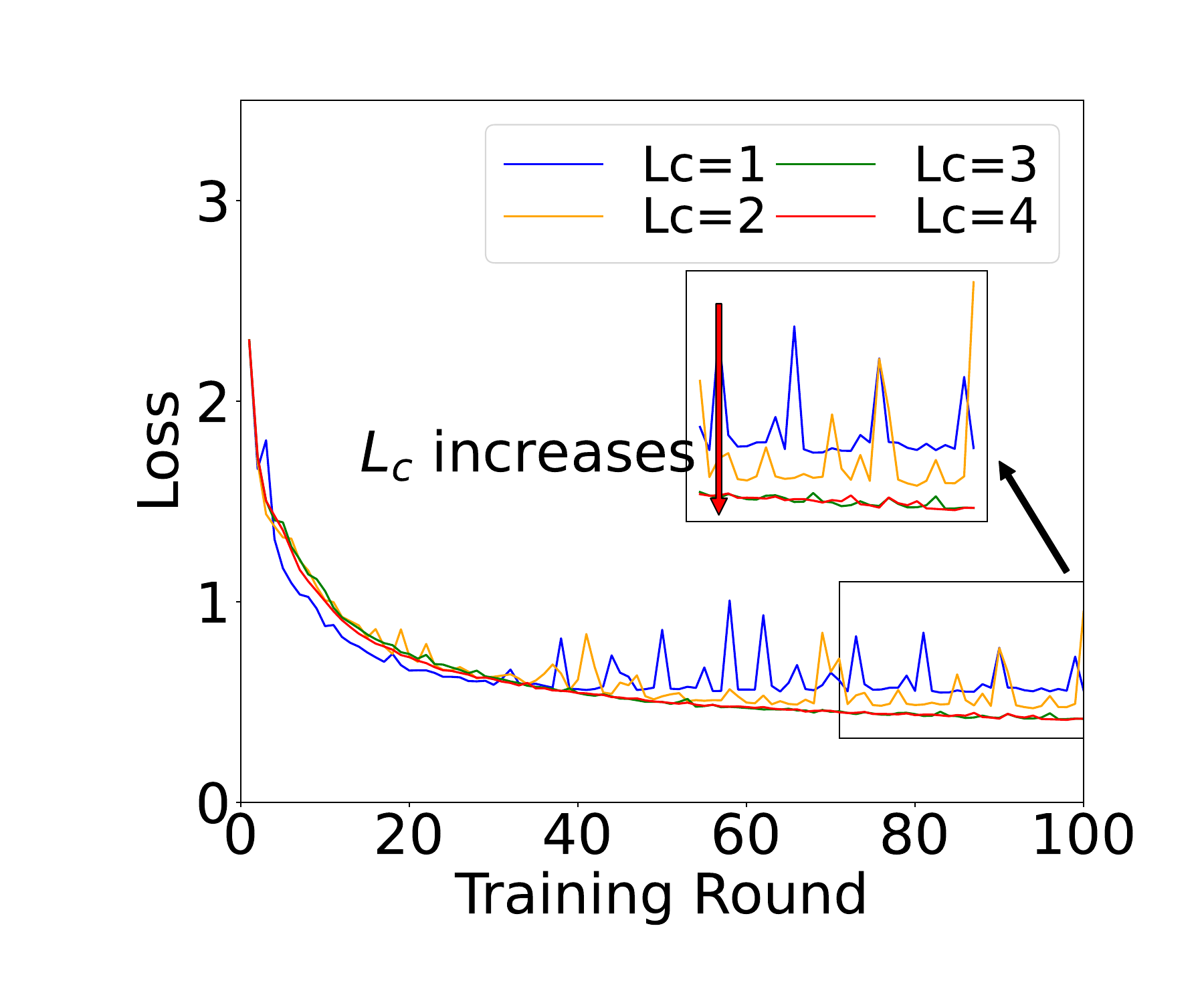}
        \caption{$r\%=0$ (CIFAR-10).}
        \label{subfig:accuracy0.5}
    \end{subfigure}
    \hfil
    \begin{subfigure}{0.24\textwidth}
        \centering
        \includegraphics[height=3.4cm]{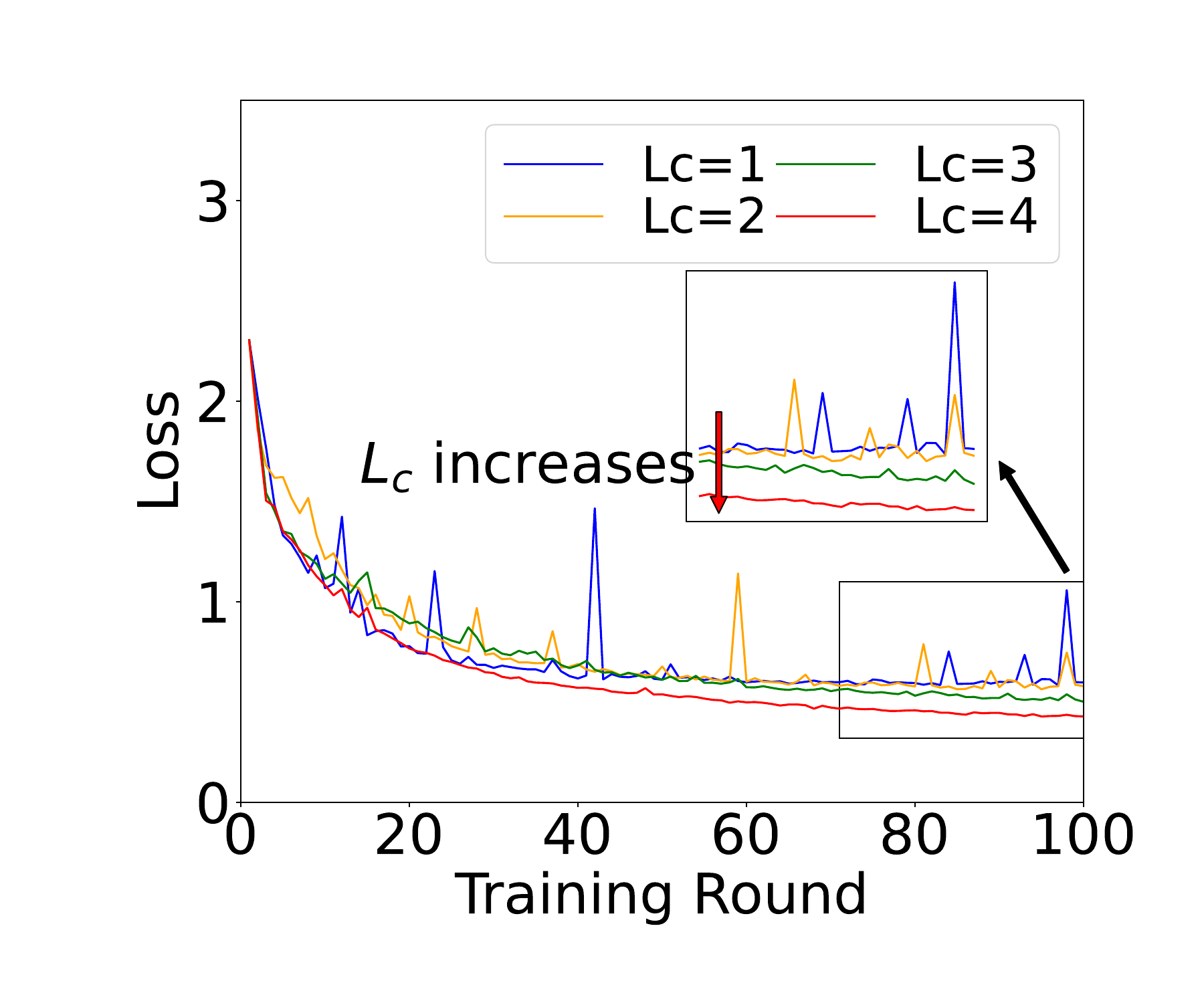}
        \caption{$r\%=0.5$ (CIFAR-10).}
        \label{subfig:accuracy0.5}
    \end{subfigure}
    \hfil
    \begin{subfigure}{0.24\textwidth}
        \centering
        \includegraphics[height=3.4cm]{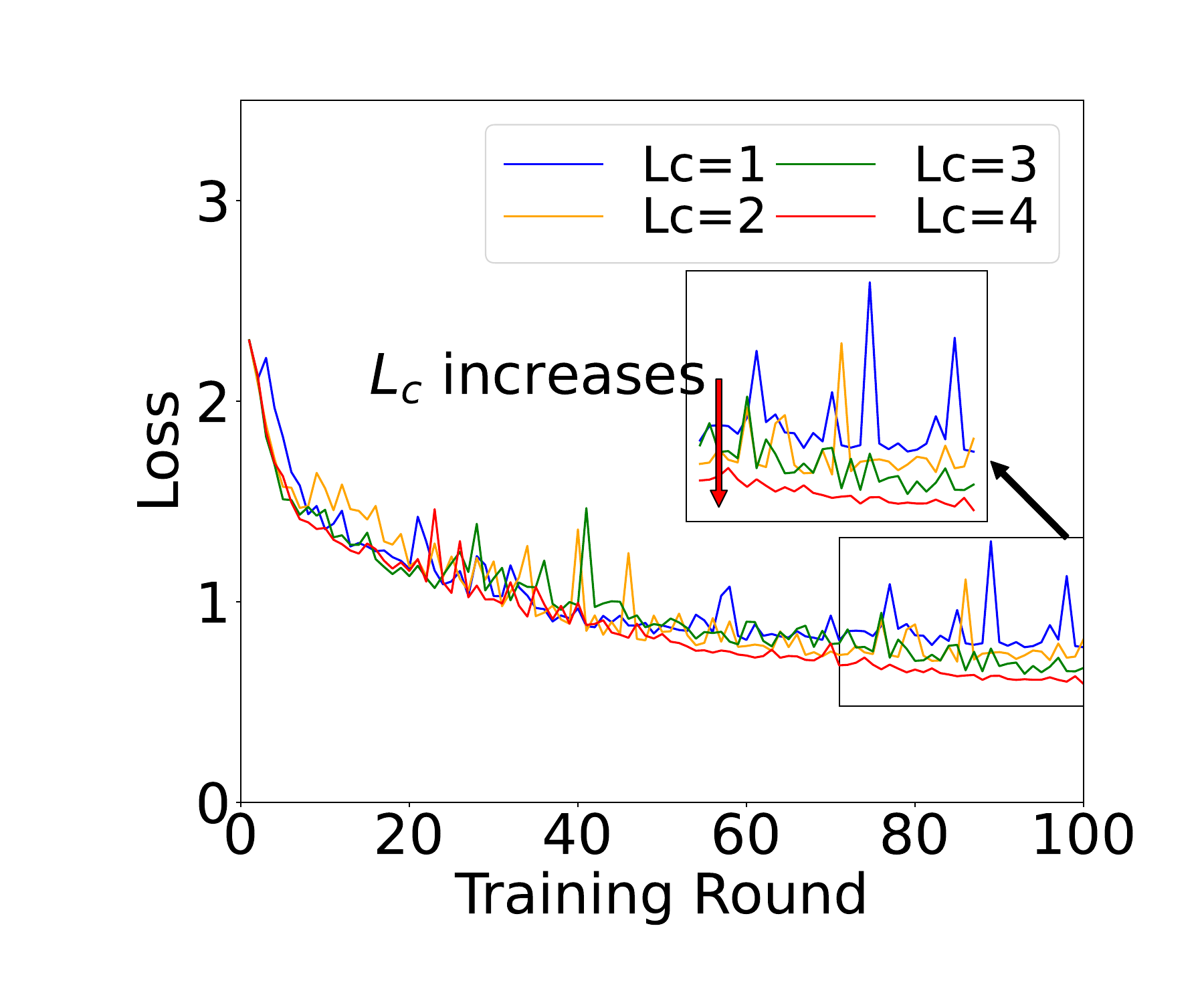}
        \caption{$r\%=0.8$ (CIFAR-10).}
        \label{subfig:accuracy0.8}
    \end{subfigure}
    \hfil 
    \begin{subfigure}{0.24\textwidth}
        \centering
        \includegraphics[height=3.4cm]{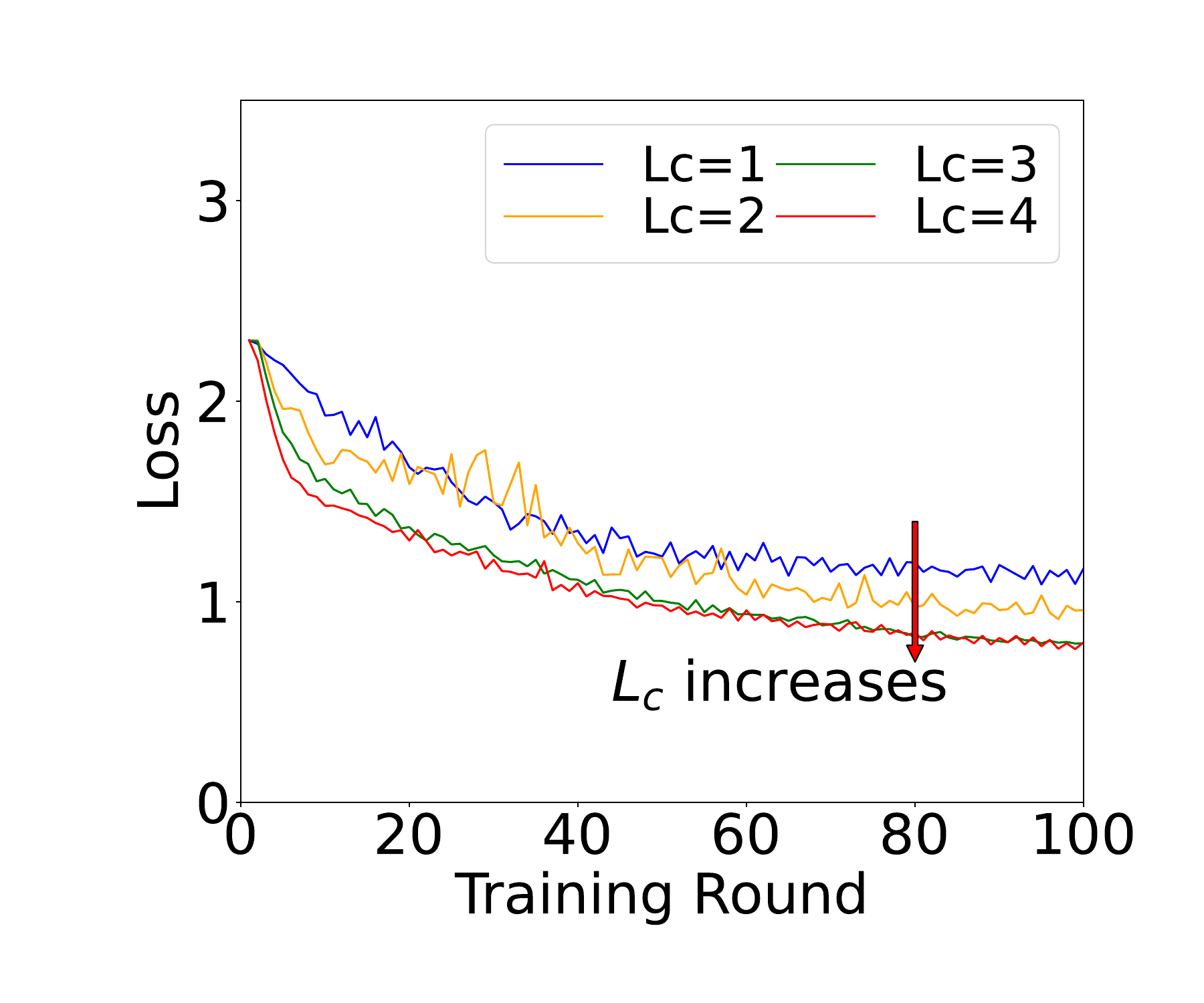}
        \caption{$r\%=0.9$ (CIFAR-10).}
        \label{subfig:accuracy0.9}
    \end{subfigure}
    \hfil \\
    \begin{subfigure}{0.24\textwidth}
        \centering
        \includegraphics[height=3.4cm]{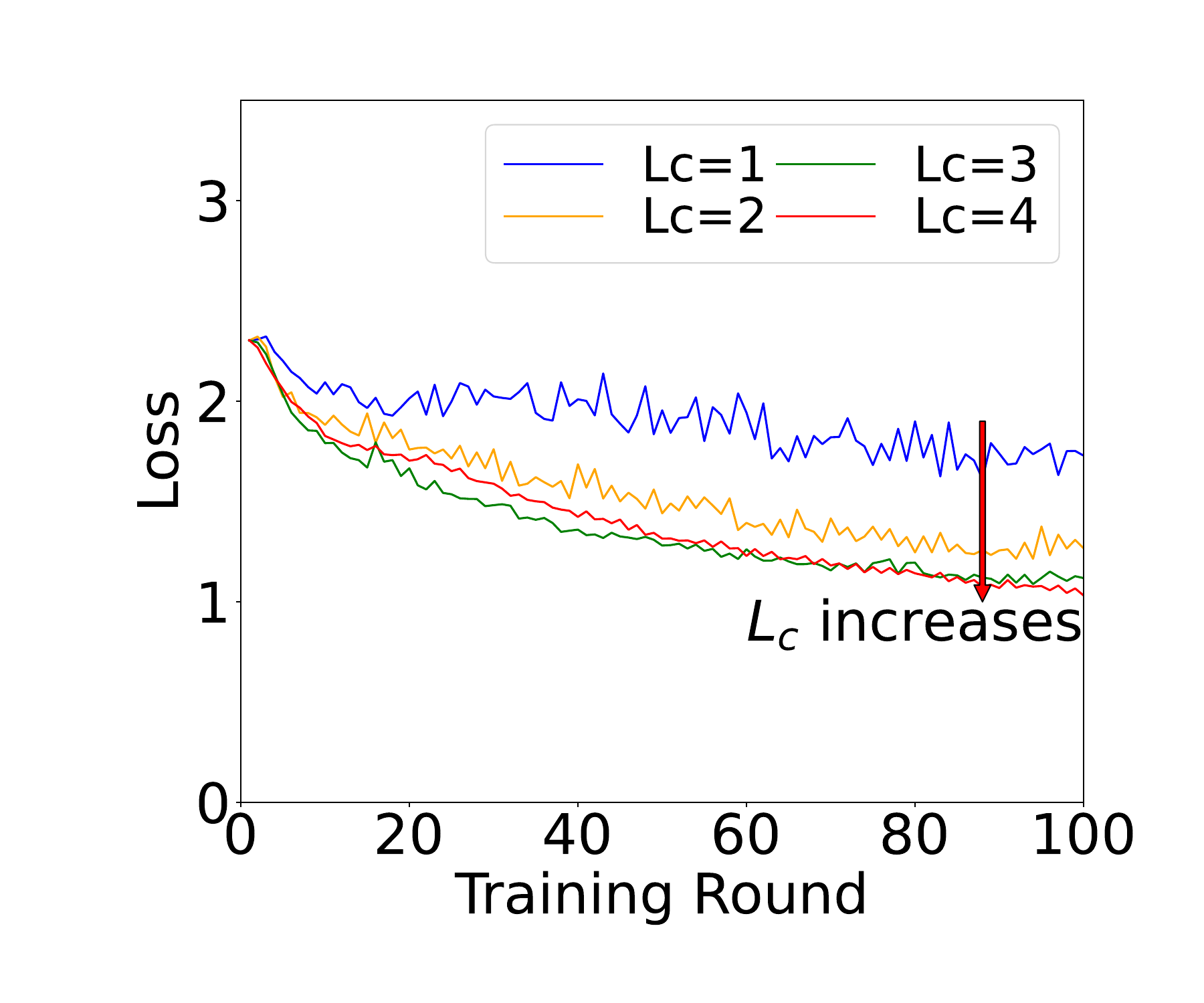}
        \caption{$r\%=0.95$ (CIFAR-10).}
        \label{subfig:accuracy0.95}
    \end{subfigure}
    \hfil
        \begin{subfigure}{0.24\textwidth}
        \centering
        \includegraphics[height=3.4cm]{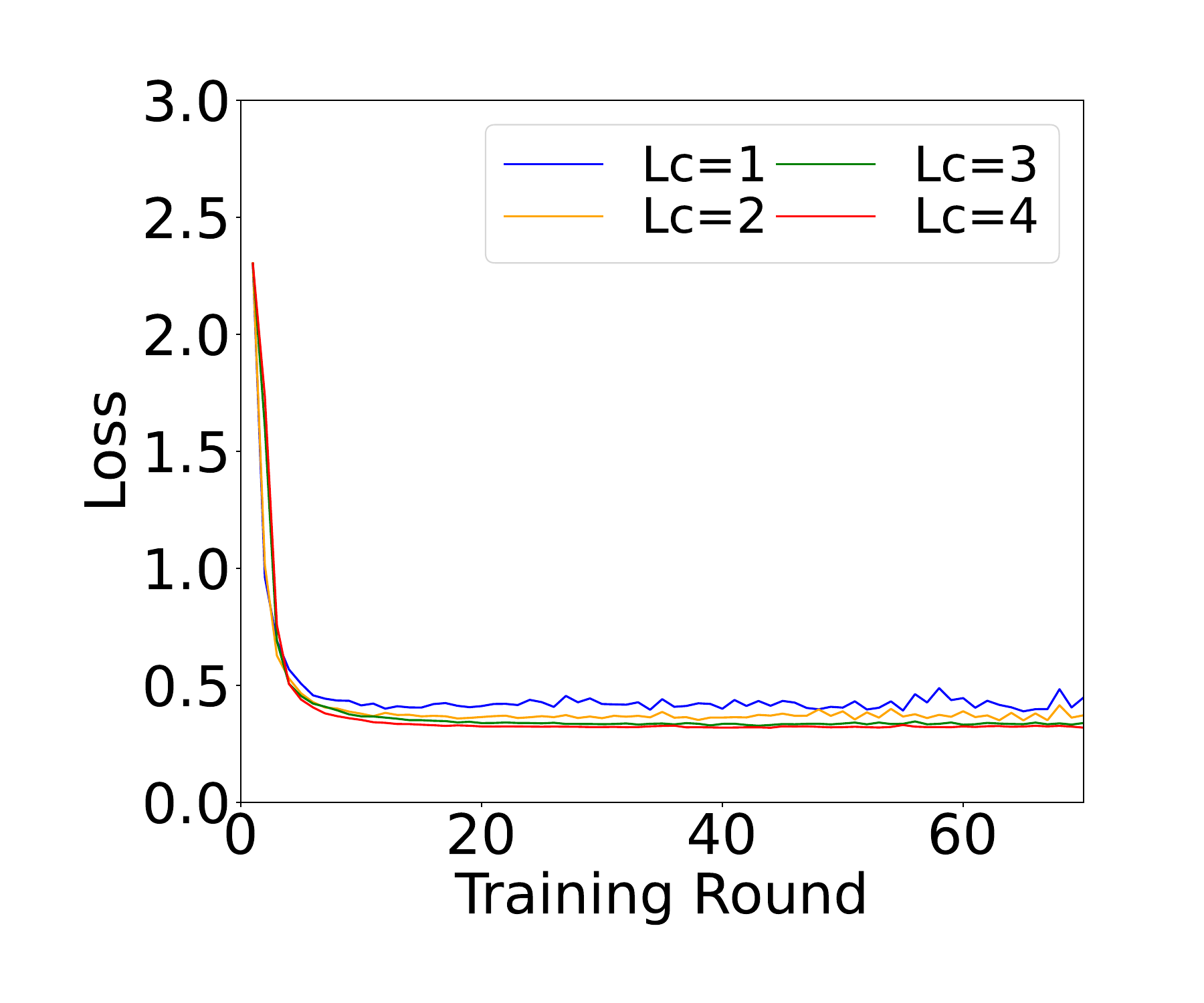}
        \caption{$r\%=0.8$ (FMNIST).}
    \end{subfigure}
    \hfil
    \begin{subfigure}{0.24\textwidth}
        \centering
        \includegraphics[height=3.4cm]{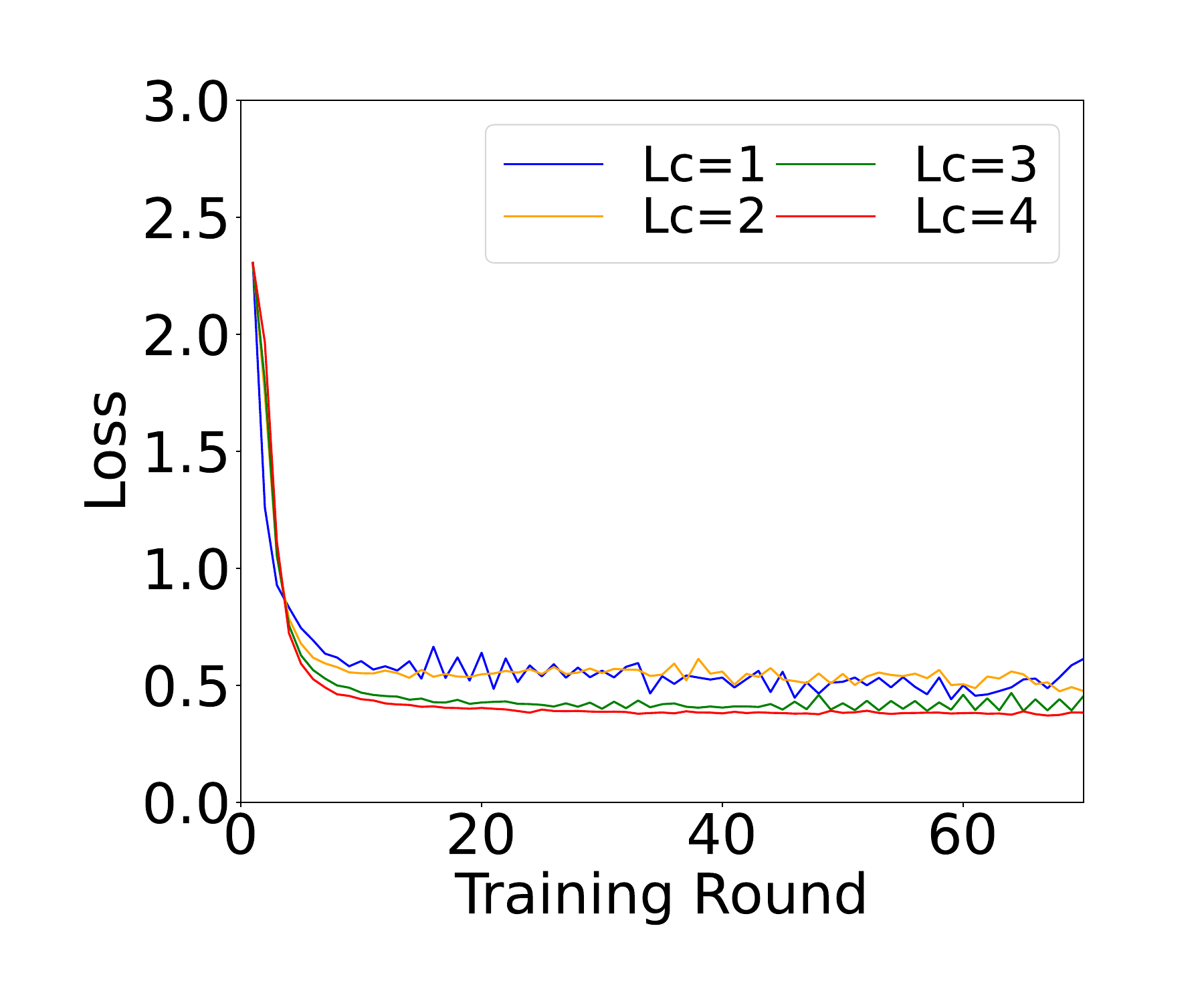}
        \caption{$r\%=0.9$ (FMNIST).}
    \end{subfigure}
    \hfil
    \begin{subfigure}{0.24\textwidth}
        \centering
        \includegraphics[height=3.4cm]{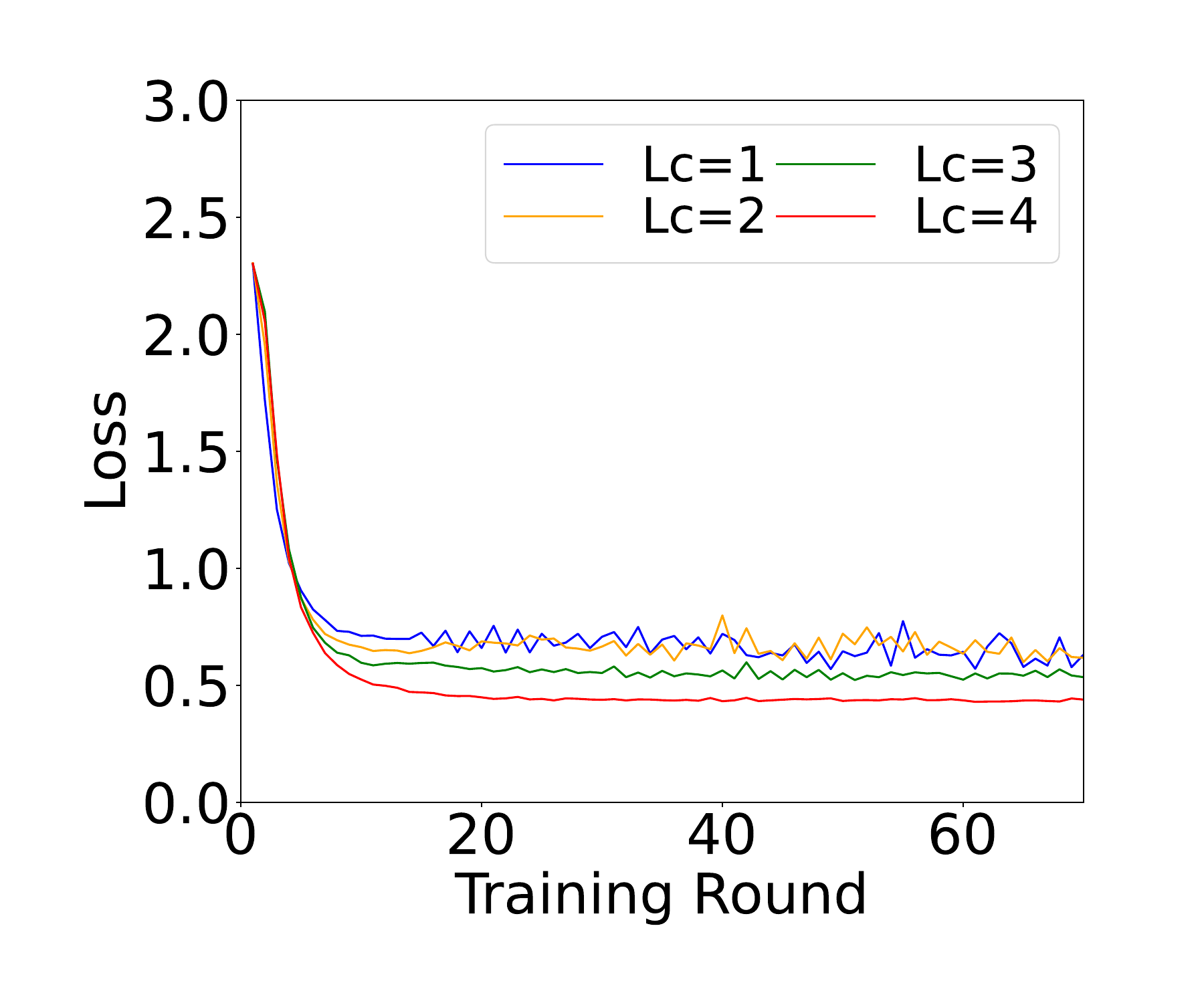}
        \caption{$r\%=0.95$ (FMNIST).}
    \end{subfigure}
    \caption{Impact of cut layer under different non-IID ratios: (a-e) CIFAR-10, (f-h) FMNIST.}
    \label{fig:cut-loss-cifar}
\end{figure}

We first plot the loss under different choices of cut layers in Fig. \ref{fig:cut-loss-cifar}, where we consider $N=10$.
From Fig. \ref{fig:cut-loss-cifar}, we observe that the loss generally decreases in $L_c$, meaning that a larger $L_c$ corresponds to a better trained model. This is consistent with Observation 1 in the main paper. 

\subsubsection{Performance comparison}
\begin{figure}[h]
       \begin{subfigure}{0.24\textwidth}
        \centering
        \includegraphics[height=3.4cm]{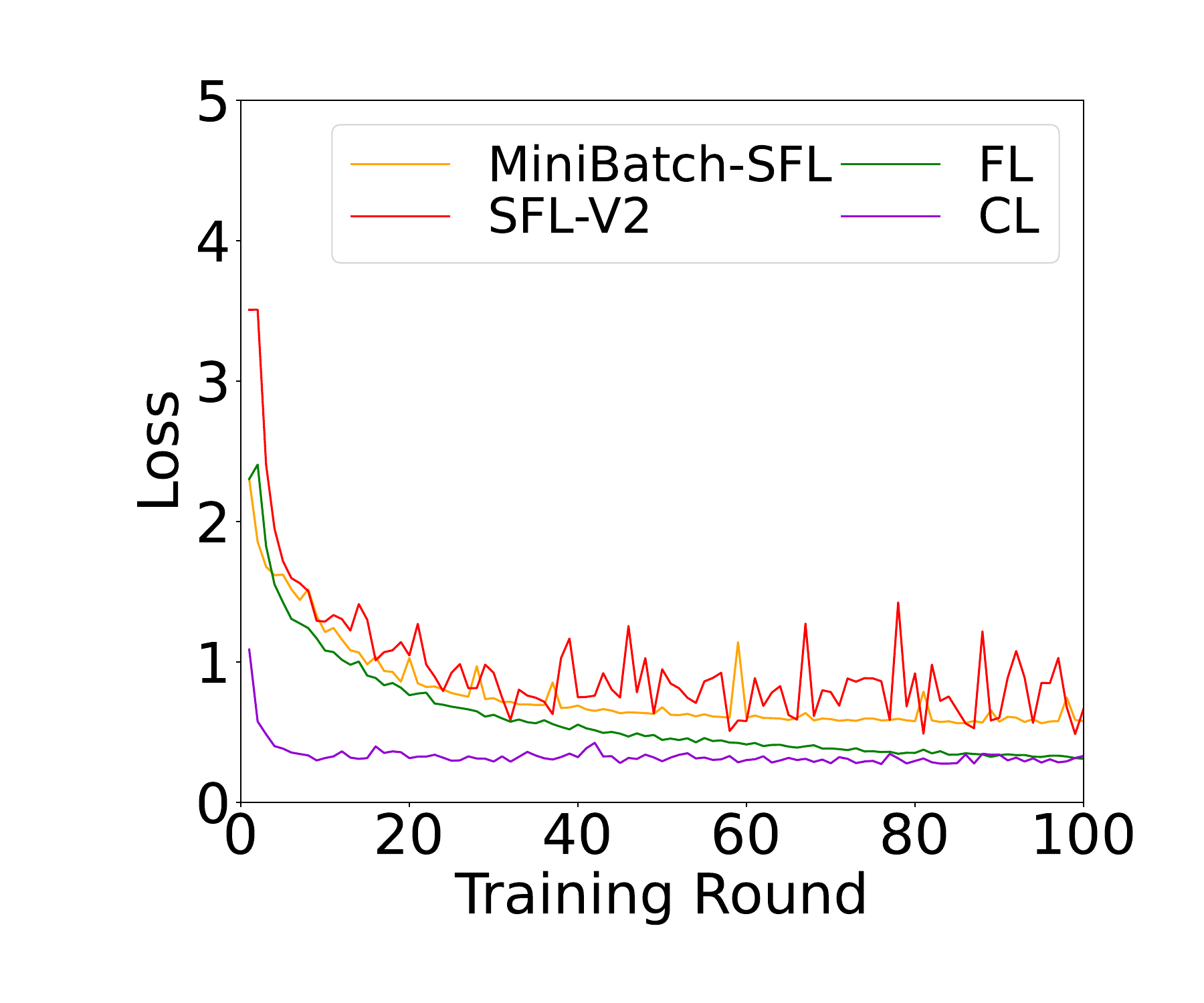}
        \caption{$r\%=0.5$.}
    \end{subfigure}
    \hfil
    \begin{subfigure}{0.24\textwidth}
        \centering
        \includegraphics[height=3.4cm]{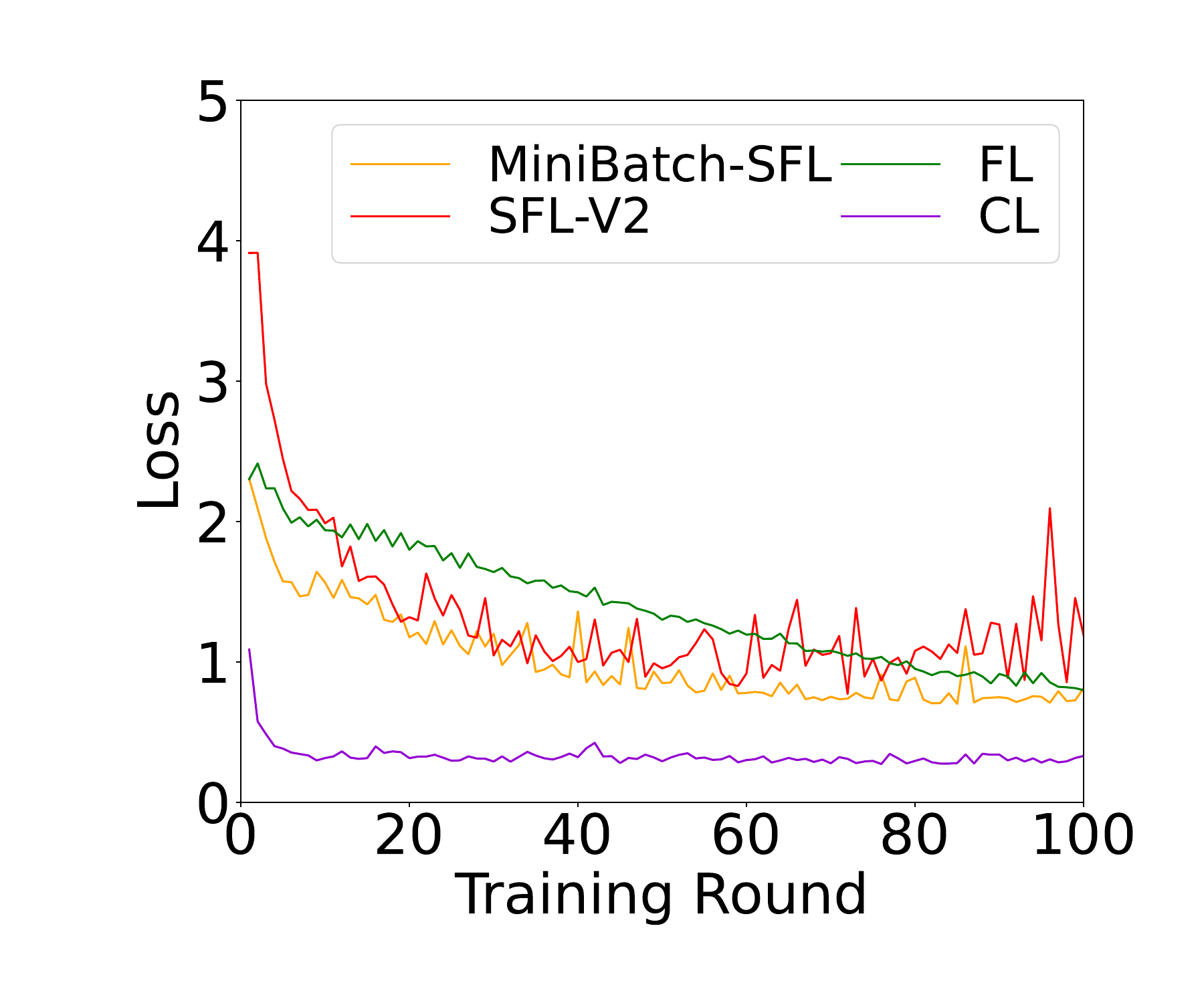}
        \caption{$r\%=0.8$.}
    \end{subfigure}
    \hfil
    \begin{subfigure}{0.24\textwidth}
        \centering
        \includegraphics[height=3.4cm]{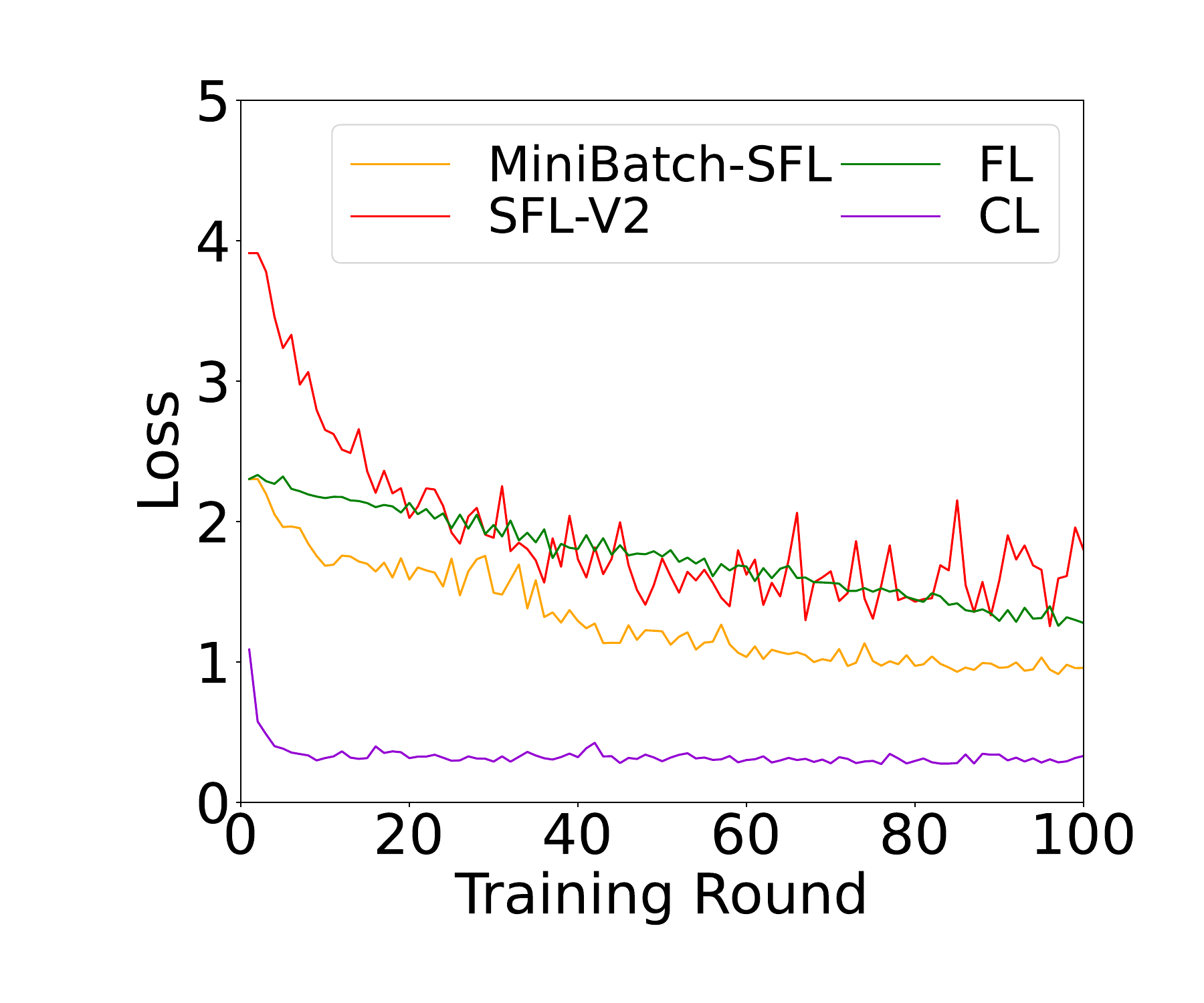}
        \caption{$r\%=0.9$.}
    \end{subfigure}
    \hfil 
    \begin{subfigure}{0.24\textwidth}
        \centering
        \includegraphics[height=3.4cm]{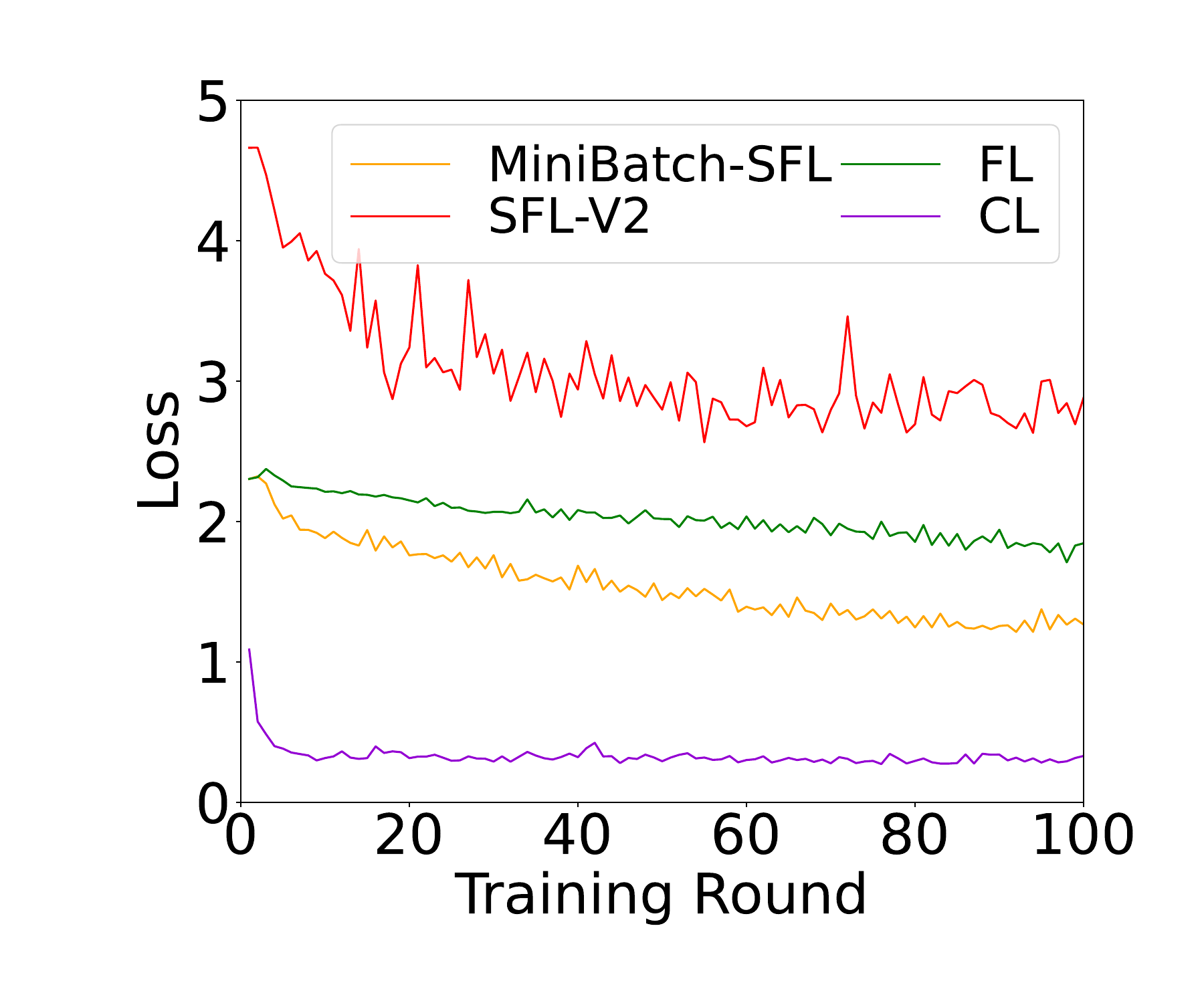}
        \caption{$r\%=0.95$.}
    \end{subfigure}
    \caption{Loss comparison with $N=10$ on CIFAR-10.}
    \label{fig:loss-comparison-10}
\end{figure}

\begin{figure}[h]
    \centering
     \begin{subfigure}{0.23\textwidth}
        \centering
        \includegraphics[height=3.4cm]{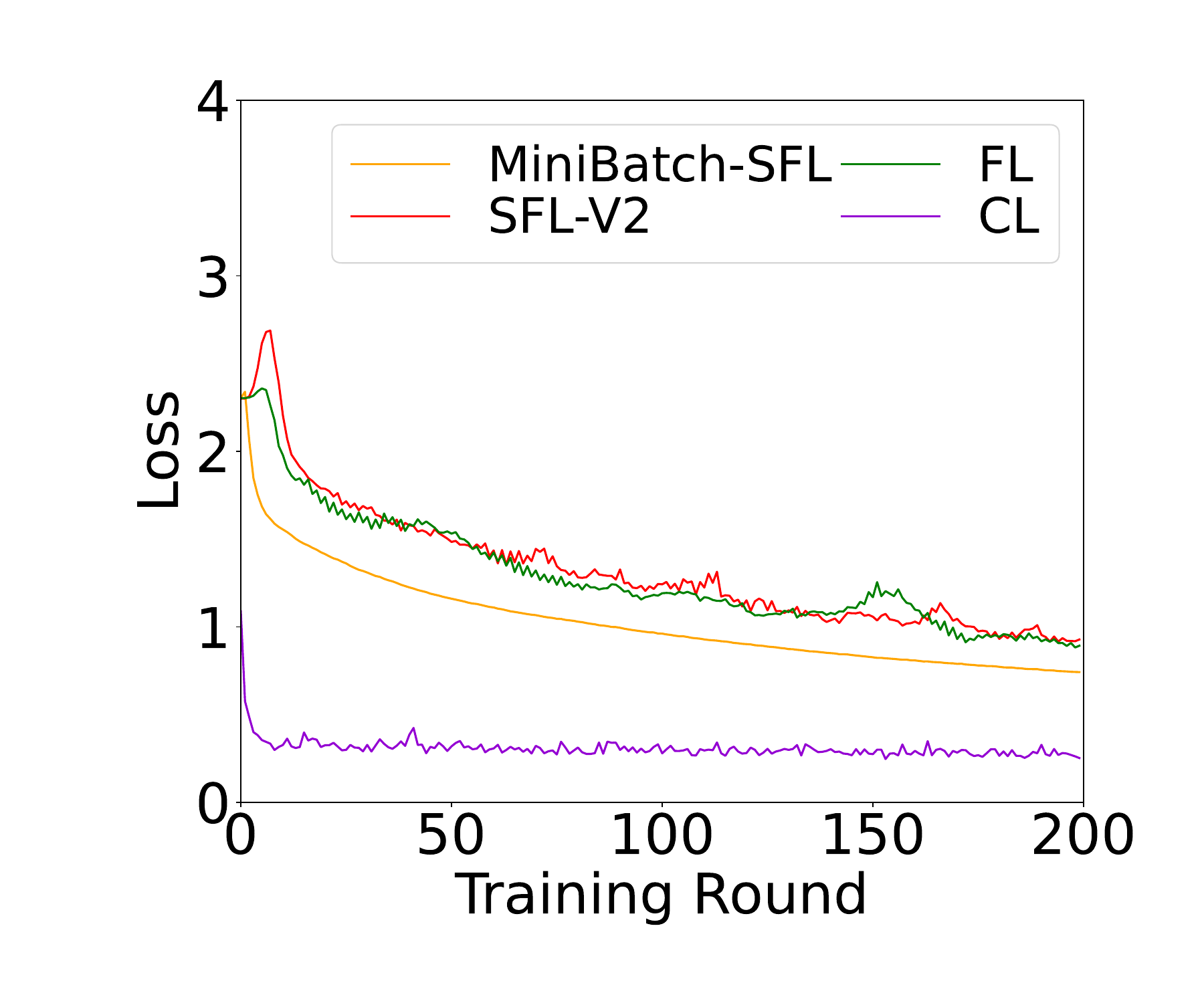}
        \caption{$r\%=0.5$.}
    \end{subfigure}
    \hfil
    \begin{subfigure}{0.23\textwidth}
        \centering
        \includegraphics[height=3.4cm]{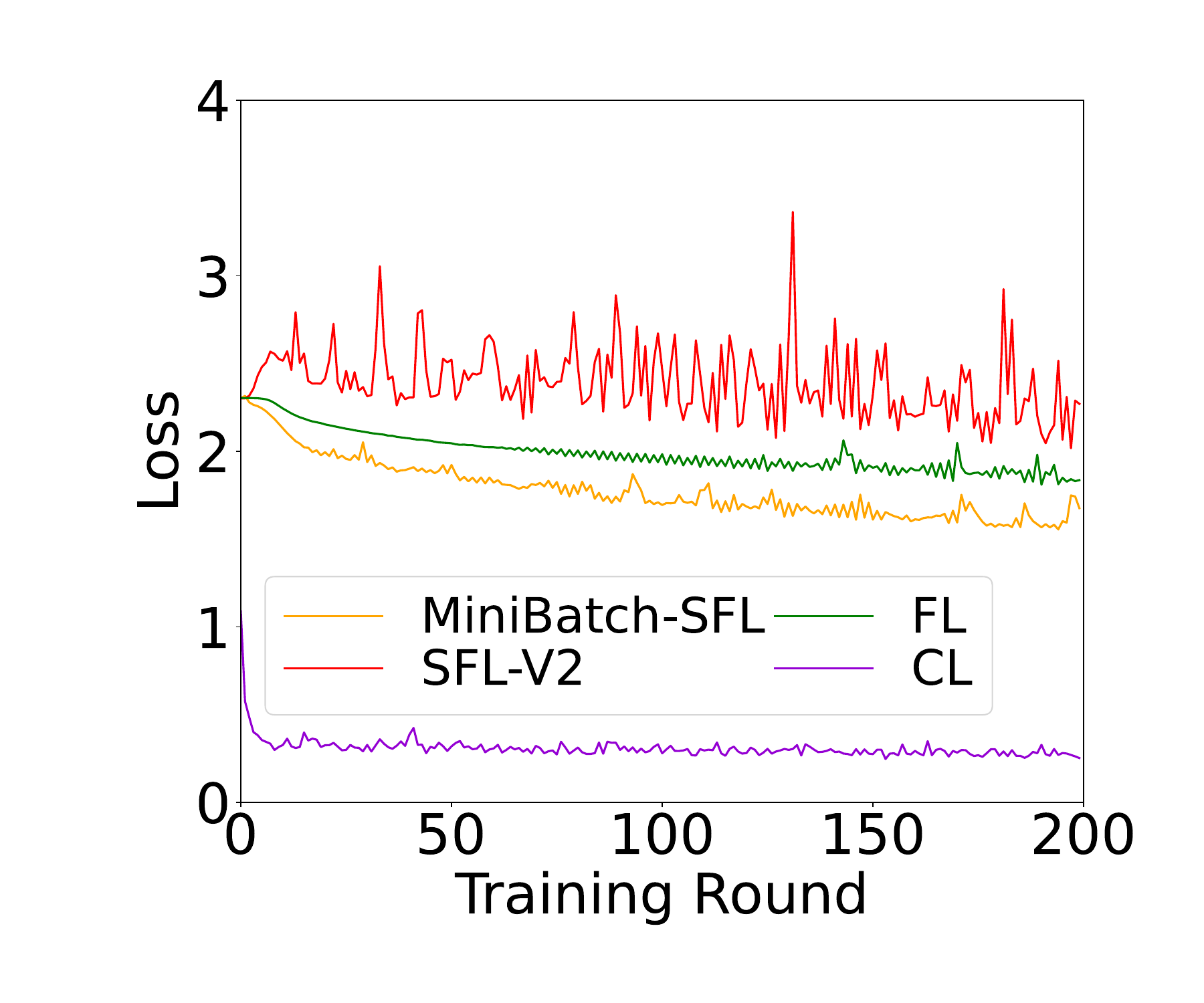}
        \caption{$r\%=0.95$.}
    \end{subfigure}
    \caption{Performance comparison with $N=100$ on CIFAR-10.}
    \label{fig:loss-comparison-100}
\end{figure}

Next, we present the loss across different algorithms in Fig. \ref{fig:loss-comparison-10} ($N=10$) and Fig. \ref{fig:loss-comparison-100} ($N=100$).  

From both Figs. \ref{fig:loss-comparison-10} and \ref{fig:loss-comparison-100}, we observe that MiniBatch-SFL achieves a smaller loss than FL and SFL-V2. This is consistent with Observation 5 in the main paper, showing that the proposed MiniBatch-SFL leads to a better trained model. 

\newpage 
\subsection{Section 4.3: More Comparison Results on FMNIST}

\subsubsection{Impact of cut layer in MiniBatch-SFL} 
\begin{figure}[h]
       \begin{subfigure}{0.24\textwidth}
        \centering
        \includegraphics[height=3.4cm]{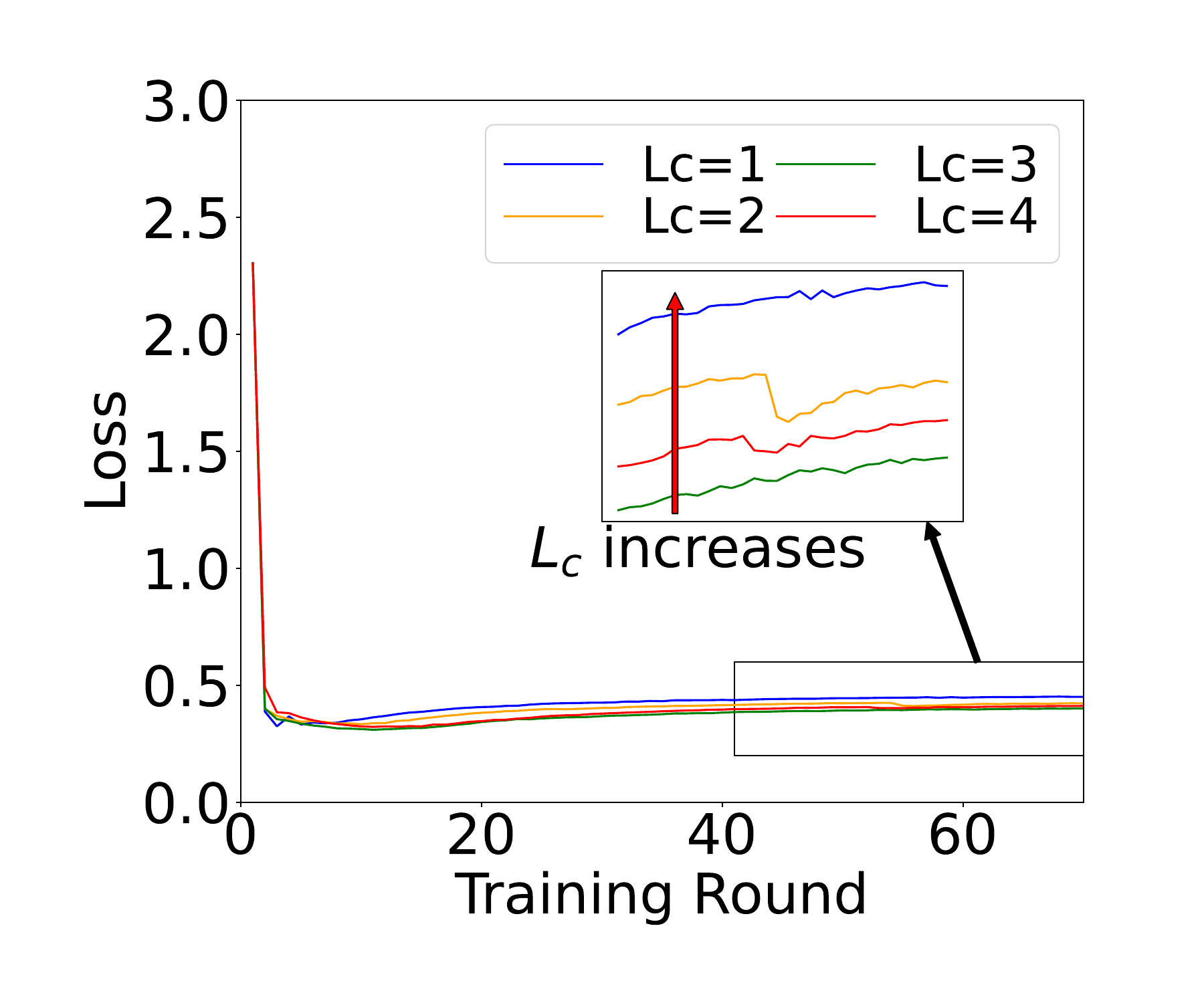}
        \caption{loss with $r\%=0$.}
    \end{subfigure}
    \hfil
    \begin{subfigure}{0.24\textwidth}
        \centering
        \includegraphics[height=3.4cm]{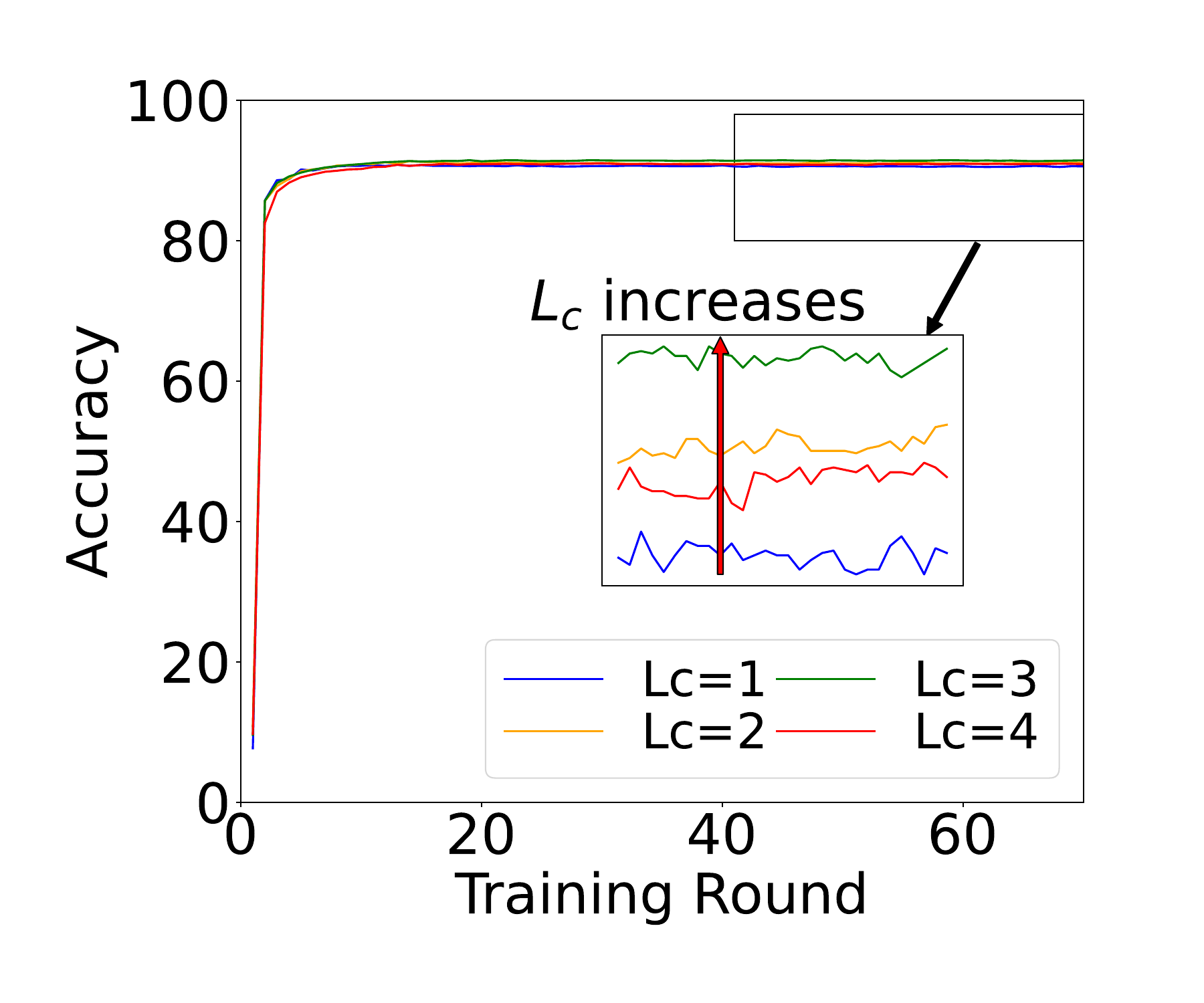}
        \caption{accuracy with $r\%=0$.}
    \end{subfigure}
    \hfil
    \begin{subfigure}{0.24\textwidth}
        \centering
        \includegraphics[height=3.4cm]{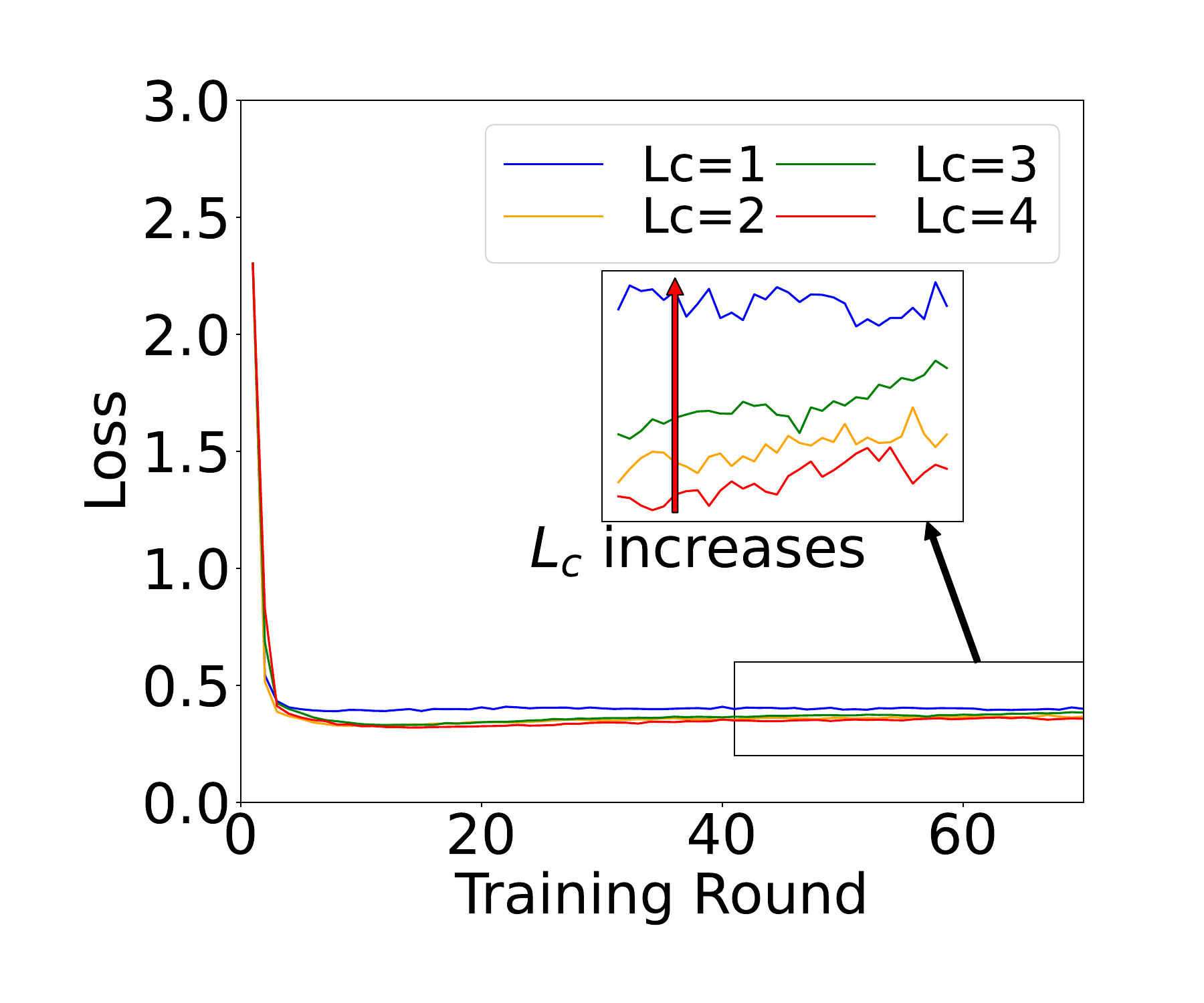}
        \caption{loss with $r\%=0.5$.}
    \end{subfigure}
    \hfil 
    \begin{subfigure}{0.24\textwidth}
        \centering
        \includegraphics[height=3.4cm]{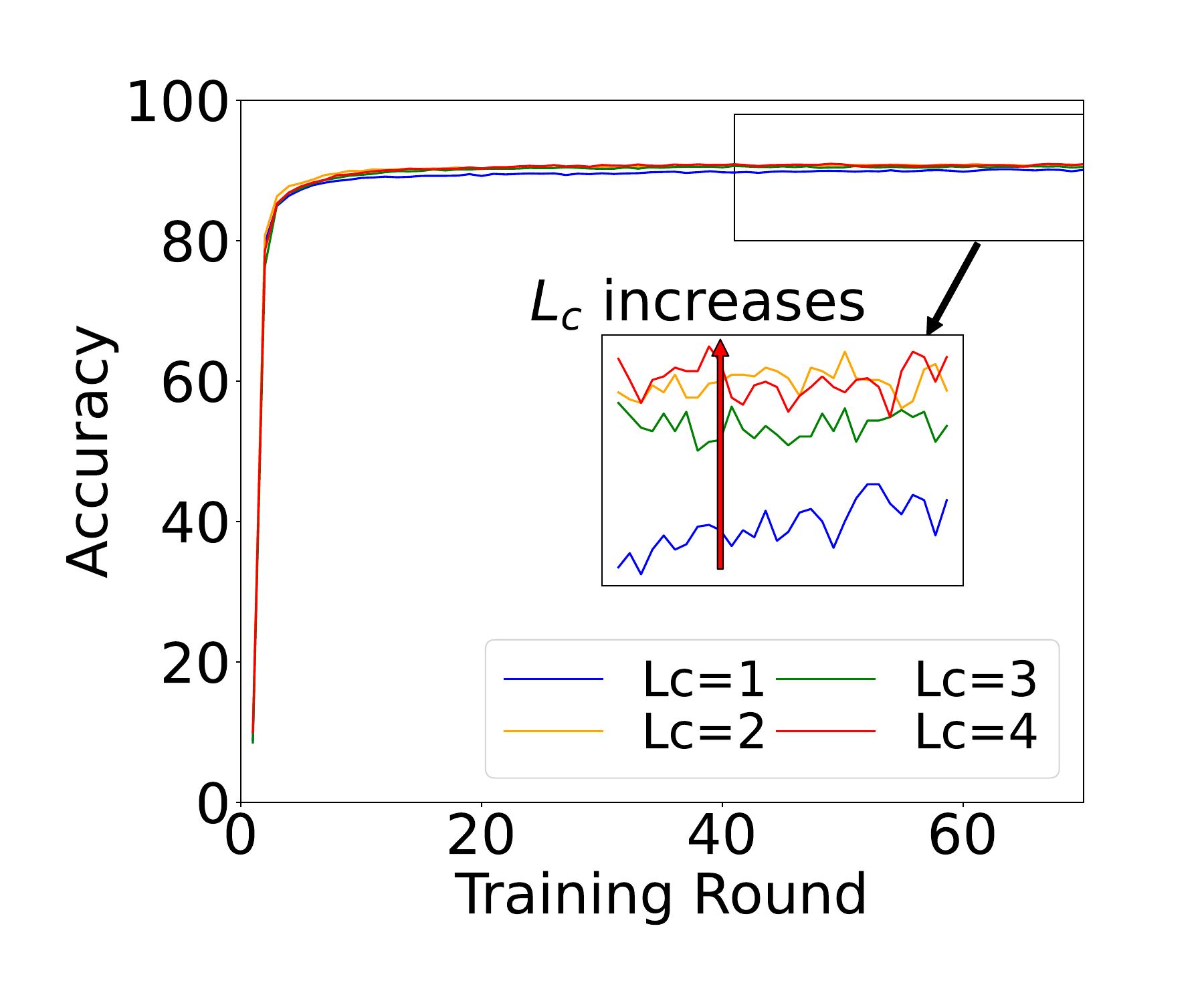}
        \caption{accuracy with $r\%=0.5$.}
    \end{subfigure}
    \caption{Impact of cut layer with $N=10$ on FMNIST.}
    \label{fig:cut-fmnist-10}
\end{figure}

We first study the impact of cut layer on MiniBatch-SFL and report the results in Fig. \ref{fig:cut-fmnist-10}. From Fig. \ref{fig:cut-fmnist-10}, we observe that the impact of cut later is minor at $r\%=0$. The impact becomes a bit more significant when $r\%=0.5$, where a larger $L_c$ corresponds to a smaller loss and larger accuracy.  

\subsubsection{Performance comparison}
\begin{figure}[h]
       \begin{subfigure}{0.24\textwidth}
        \centering
        \includegraphics[height=3.4cm]{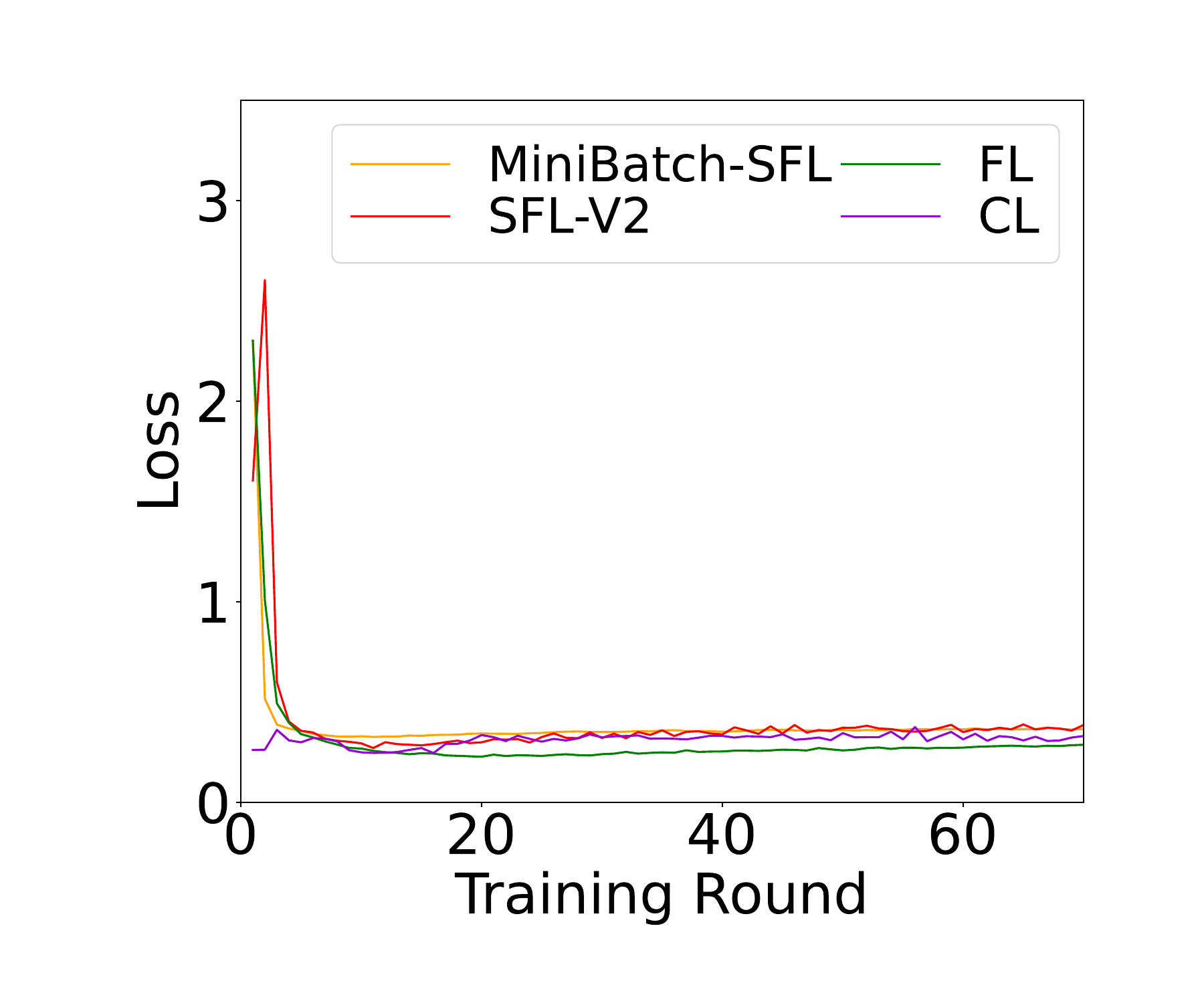}
        \caption{$r\%=0.5$.}
    \end{subfigure}
    \hfil
    \begin{subfigure}{0.24\textwidth}
        \centering
        \includegraphics[height=3.4cm]{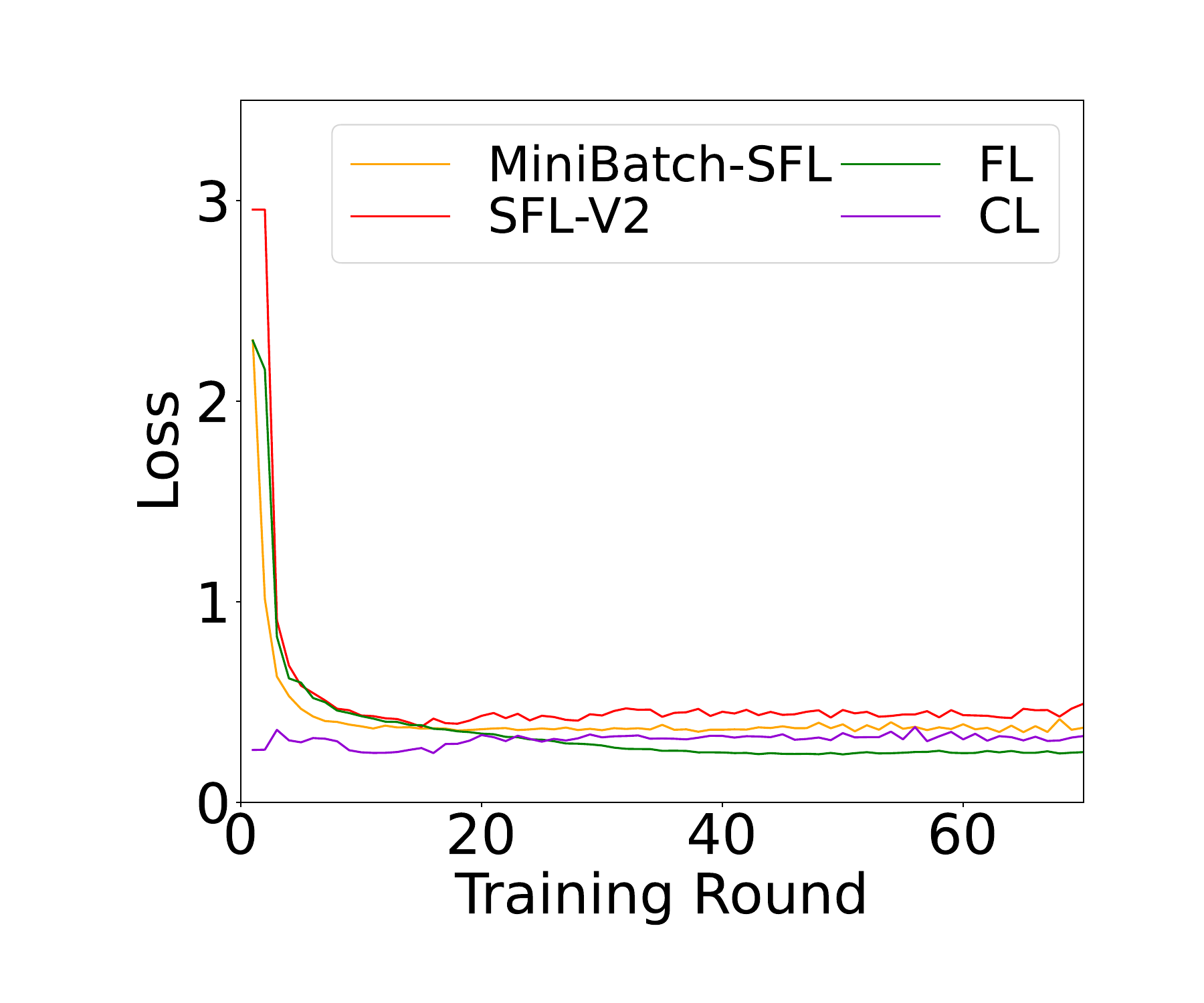}
        \caption{$r\%=0.8$.}
    \end{subfigure}
    \hfil
    \begin{subfigure}{0.24\textwidth}
        \centering
        \includegraphics[height=3.4cm]{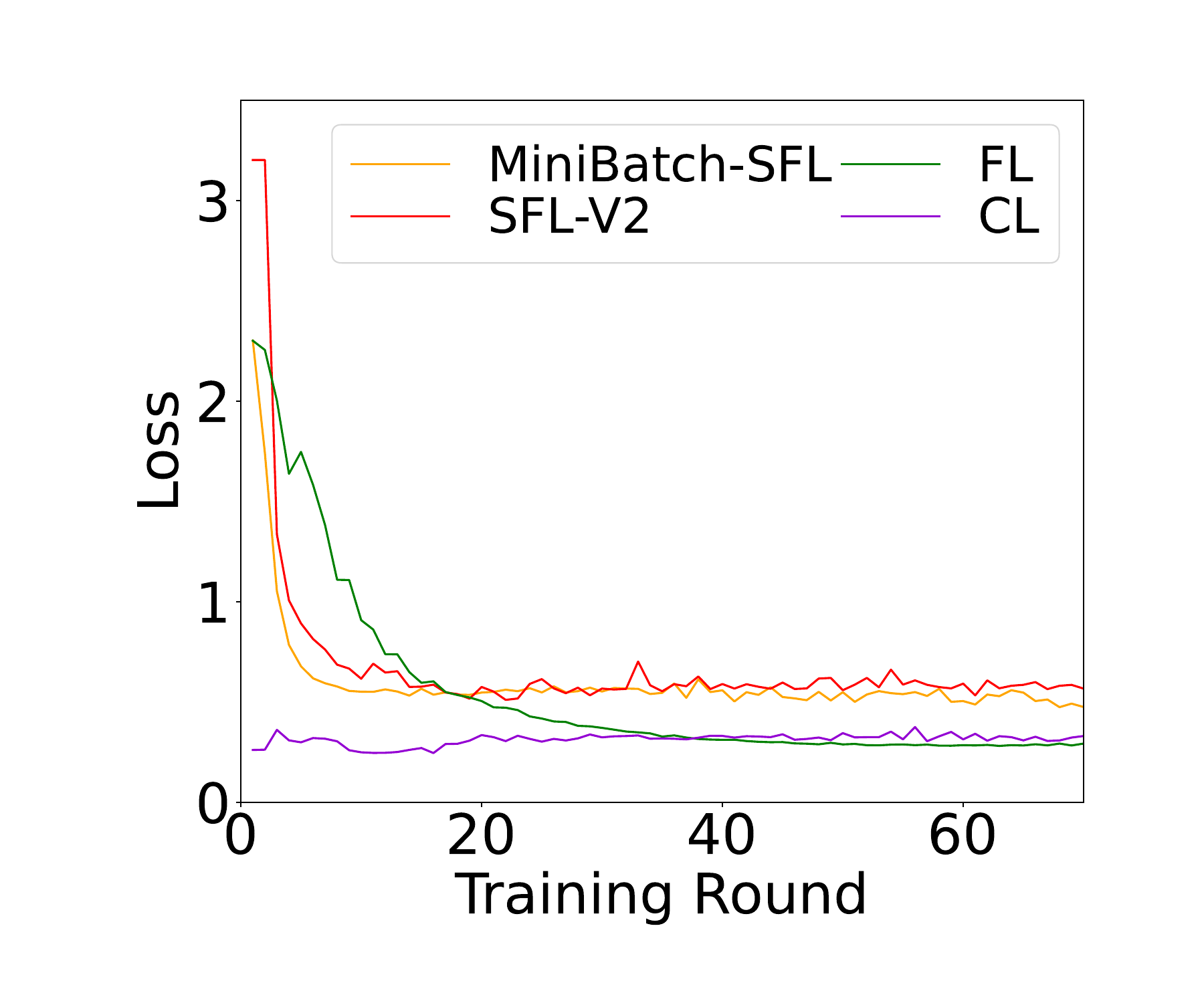}
        \caption{$r\%=0.9$.}
    \end{subfigure}
    \hfil 
    \begin{subfigure}{0.24\textwidth}
        \centering
        \includegraphics[height=3.4cm]{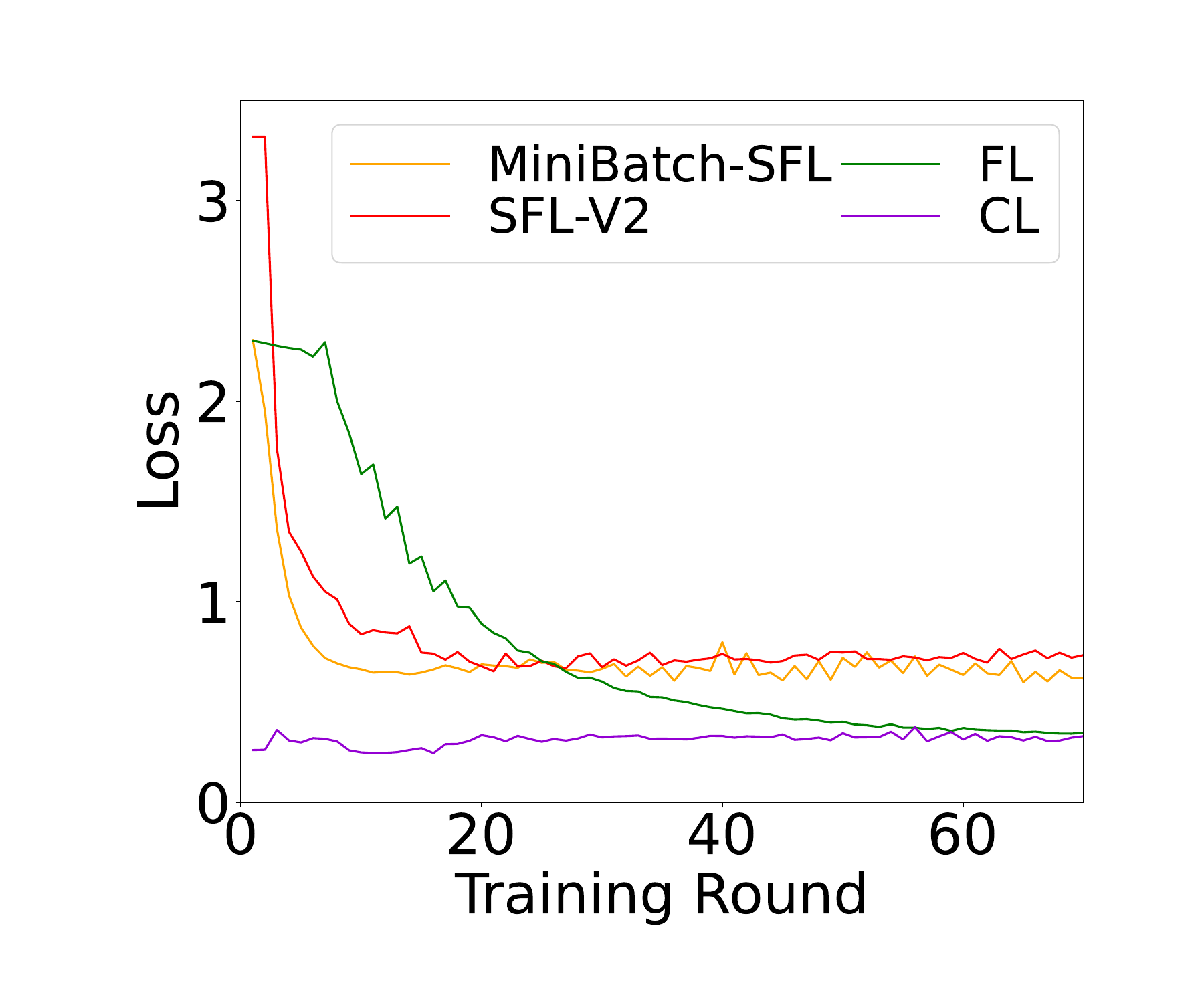}
        \caption{$r\%=0.95$.}
    \end{subfigure}
    \caption{Loss comparison with $N=10$ on FMNSIT.}
    \label{fig:loss-comparison-10-fmnist}
\end{figure}

\begin{figure}[h]
       \begin{subfigure}{0.24\textwidth}
        \centering
        \includegraphics[height=3.4cm]{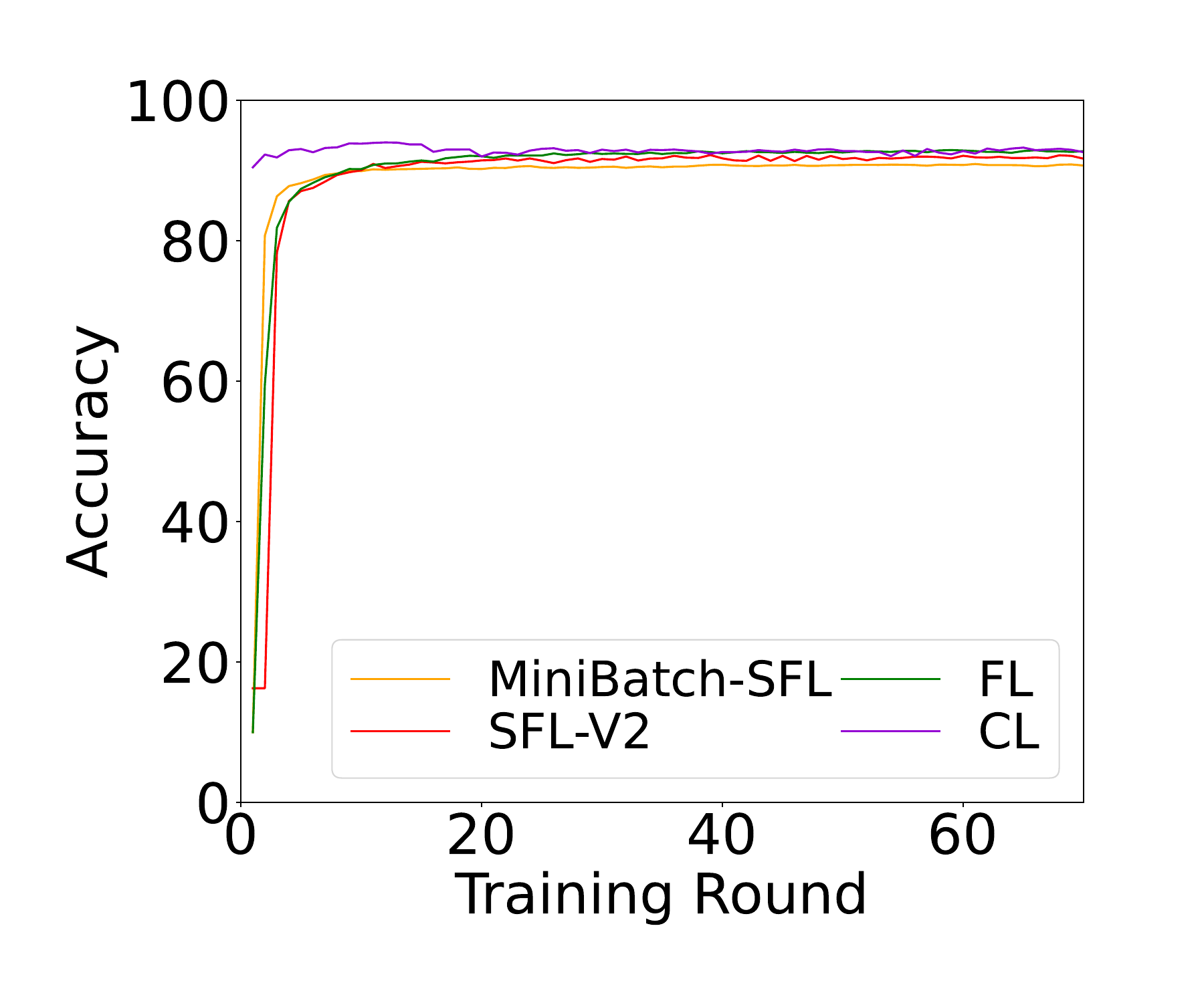}
        \caption{$r\%=0.5$.}
    \end{subfigure}
    \hfil
    \begin{subfigure}{0.24\textwidth}
        \centering
        \includegraphics[height=3.4cm]{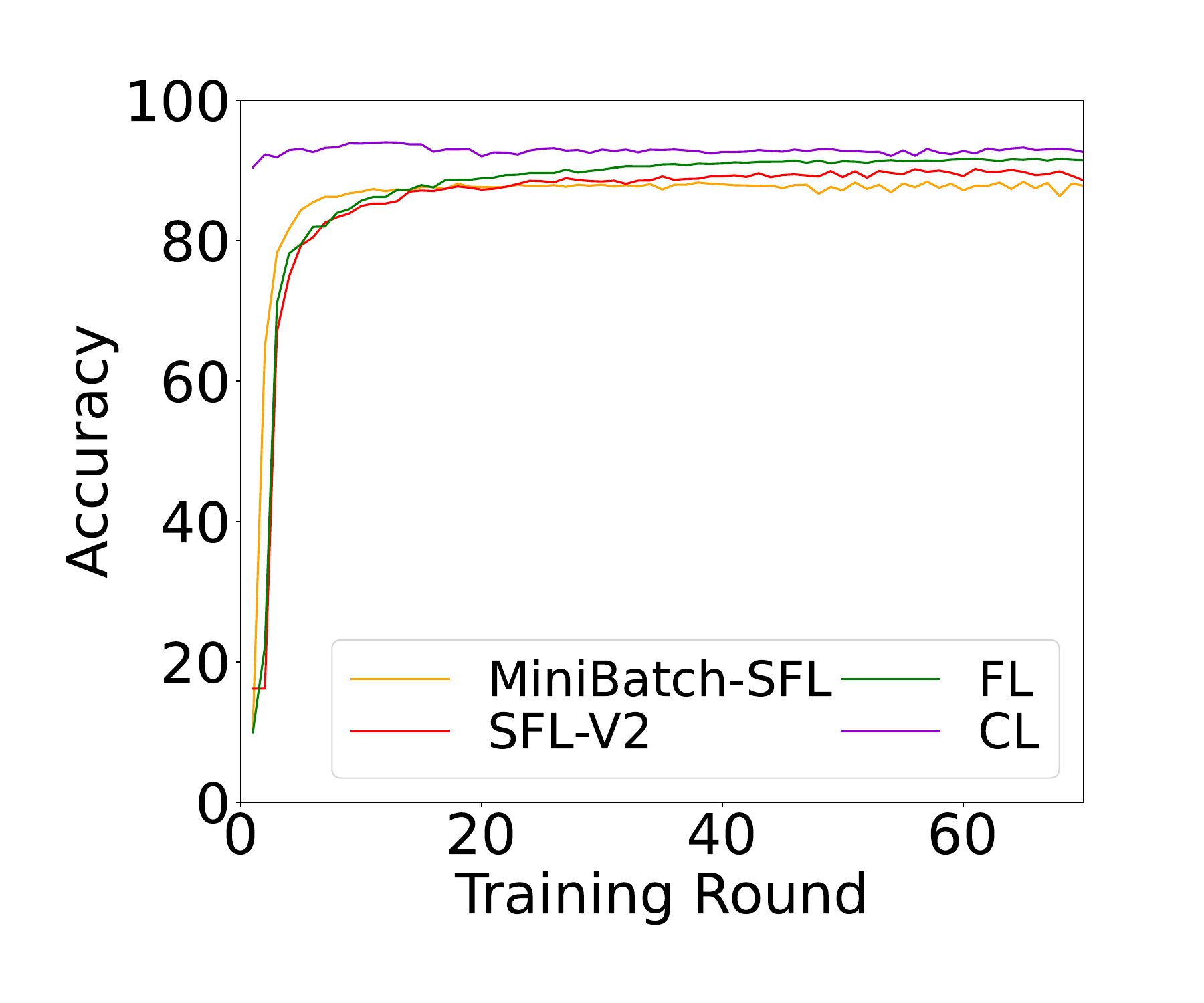}
        \caption{$r\%=0.8$.}
    \end{subfigure}
    \hfil
    \begin{subfigure}{0.24\textwidth}
        \centering
        \includegraphics[height=3.4cm]{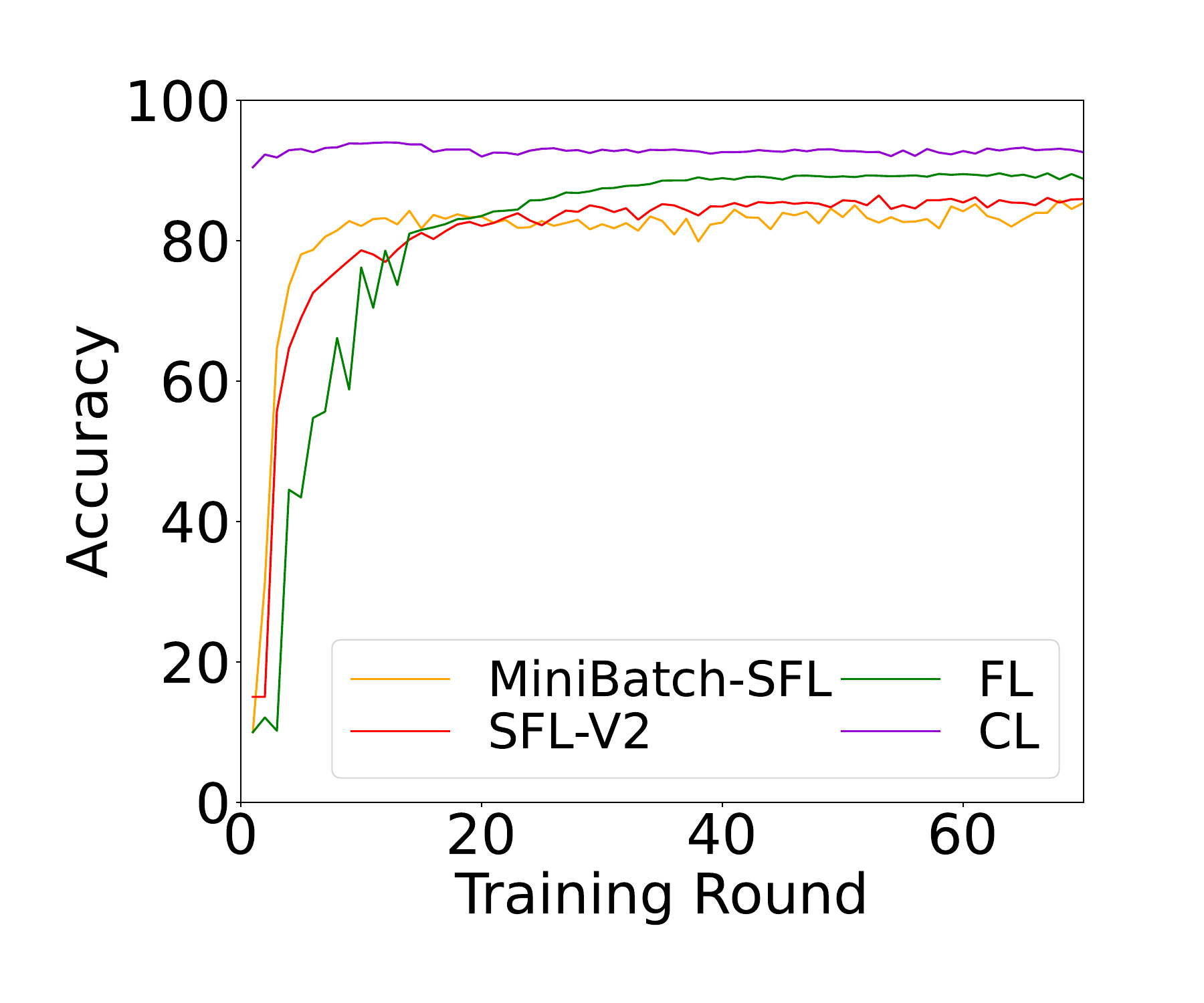}
        \caption{$r\%=0.9$.}
    \end{subfigure}
    \hfil 
    \begin{subfigure}{0.24\textwidth}
        \centering
        \includegraphics[height=3.4cm]{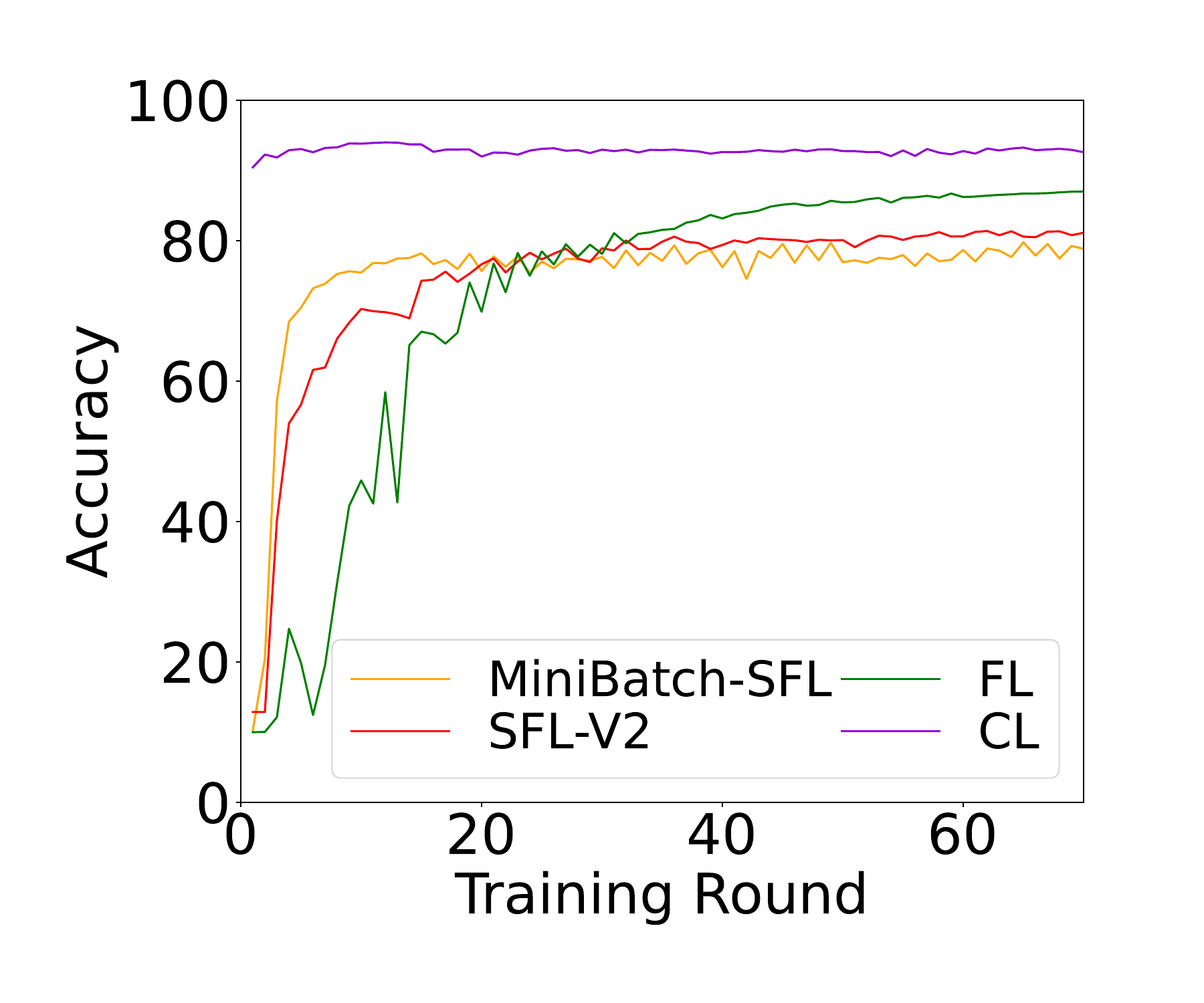}
        \caption{$r\%=0.95$.}
    \end{subfigure}
    \caption{Accuracy comparison with $N=10$ on FMNSIT.}
    \label{fig:acc-comparison-10-fmnist}
\end{figure}

\begin{figure}[h]
       \begin{subfigure}{0.24\textwidth}
        \centering
        \includegraphics[height=3.4cm]{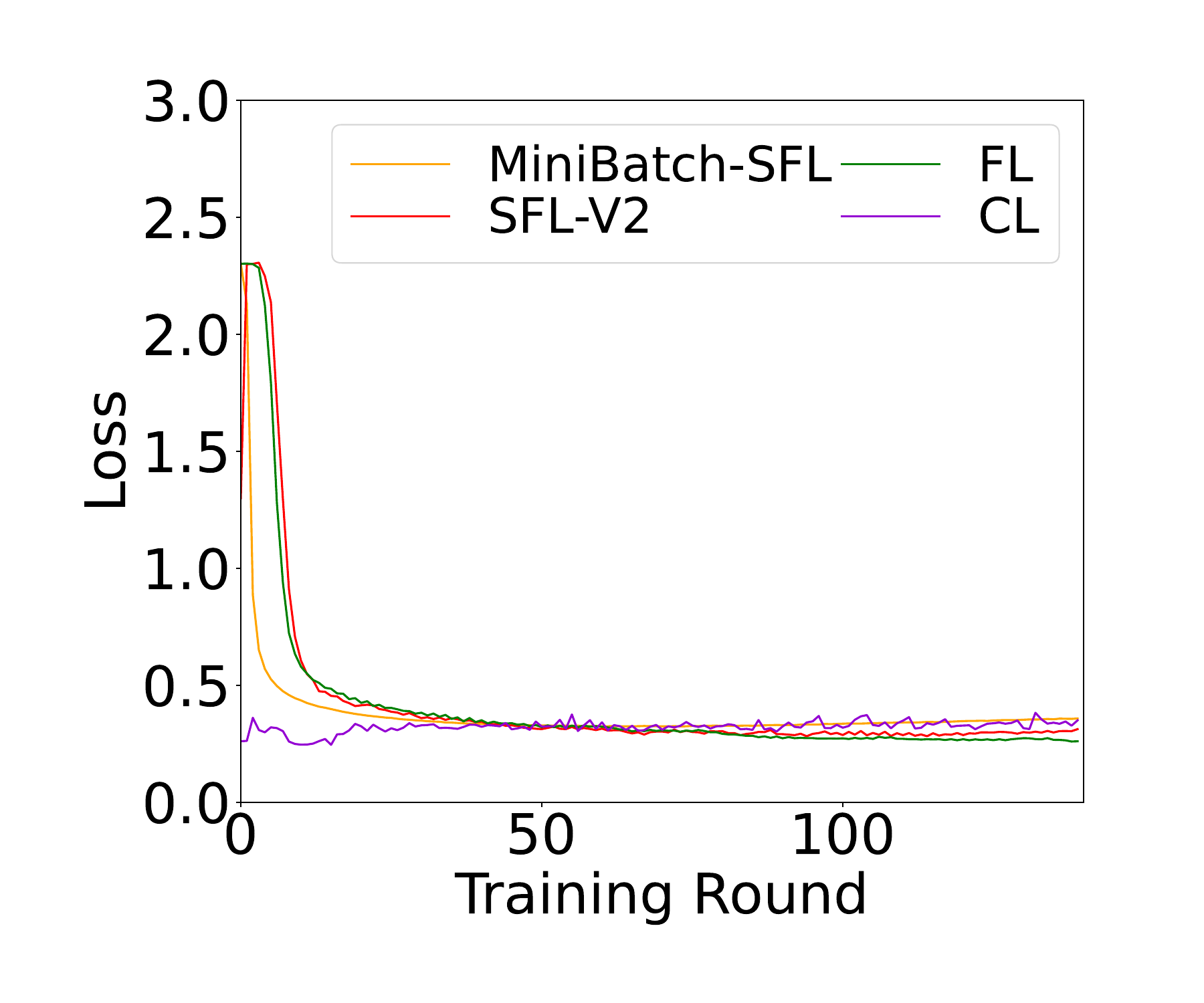}
        \caption{loss at $r\%=0.5$.}
    \end{subfigure}
    \hfil
    \begin{subfigure}{0.24\textwidth}
        \centering
        \includegraphics[height=3.4cm]{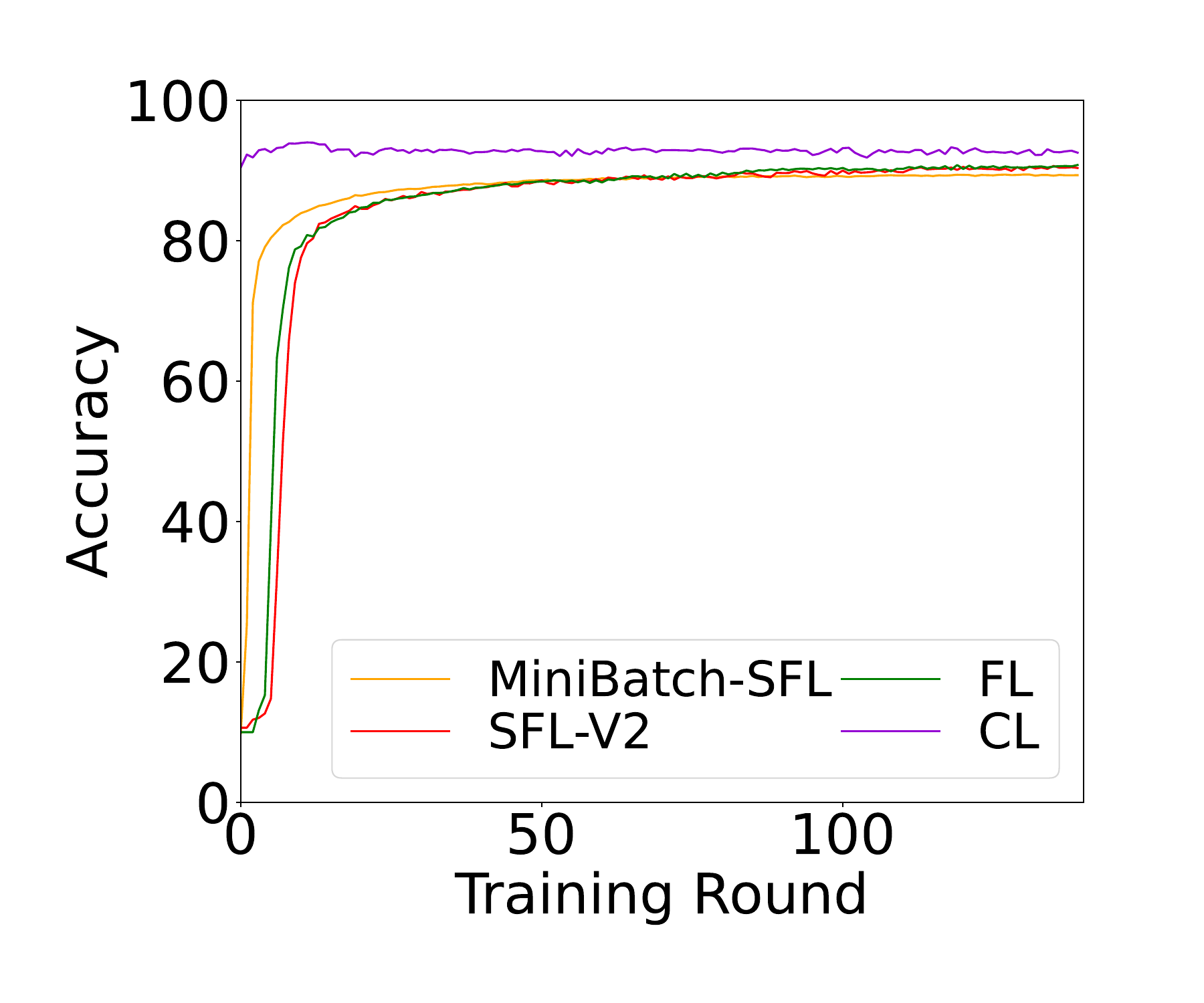}
        \caption{accuracy at $r\%=0.5$.}
    \end{subfigure}
    \hfil
    \begin{subfigure}{0.24\textwidth}
        \centering
        \includegraphics[height=3.4cm]{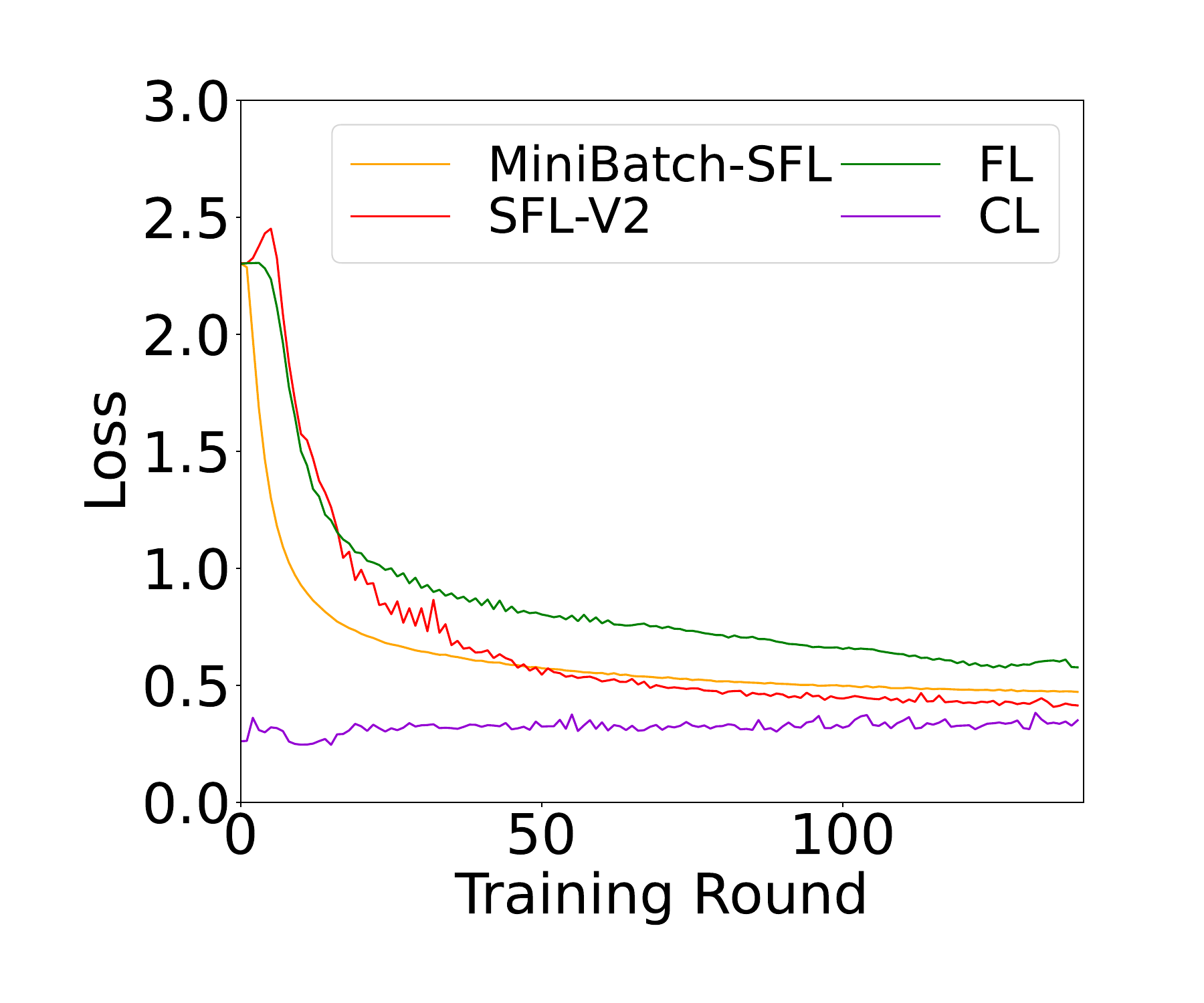}
        \caption{loss at $r\%=0.95$.}
    \end{subfigure}
    \hfil 
    \begin{subfigure}{0.24\textwidth}
        \centering
        \includegraphics[height=3.4cm]{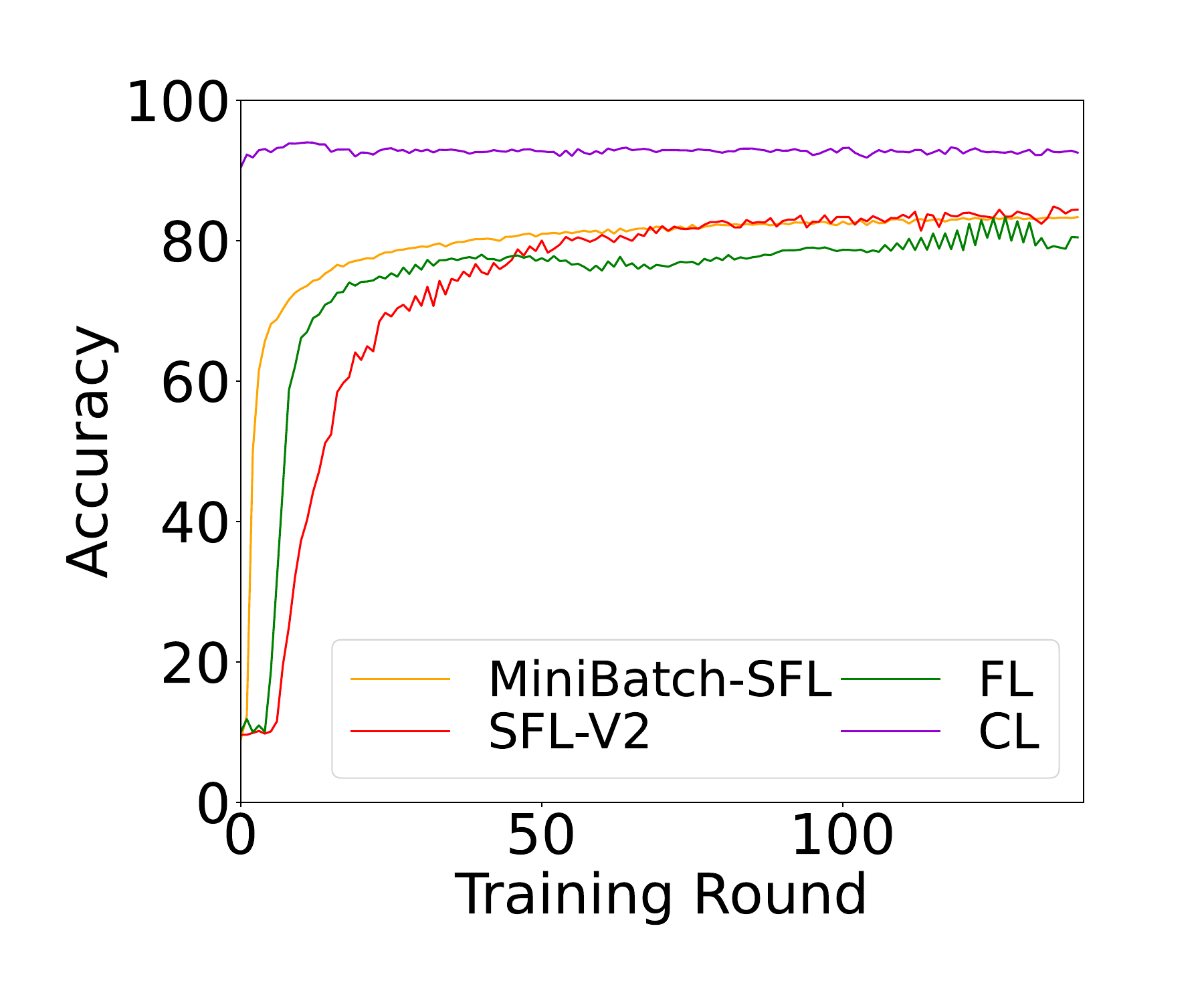}
        \caption{accuracy at $r\%=0.95$.}
    \end{subfigure}
    \caption{Performance comparison with $N=100$ on FMNSIT.}
    \label{fig:comparison-100-fmnist}
\end{figure}

Now we compare the algorithm performance on FMNIST. The results are reported in Figs. \ref{fig:loss-comparison-10-fmnist}-\ref{fig:acc-comparison-10-fmnist} ($N=10$) and Fig. \ref{fig:comparison-100-fmnist} ($N=100$).

From Fig. \ref{fig:comparison-100-fmnist}, we observe that the proposed MiniBatch-SFL has has a slightly lower loss and higher accuracy than FL and SFL. This is consistent with Observation 5 in the main paper. We argue that the improvement is not so significant mainly because FMNIST is easier to train than CIFAR-10.  It is an interesting future direction to test our algorithm on more complex datasets. 

Interestingly, from Figs. \ref{fig:loss-comparison-10-fmnist}-\ref{fig:acc-comparison-10-fmnist}, we observe that FL has a better performance than MiniBatch-SFL. This seems to contradict with our previous results. Some possible explanations for this can be due to randomness and a poor choice of hyper-parameters for MiniBatch-SFL on FMNIST. We are currently running more experiments to validate our hypothesis on this. Nevertheless, even if MiniBatch-SFL perform slightly worse than FL, it still converges faster than FL and SFL-V2.
\end{document}